\pdfoutput=1

\documentclass[11pt]{article}

\usepackage[preprint]{acl}

\usepackage{times}
\usepackage{latexsym}
\usepackage{amsmath}

\usepackage[T1]{fontenc}

\usepackage[utf8]{inputenc}

\usepackage{microtype}

\usepackage{inconsolata}

\usepackage{graphicx}
\usepackage{tabularx}
\usepackage{booktabs}
\usepackage{multirow}
\usepackage{rotating}
\usepackage[table]{xcolor}
\usepackage{listings}
\usepackage{xcolor}
\usepackage{soul}
\DeclareMathOperator*{\argmax}{arg\,max}

\title{Enhancing Factuality through Consensus and Consistency in Summarization Using Minimum Bayes Risk Decoding}

\author{
 \textbf{Riza Setiawan Soetedjo}\textsuperscript{1},
 \textbf{Yusuke Sakai}\textsuperscript{1},
 \textbf{Hidetaka Kamigaito}\textsuperscript{1},
 \textbf{Jingun Kwon}\textsuperscript{2},
\\
 \textbf{Manabu Okumura}\textsuperscript{3},
 \textbf{Taro Watanabe}\textsuperscript{1}
\\
 \textsuperscript{1}Nara Institute of Science and Technology (NAIST)
 \textsuperscript{2}Chungnam National University
\\
 \textsuperscript{3}Institute of Science Tokyo
\\
 \texttt{riza.setiawan\_soetedjo.rs6@naist.ac.jp}
\\
 \texttt{\{sakai.yusuke.sr9, kamigaito.h, taro.watanabe\}@is.naist.jp}
\\
 \texttt{jingun.kwon@cnu.ac.kr, oku@first.iir.isct.ac.jp}
}

\begin{document}
\maketitle
\begin{abstract}
Improving the quality of model-generated summaries, especially factuality, the accuracy of a summary with respect to its source content, remains a challenge. While reranking could select the optimal output from multiple generated candidates, it is limited to only using the source as guidance, resulting in unreliable summaries. To address this limitation, we propose ConSUM that reranks candidate summaries by considering two factors: \emph{consistency} to the source document and \emph{consensus} among the other candidates. Consensus is established using Minimum Bayes Risk (MBR) decoding over the set of generated summaries, while ensuring consistency by employing factuality-aware metrics that compare the summary against the source. Rigorous testing demonstrates that our system is competitive with existing methods, with human evaluations further confirming that its generated summaries are preferred over those from other systems. Our code is available at \href{https://github.com/naist-nlp/ConSUM}{https://github.com/naist-nlp/ConSUM}.

\end{abstract}

\section{Introduction}

Document Summarization is a task to summarize a lengthy document while retaining its most important information. Thus, to evaluate a generated summary, we should focus on its factuality, i.e., how the generated summary aligns with the original document, as well as commonly used metrics in Natural Language Generation (NLG), such as fluency and coherence~\cite{Fabbri_EtAl-2021-SummEvalReevaluatingSummarization}. 

Reranking is an established method to improve summary quality, including factuality \cite{Sul_Choi-2023-BalancingLexicalSemantic, Ravaut_EtAl-2022-SummaRerankerMultiTaskMixtureofExperts}, which involves generating multiple candidate summaries and ranking them using specific metrics reflecting summarization quality. Since we cannot use gold references, reference-free metrics are used to rerank the candidates by using, e.g., the source document as their ground truth~\cite{Dixit_EtAl-2023-ImprovingFactualityAbstractive}.

\begin{figure}[t]
    \centering
    \includegraphics[width=1.0\linewidth]{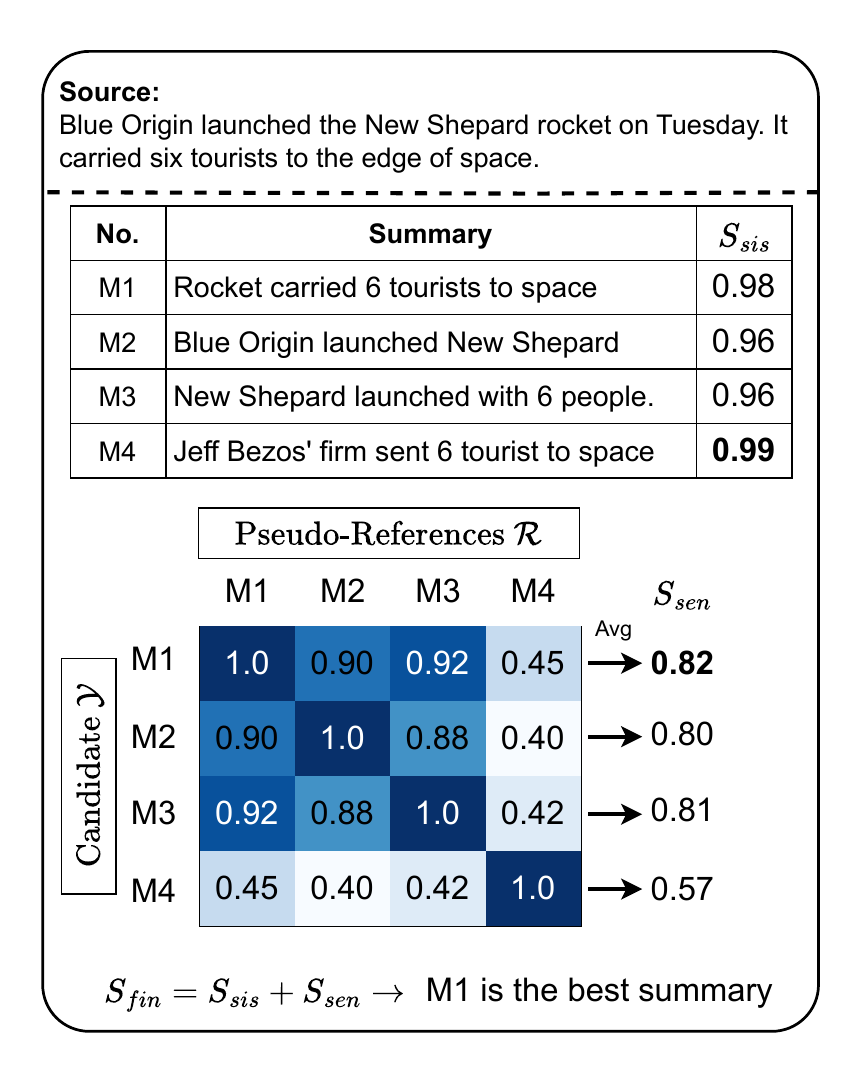}
    \caption{Case where reference-free metric resulted in factual inconsistency. $S_{sis}$ is consistency score to the source document; $S_{sen}$ is consensus score over among candidate summaries; $S_{fin}$ is the combined score.}
    \label{fig:analogy}
\end{figure}

Even though the success, the current reranking in summarization has some limitations: First, it cannot use reference-based metrics that correlate well to human evaluation due to requiring gold references; 
Second, relying only on the source can result in factual inconsistencies by failing to penalize specific errors due to the broad scoring nature of reference-free metrics, as shown in Figure \ref{fig:analogy}; 
Third, relying only on a single metric may cause overfitting to that metric and inherit its bias.

To address these problems, we propose a novel reranking method, \textbf{Con}sistency and \textbf{Con}sensus in \textbf{Sum}marization (ConSUM), that combines two factors: the \emph{consensus} among candidate summaries and the \emph{consistency} of each summary to the source document for factual summary generation. Inspired by Minimum Bayes Risk (MBR) decoding \cite{Eikema_Aziz-2020-MAPDecodingAll}, we achieve consensus by identifying the most representative summary among the generated candidates, relying on a reference-based factuality metric using pseudo-references sampled from the same model to serve as proxies for the ground truth. Consistency for the source document, meanwhile, is measured using reference-free metrics. Our method operates in a simple three-step process: (1) \textbf{Generate} multiple candidate summaries and pseudo-references given the source text; (2) \textbf{Score} each candidate using both MBR decoding and reference-free metric; and (3) \textbf{Rerank} each candidate by combining both scores and select the one with the highest final score.
 
We benchmarked our method on the CNN/DailyMail \cite{Nallapati_EtAl-2016-AbstractiveTextSummarization} and XSum \cite{Narayan_EtAl-2018-Don`tGiveMe} datasets using three Pretrained Language Models (PLMs) and a Large Language Model (LLM), evaluated in two categories of metrics: Quality, e.g., ROUGE \cite{Lin-2004-ROUGEPackageAutomatic} and Factuality, e.g, MENLI \cite{Chen_Eger-2023-MENLIRobustEvaluation}. Furthermore, we did extensive experiments to find the optimal hyperparameters in ConSUM and their influence on summarization quality. Finally, we conducted a human evaluation to assess our method qualitatively. Experimental results consistently demonstrate that our method achieves superior factuality scores across both automatic and human evaluations. This shows that combining \emph{consensus} and \emph{consistency} succeeded in improving the factuality of a summary.

\section{Related Work}
\label{sec:related_work}

\paragraph{Improving Factuality in Summary}
Current approaches to improving factuality often use existing evaluation metrics to improve the summarization model. Research in this area typically either specifies only improving factuality \cite{Liu_EtAl-2023-ImprovingSummarizationFactual} or attempts to balance multiple aspects simultaneously \cite{Ryu_EtAl-2024-MultiDimensionalOptimizationTexta, Dixit_EtAl-2023-ImprovingFactualityAbstractive}. However, these methods face two limitations: requiring high-quality gold references, or relying solely on the source document for guidance. This reliance on a single reference signal arises because creating human-written ``gold'' summaries for comparison is notoriously labor-intensive. On the other hand, only using the source document is unreliable. 

\paragraph{Reranking in Summarization}
Reranking in summarization is a two-step approach: (1) a summarization model generates multiple summaries based on the source document; (2) a reranker model reranks the summaries based on a specific evaluation metric. \citet{liu_simcls_2021} trains a reranker model using ROUGE, while \citet{Dixit_EtAl-2023-ImprovingFactualityAbstractive} trains a reranker model using FactCC and ROUGE to balance between factuality and quality. As an alternative to train a supervised reranker model, some studies have proposed methods to rerank in an unsupervised setup \cite{ravaut-etal-2023-unsupervised, suzgun-etal-2023-follow}. However, they only use quality metrics such as ROUGE and BERTScore. 

\paragraph{Minimum Bayes Risk (MBR) Decoding}
MBR decoding was used in Neural Machine Translation (NMT) to address the mode-seeking problem inherent in standard Maximum a Posteriori (MAP) decoding \cite{Eikema_Aziz-2020-MAPDecodingAll}. Instead of selecting a single most probable output, MBR first generates a pool of candidate texts, then selects the candidate that maximizes an expected utility function when scored against a set of references, which are usually generated by the same model \cite{ohashi_true_2024}. The utility function itself is typically an existing evaluation metric. While studies show that MBR effectively optimizes for whichever metric is chosen as the utility function \cite{bertsch_its_2023}, this can lead to a form of reward hacking or metric bias. Although this issue has been extensively researched in NMT \cite{kovacs_mitigating_2024, muller_understanding_2021}, its impact within the summarization task remains largely unexplored.

\begin{figure}[t]
    \includegraphics[width=0.85\columnwidth]{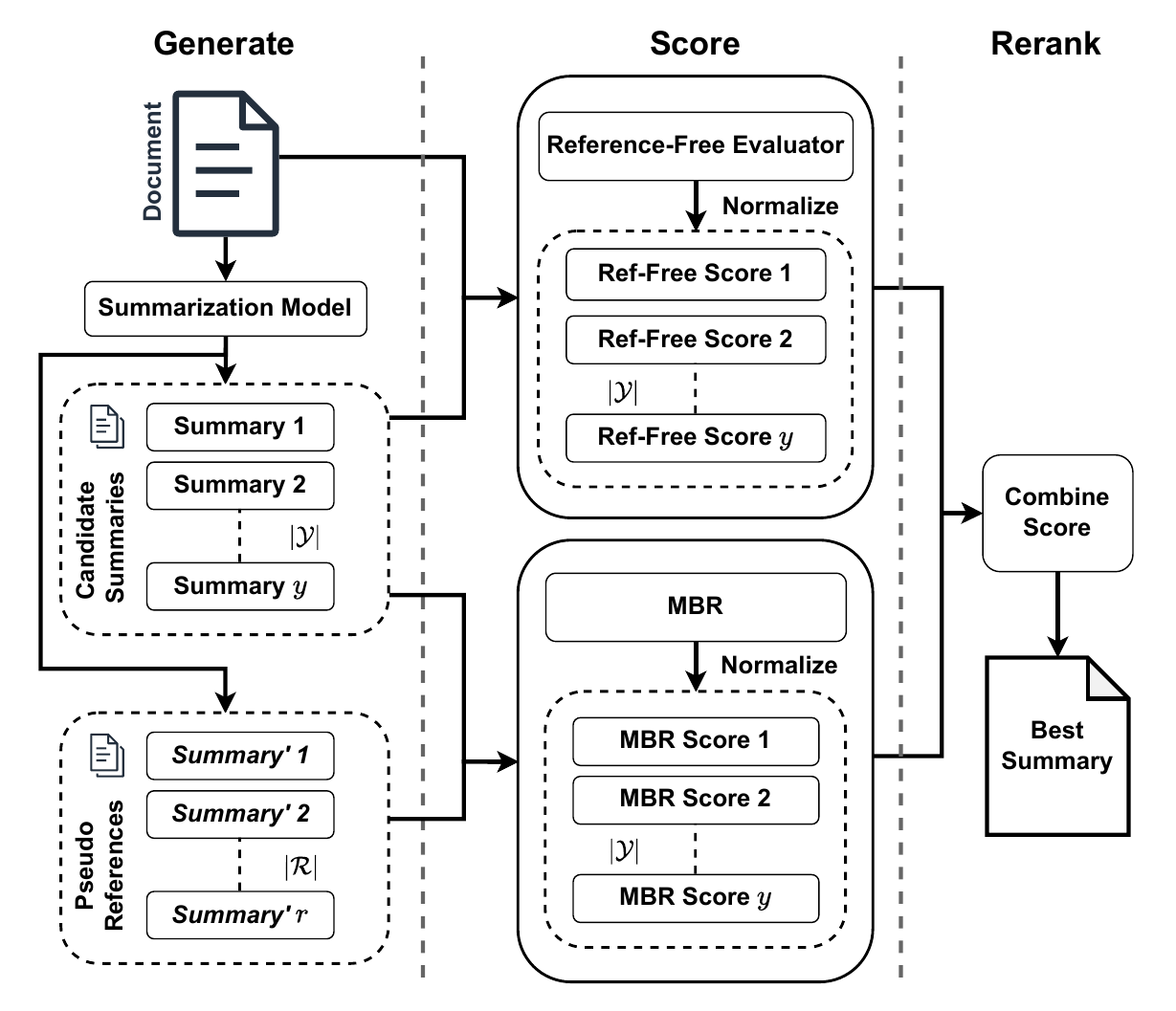}
    \caption{Overview of ConSUM comprising three steps: Generate, Score, and Rerank.}
    \label{fig:methodology}
\end{figure}

\section{Proposed Method: ConSUM}

The overview of ConSUM is presented in Figure \ref{fig:methodology}. It comprises three steps: (1) \textit{Generate Candidates and Pseudo-references} -- The summarization model generates multiple candidate and pseudo-references summaries given the source document; (2) \textit{Score Candidates} -- Each candidate is scored using two distinct evaluators, one assessing its consistency to the source document and the other assessing the consensus between the candidates and the pseudo-references; (3) \textit{Rerank Candidates} -- Both scores are combined and the candidate with the highest overall score is selected as the final output.

\subsection{Generate Candidates and Pseudo-references}
\label{subsec:gen_cand}

To enhance factuality, we distinguish between the \textit{candidates} $\mathcal{Y} = \{\boldsymbol{y}_1, \cdots, \boldsymbol{y}_{|\mathcal{Y}|}\}$ (the pool of potential outputs) and the \textit{pseudo-references} $\mathcal{R} = \{\boldsymbol{r}_1, \cdots, \boldsymbol{r}_{|\mathcal{R}|}\}$ (proxies for the ground truth). Both are sampled from the same model $\theta$ given a source $s$, but they serve distinct roles and may use different decoding strategies, denoted as $\theta_{\text{cand}}$ and $\theta_{\text{ref}}$:
\begin{equation}
\label{eq:sample_gen}
    \mathcal{Y} \sim p(\boldsymbol{y}|s; \theta_{\text{cand}}) \quad \text{and} \quad \mathcal{R} \sim p(\boldsymbol{r}|s; \theta_{\text{ref}}).
\end{equation}
While prior work often sets $\mathcal{Y} = \mathcal{R}$ \cite{suzgun-etal-2023-follow}, we treat them as distinct sets. This allows us to optimize $\mathcal{R}$ to better capture the model's true distribution for consensus estimation \cite{kamigaito-etal-2025-diversity}, independent of the strategy used to generate diverse \textit{candidates} $\mathcal{Y}$ (see Appendix \ref{appendix:optimal_pseudo}).

\subsection{Score Candidates}

We assess the validity of a candidate $y$ using two signals: its \emph{consistency} with the source document and its \emph{consensus} with the pseudo-references $\mathcal{R}$. Consistency-based scores have been used widely to improve summarization \cite{liu_simcls_2021, Dixit_EtAl-2023-ImprovingFactualityAbstractive}. However, we posit that relying solely on the source is insufficient due to metric bias and a lack of granularity, as illustrated in Figure~\ref{fig:analogy}. Since high-quality gold references are often inaccessible, we leverage pseudo-references $\mathcal{R}$ as a proxy reference signal, used in conjunction with the source document. This allows us to mitigate counterfactual outliers by penalizing information that diverges from the model's consensus distribution.

\paragraph{Consistency-based Scoring}

We define the consistency between the source document $\boldsymbol{s}$ and its candidate $\boldsymbol{y}_i \in \mathcal{Y}$, generated in \S\ref{subsec:gen_cand}, as:
\begin{equation}
    S_{sis}(\boldsymbol{y}_i,\boldsymbol{s}) = FM(\boldsymbol{y}_i,\boldsymbol{s}),
\end{equation}
where $FM(\boldsymbol{y}_i,\boldsymbol{s})$ is a reference-free factuality metric. In this study, we use FENICE \cite{Scire_EtAl-2024-FENICEFactualityEvaluation} and FIZZ \cite{Yang_EtAl-2024-FIZZFactualInconsistency} as the reference-free metric. We chose the metrics because they represent the state-of-the-art (SOTA) factuality-based reference-free metrics for summarization task\footnote{as of January 2025}. Both metrics operate on a similar principle (see Appendix \ref{appendix:metric_def}).

\paragraph{Consensus-based Scoring}
\label{subsubsec:consensus}

We measure the consensus between a candidate $\boldsymbol{y}_i \in \mathcal{Y}$ and pseudo-references $\mathcal{R}$ by incorporating MBR decoding as follows:
\begin{equation}
    S_{sen}(\boldsymbol{y}_i,\mathcal{R}) = \frac{1}{|\mathcal{R}|}\sum_{j=1}^{|\mathcal{R}|} u(\boldsymbol{y}_i, \boldsymbol{r}_j),
\end{equation}
where $u(\boldsymbol{y}_i, \boldsymbol{r}_j)$ is a utility function to calculate the validity of each candidate $\boldsymbol{y}_i$ using each pseudo-reference $\boldsymbol{r}_j$. In this study, we specifically choose MENLI \cite{Chen_Eger-2023-MENLIRobustEvaluation} as the utility function, as it is a SOTA NLI-based metric designed specifically for summarization. The metric aligns with our goal of ``consensus'' driven by factual agreement, rather than just lexical or semantic popularity which are measured through ROUGE and BERTScore~\cite{ravaut-etal-2023-unsupervised,suzgun-etal-2023-follow}.

\subsection{Rerank Candidates}

To ensure a balanced contribution, we normalize and combine $S_{sis}$ and $S_{sen}$ as follows:
\begin{align}
    &S_{fin}(\boldsymbol{y}_i,\boldsymbol{s},\mathcal{R}) \nonumber\\ 
    =& w Z(S_{sen}(\boldsymbol{y}_i,\mathcal{R})) \!+\! (1\!-\!w) Z(S_{sis}(\boldsymbol{y}_i,\boldsymbol{s})).
\end{align}
$Z$ indicates z-score normalization and $w$ $(0 \leq w \leq 1)$ is a hyperparameter to adjust the importance of  $S_{sen}$, where 0 and 1 means the scoring only uses $S_{sis}$ and $S_{sen}$, respectively.
Finally, we choose the best candidate $\hat{\boldsymbol{y}}$ based on $S_{final}$ as follows:
\begin{equation}
    \hat{\boldsymbol{y}} = \argmax_{\boldsymbol{y}} S_{fin}(\boldsymbol{y},\boldsymbol{s},\mathcal{R}).
\end{equation}
Unlike prior work relying solely on reference-free metrics \cite{Dixit_EtAl-2023-ImprovingFactualityAbstractive} or limiting MBR to only use BERTScore \citep{suzgun-etal-2023-follow}, we combine both paradigms. We identify the best candidate by aggregating its \emph{consensus} score (using MENLI against the pseudo-references $\mathcal{R}$) and its \emph{consistency} score (using FENICE or FIZZ against the source document).

\section{Experiments}
We test our method on two news summarization datasets, using fine-tuned PLMs and a LLM as the summarization models, evaluated on metrics from two aspects: Quality and Factuality.

\begin{table}[!t]
    \centering
    \small
    \begin{tabular}{@{}lp{0.35\linewidth}p{0.35\linewidth}@{}}
    \toprule
    \textbf{Group} & \textbf{Reference-Based} & \textbf{Reference-Free} \\
    \midrule
    Quality & ROUGE, BERTScore, MoverScore & --- \\
    \midrule
    Factuality & MENLI & UniEval, FENICE, FIZZ, SimCLS \\
    \bottomrule
    \end{tabular}
    \caption{Metrics by group and type by reference.}
    \label{tab:eval_metrics}
\end{table}

\subsection{Experimental Settings}
\label{subsec:exp_setting}

\paragraph{Datasets}
We evaluated our method on two widely-used news summarization datasets: CNN/DailyMail (CNN/DM) \cite{Nallapati_EtAl-2016-AbstractiveTextSummarization} and XSum \cite{Narayan_EtAl-2018-Don`tGiveMe}. CNN/DM is characterized by its relatively extractive summaries, where summary sentences are often copied directly from the source article. In contrast, XSum is known for its highly abstractive, single-sentence summaries that condense the main point of an article. Although these are standard benchmarks, recent work \citep{Zhang_EtAl-2024-BenchmarkingLargeLanguage} has highlighted the low quality of their ``gold'' references. 
These noise allow us to demonstrate that ConSUM can improve factuality through the model's consensus and consistency with the source document.

\paragraph{Models}
We used three PLMs, i.e, BART~\cite{Lewis_EtAl-2020-BARTDenoisingSequencetoSequence}, PEGASUS~\cite{Zhang_EtAl-2020-PEGASUSPretrainingExtracted}, and T5~\cite{Raffel_EtAl-2023-ExploringLimitsTransfer}, which were fine-tuned to the respective datasets. In addition, we experimented Llama-3~\cite{grattafiori2024llama3herdmodels} to test our method with LLMs using nucleus sampling \cite{Holtzman2020The}\footnote{Implementation details are in Appendix \ref{appendix:summ_models}.}.
Despite generally outperforming PLMs, LLMs exhibit comparable summarization performance due to hallucinations \cite{chhabra-etal-2024-revisiting},  motivating us to evaluate both in this study.
Furthermore, we explored two types of decoding for the PLMs: Diverse Beam Search (DBS)~\cite{vijayakumar_diverse_2018} and Epsilon Sampling~\cite{hewitt-etal-2022-truncation}. For DBS, we implemented the setting from \citet{liu_simcls_2021} and refer to it as \texttt{beam-sim} in this study. On the other hand, we implemented the best epsilon setting from NMT \cite{Freitag_EtAl-2023-EpsilonSamplingRocks} and refer to it as \texttt{epsilon} in this study. We also explored alternative settings, as discussed in Appendix \ref{appendix:sampling_setting}.

\paragraph{Metrics}
Generated summaries are assessed on two metric groups: Quality and Factuality, as shown in Table \ref{tab:eval_metrics}. We used three reference-based metrics in the Quality group: ROUGE~\cite{Lin-2004-ROUGEPackageAutomatic}, BERTScore~\cite{Zhang_EtAl-2020-BERTScoreEvaluatingTexta}, and MoverScore~\cite{zhao_moverscore_2019}. In Factuality group, we utilized three scores provided by MENLI~\cite{Chen_Eger-2023-MENLIRobustEvaluation}, i.e. entailment (EM), contradiction (CM), and summarization (SM) as the reference-based metric,  UniEval~\cite{zhong_towards_2022}, FENICE~\cite{Scire_EtAl-2024-FENICEFactualityEvaluation}, and FIZZ~\cite{Yang_EtAl-2024-FIZZFactualInconsistency} as the reference-free metrics. Quality metrics are the standard benchmarks in summarization studies.  We also included UniEval to measure multiple dimensions alongside factuality (i.e. fluency and coherence). Although SimCLS underperformed as a reranking component (see \S\ref{subsec:optimal_weight}), we retained it as a strong reference-free evaluator to verify our results against a finetuned model. Statistical significance is assessed using paired-bootstrap resampling \cite{koehn-2004-statistical} with 10,000 iterations. We established a significance level of $p<0.05$, applying the Bonferroni correction to account for multiple comparisons against the three baselines\footnote{The significance test is done before the results are rounded.}. Implementation details are found in Appendix \ref{appendix:metric_def}.

\paragraph{Hyperparameters}
Our preliminary study (Appendix \ref{appendix:preliminary}) highlights the importance of using diverse and unbiased pseudo-reference sets, as well as maintaining the weight between consensus and consistency, to achieve optimal performance.
We determined the best hyperparameters to be \textbf{16 candidates}; \textbf{64 pseudo-references, generated via epsilon sampling}; and \textbf{a combination weight ($w$) of 0.75}. The weight ($w$) means the contribution of $S_{sen}$ and $S_{sis}$ is 75:25 to the $S_{final}$ score. In this study, our scorer methods are formatted as \texttt{Scorer-$w$}, where \texttt{Scorer} refers to either FENICE or FIZZ and $w$ represents the weight assigned to the MBR score ($S_{sen}$). The setting with $w=0.0$ uses only the named scorer (consistency-only), which serves as a baseline. On the other hand, $w=1.0$ uses only the MBR score (consensus-only) and its \texttt{Scorer} is denoted as \texttt{MBR}. Conversely, a setting with $0 < w < 1$ indicates a linear combination of the named scorer and MBR score. 

\paragraph{Baselines}
We evaluate our proposed configurations (MBR-1.0, FENICE-0.75, and FIZZ-0.75) against three baselines: \texttt{Baseline} (standard decoding without MBR) and the consistency-only rerankers (FENICE-0.0 and FIZZ-0.0).

\subsection{Optimal Weight Combination ($w$)}
\label{subsec:optimal_weight}

\paragraph{Experiment Settings}
To leverage both \emph{consensus} and \emph{consistency}, our system combines MBR scores with reference-free metric scores. We investigated the optimal combination weight ($w$) using three distinct metrics: the factuality-focused FENICE \cite{Scire_EtAl-2024-FENICEFactualityEvaluation} and FIZZ \cite{Yang_EtAl-2024-FIZZFactualInconsistency}, and SimCLS \cite{liu_simcls_2021}, which balances factuality and quality. We evaluated weights $w \in \{0.0, 0.25, 0.5, 0.75, 1.0\}$, spanning from only using reference-free scoring (0.0) to only using MBR (1.0). Experiments utilized the candidate samples generated using epsilon sampling for PLMs and nucleus sampling for LLM (see Appendix \ref{appendix:optimal_pseudo}). We fixed the pseudo-references using 64 samples generated via epsilon sampling.

\begin{table}[t!]
\centering
\setlength{\tabcolsep}{2pt}
\resizebox{1.0\linewidth}{!}{
\begin{tabular}{@{}lcccccc@{}}
\toprule
& \multicolumn{2}{c}{\textbf{FENICE}} & \multicolumn{2}{c}{\textbf{FIZZ}} & \multicolumn{2}{c}{\textbf{SimCLS}} \\
\cmidrule(lr){2-3} \cmidrule(lr){4-5} \cmidrule(lr){6-7}
\textbf{$w$} & \textbf{CNNDM} & \textbf{XSUM} & \textbf{CNNDM} & \textbf{XSUM} & \textbf{CNNDM} & \textbf{XSUM} \\
\midrule
\textbf{0.00}      & 81.05          & 34.67          & 14.15          & 17.37          & 12.76          & 8.62           \\
\textbf{0.25}   & 62.29          & \textbf{77.83} & 36.88          & 13.23          & 24.51          & 15.60          \\
\textbf{0.50}   & 72.64          & 75.95          & 49.76          & 26.86          & 39.57          & 26.87          \\
\textbf{0.75}   & \textbf{81.05} & 77.52          & \textbf{71.08} & 55.03          & 63.51          & 55.06          \\
\textbf{1.00}      & 68.86          & 39.70          & 69.69          & \textbf{76.19} & \textbf{65.35} & \textbf{90.91} \\
\bottomrule
\end{tabular}
}
\caption{Average normalized of all metrics for CNN/DM and XSum datasets. The 'no mbr' and 'mbr only' settings are denoted by 0 and 1, respectively. Best scores for each reranker are in bold.}
\label{tab:weight_results}
\end{table}

\begin{table*}[ht]
\centering
\footnotesize
\setlength{\tabcolsep}{2pt}
\resizebox{1.0\textwidth}{!}{

\begin{tabular}{@{}ll | l >{\columncolor[gray]{0.9}}l l >{\columncolor[gray]{0.9}}l l |>{\columncolor[gray]{0.9}}l l >{\columncolor[gray]{0.9}}l l >{\columncolor[gray]{0.9}}l l >{\columncolor[gray]{0.9}}l  @{}}
\toprule
& & \multicolumn{5}{c|}{\textbf{Quality}} & \multicolumn{7}{c}{\textbf{Factuality}} \\
\cmidrule(lr){3-7} \cmidrule(lr){8-14}
\textbf{Setting} & \textbf{Reranker} & \multicolumn{1}{c}{\textbf{R1}} & \multicolumn{1}{c}{\textbf{R2}} & \multicolumn{1}{c}{\textbf{RL}} & \multicolumn{1}{c}{\textbf{BS}} & \multicolumn{1}{c|}{\textbf{MS}} & \multicolumn{1}{c}{\textbf{EM}} & \multicolumn{1}{c}{\textbf{CM}} & \multicolumn{1}{c}{\textbf{SM}} & \multicolumn{1}{c}{\textbf{UE}} & \multicolumn{1}{c}{\textbf{Fe}} & \multicolumn{1}{c}{\textbf{Fi}} & \multicolumn{1}{c}{\textbf{SC}} \\
\midrule

\multirow{8}{*}{epsilon}& Baseline & 39.96 & 16.88 & 33.38 & 65.52 & 58.03 & 4.46 & -6.62 & 1.63 & 81.37 & 95.52 & 39.36 & \underline{99.74} \\
& FENICE-0.0 & 40.52 & 17.46 & 34.06 & \underline{66.02} & 58.19 & 5.96 & -5.21 & 2.14 & \underline{85.18} & --- & \underline{55.83} & 99.71 \\
& FIZZ-0.0 & 40.54 & 17.81 & 34.23 & 65.91 & 58.17 & 5.18 & -5.59 & 2.39 & 84.65 & 98.39 & --- & 99.72 \\
\cmidrule(lr){2-14}
& FENICE-0.75 & \textbf{40.60}* & 17.47* & 34.11* & 65.86* & \textbf{\underline{58.25}}*\textdagger\textdaggerdbl & \textbf{10.44}*\textdagger\textdaggerdbl & \textbf{-3.82}*\textdagger\textdaggerdbl & 2.36* & 83.34* & --- & 52.44* & 99.70 \\
& FIZZ-0.75 & \textbf{\underline{40.67}}*\textdagger\textdaggerdbl & \textbf{\underline{17.87}}*\textdagger & \textbf{\underline{34.36}}*\textdagger\textdaggerdbl & 65.94* & \textbf{58.24}*\textdagger\textdaggerdbl & \textbf{7.68}*\textdagger\textdaggerdbl & \textbf{-4.46}*\textdagger\textdaggerdbl & \textbf{\underline{2.44}}* & 84.16* & \textbf{\underline{98.45}}*\textdaggerdbl & --- & 99.70 \\
& MBR-1.0 & 40.36* & 17.23* & 33.87* & 65.60* & 58.19* & \textbf{\underline{11.29}}*\textdagger\textdaggerdbl & \textbf{\underline{-3.78}}*\textdagger\textdaggerdbl & --- & 81.41 & 97.35* & 47.26* & 99.69 \\\cmidrule(lr){2-14}
& Oracle & \textit{51.85} & \textit{28.45} & \textit{45.15} & \textit{71.35} & \textit{61.29} & \textit{23.82} & \textit{-0.47} & \textit{35.02} & \textit{95.20} & \textit{99.80} & \textit{84.22} & \textit{99.89} \\
\midrule

\multirow{8}{*}{beam-sim}& Baseline & 36.37 & \underline{17.34} & \underline{31.64} & 64.78 & 56.79 & 12.26 & -5.67 & 2.24 & 89.86 & 97.02 & 68.44 & 97.63 \\
& FENICE-0.0 & 35.69 & 15.94 & 30.67 & 64.61 & 56.74 & \underline{14.56} & -5.49 & -0.13 & 89.54 & --- & \underline{68.89} & 97.65 \\
& FIZZ-0.0 & 35.60 & 16.27 & 30.78 & 64.37 & 56.65 & 12.80 & -5.91 & 0.45 & 88.94 & 97.66 & --- & 97.62 \\
\cmidrule(lr){2-14}
& FENICE-0.75 & \textbf{36.50}*\textdagger\textdaggerdbl & 16.63\textdagger\textdaggerdbl & 31.23\textdagger\textdaggerdbl & \textbf{\underline{64.97}}*\textdagger\textdaggerdbl & \textbf{\underline{56.90}}*\textdagger\textdaggerdbl & 13.37*\textdaggerdbl & \textbf{-4.54}*\textdagger\textdaggerdbl & \textbf{\underline{6.60}}*\textdagger\textdaggerdbl & \textbf{\underline{90.14}}*\textdagger\textdaggerdbl & --- & 65.85 & \textbf{97.71}*\textdagger\textdaggerdbl \\
& FIZZ-0.75 & 36.14\textdagger\textdaggerdbl & 16.65\textdagger\textdaggerdbl & 31.14\textdagger\textdaggerdbl & 64.68\textdagger\textdaggerdbl & 56.79\textdagger\textdaggerdbl & 13.19*\textdaggerdbl & \textbf{-4.99}*\textdagger\textdaggerdbl & \textbf{4.59}*\textdagger\textdaggerdbl & 89.48\textdaggerdbl & \textbf{\underline{97.80}}*\textdaggerdbl & --- & 97.65*\textdaggerdbl \\
& MBR-1.0 & \textbf{\underline{36.56}}*\textdagger\textdaggerdbl & 16.65\textdagger\textdaggerdbl & 31.14\textdagger\textdaggerdbl & \textbf{64.88}*\textdagger\textdaggerdbl & \textbf{56.87}*\textdagger\textdaggerdbl & 12.03 & \textbf{\underline{-4.44}}*\textdagger\textdaggerdbl & --- & 89.67\textdaggerdbl & 97.32* & 60.60 & \textbf{\underline{97.73}}*\textdagger\textdaggerdbl \\\cmidrule(lr){2-14}
& Oracle & \textit{46.68} & \textit{27.01} & \textit{41.80} & \textit{69.90} & \textit{59.54} & \textit{39.80} & \textit{-0.47} & \textit{23.46} & \textit{96.12} & \textit{99.86} & \textit{92.69} & \textit{98.62} \\
\midrule

\multirow{8}{*}{llm}& Baseline & 34.83 & 13.51 & 28.34 & 64.04 & 56.56 & 4.80 & -2.12 & 21.70 & 92.58 & 98.43 & 24.98 & 99.91 \\
& FENICE-0.0 & 35.15 & 13.77 & 28.66 & 64.20 & 56.64 & 5.53 & -2.05 & 21.19 & 92.63 & --- & \underline{31.12} & 99.90 \\
& FIZZ-0.0 & 35.26 & 13.80 & 28.74 & 64.23 & 56.66 & 5.13 & -2.27 & 20.92 & 92.55 & 98.89 & --- & 99.90 \\
\cmidrule(lr){2-14}
& FENICE-0.75 & \textbf{35.31}*\textdagger & \textbf{\underline{14.11}}*\textdagger\textdaggerdbl & 28.71* & \textbf{\underline{64.38}}*\textdagger\textdaggerdbl & 56.66*\textdagger & \textbf{5.95}*\textdagger\textdaggerdbl & \textbf{-1.60}*\textdagger\textdaggerdbl & \textbf{\underline{31.14}}*\textdagger\textdaggerdbl & \textbf{92.91}*\textdagger\textdaggerdbl & --- & 28.50* & 99.91*\textdagger\textdaggerdbl \\
& FIZZ-0.75 & \textbf{\underline{35.36}}*\textdagger\textdaggerdbl & \textbf{13.98}*\textdagger\textdaggerdbl & \textbf{\underline{28.80}}*\textdagger\textdaggerdbl & \textbf{64.34}*\textdagger\textdaggerdbl & \textbf{\underline{56.68}}*\textdagger\textdaggerdbl & 5.34*\textdaggerdbl & \textbf{-1.95}*\textdaggerdbl & \textbf{25.08}*\textdagger\textdaggerdbl & \textbf{92.70}*\textdaggerdbl & \textbf{\underline{98.90}}* & --- & 99.90\textdaggerdbl \\
& MBR-1.0 & 35.15* & \textbf{14.07}*\textdagger\textdaggerdbl & 28.52* & \textbf{64.36}*\textdagger\textdaggerdbl & 56.63* & \textbf{\underline{6.02}}*\textdagger\textdaggerdbl & \textbf{\underline{-1.51}}*\textdagger\textdaggerdbl & --- & \textbf{\underline{92.99}}*\textdagger\textdaggerdbl & 98.50 & 24.73 & \textbf{\underline{99.92}}*\textdagger\textdaggerdbl \\\cmidrule(lr){2-14}
& Oracle & \textit{40.88} & \textit{18.83} & \textit{34.05} & \textit{67.03} & \textit{57.93} & \textit{14.81} & \textit{-0.45} & \textit{49.41} & \textit{95.63} & \textit{99.81} & \textit{55.71} & \textit{99.95} \\

\bottomrule
\end{tabular}
}
\caption{Results for each metric on the CNN/DM dataset. \underline{Underline} indicates the highest scores for each metric and \textbf{bold} indicates better scores than all baselines. *, \textdagger, and \textdaggerdbl ~represent the statistical significance against Baseline, FENICE-0.0, and FIZZ-0.0, respectively (See \S\ref{subsec:exp_setting}). ``---'' indicates the skipped settings because the metrics used in the reranking and evaluation are identical. Each abbreviation represents the following metric, \textbf{R1} - ROUGE-1, \textbf{R2} - ROUGE-2, \textbf{RL} - ROUGE-L, \textbf{BS} - BERTScore, \textbf{MS} - MoverScore, \textbf{EM} - MENLI-Entailment, \textbf{CM} - MENLI-Contradiction, \textbf{SM} - MENLI-Summarization, \textbf{UE} - UniEval-Overall, \textbf{Fe} - FENICE, \textbf{Fi} - FIZZ, and \textbf{SC} - SimCLS.}
\label{tab:cnndm_results}
\end{table*}

\begin{table*}[th]
\centering
\footnotesize %
\setlength{\tabcolsep}{2pt} %
\resizebox{1.0\textwidth}{!}{

\begin{tabular}{@{}ll | l >{\columncolor[gray]{0.9}}l l >{\columncolor[gray]{0.9}}l l |>{\columncolor[gray]{0.9}}l l >{\columncolor[gray]{0.9}}l l >{\columncolor[gray]{0.9}}l l >{\columncolor[gray]{0.9}}l  @{}}
\toprule
& & \multicolumn{5}{c|}{\textbf{Quality}} & \multicolumn{7}{c}{\textbf{Factuality}} \\
\cmidrule(lr){3-7} \cmidrule(lr){8-14}
\textbf{Setting} & \textbf{Reranker} & \multicolumn{1}{c}{\textbf{R1}} & \multicolumn{1}{c}{\textbf{R2}} & \multicolumn{1}{c}{\textbf{RL}} & \multicolumn{1}{c}{\textbf{BS}} & \multicolumn{1}{c|}{\textbf{MS}} & \multicolumn{1}{c}{\textbf{EM}} & \multicolumn{1}{c}{\textbf{CM}} & \multicolumn{1}{c}{\textbf{SM}} & \multicolumn{1}{c}{\textbf{UE}} & \multicolumn{1}{c}{\textbf{Fe}} & \multicolumn{1}{c}{\textbf{Fi}} & \multicolumn{1}{c}{\textbf{SC}} \\
\midrule

\multirow{8}{*}{epsilon}& Baseline & 38.27 & 15.91 & 30.60 & 75.20 & 59.38 & 9.50 & -31.15 & -23.49 & 85.08 & 45.67 & 16.91 & \underline{98.50} \\
& FENICE-0.0 & 38.67 & 16.29 & 30.96 & 75.47 & 59.44 & 17.41 & -24.20 & -17.95 & \underline{88.13} & --- & \underline{28.84} & 98.41 \\
& FIZZ-0.0 & 38.12 & 15.85 & 30.56 & 75.14 & 59.28 & 14.50 & -27.22 & -20.10 & 87.12 & 64.70 & --- & 98.39 \\
\cmidrule(lr){2-14}
& FENICE-0.75 & \textbf{\underline{38.68}}*\textdaggerdbl & \textbf{\underline{16.44}}*\textdagger\textdaggerdbl & \textbf{\underline{31.18}}*\textdagger\textdaggerdbl & \textbf{\underline{75.66}}*\textdagger\textdaggerdbl & \textbf{\underline{59.45}}*\textdaggerdbl & \textbf{27.04}*\textdagger\textdaggerdbl & \textbf{-20.36}*\textdagger\textdaggerdbl & \textbf{\underline{-16.40}}*\textdagger\textdaggerdbl & 88.01*\textdaggerdbl & --- & 27.79* & 98.32 \\
& FIZZ-0.75 & 38.41\textdaggerdbl & 16.17*\textdaggerdbl & 30.91*\textdaggerdbl & 75.41*\textdaggerdbl & 59.37\textdaggerdbl & \textbf{21.03}*\textdagger\textdaggerdbl & \textbf{-23.57}*\textdagger\textdaggerdbl & \underline{-18.07}*\textdaggerdbl & 87.63*\textdaggerdbl & \textbf{\underline{66.68}}*\textdaggerdbl & --- & 98.34 \\
& MBR-1.0 & 38.57*\textdaggerdbl & \textbf{16.38}*\textdaggerdbl & \textbf{31.11}*\textdagger\textdaggerdbl & \textbf{75.59}*\textdagger\textdaggerdbl & 59.39\textdaggerdbl & \textbf{\underline{28.20}}*\textdagger\textdaggerdbl & \textbf{\underline{-18.07}}*\textdagger\textdaggerdbl & --- & 87.53*\textdaggerdbl & 63.02* & 26.21* & 98.31 \\
\cmidrule(lr){2-14}
& Oracle & \textit{53.82} & \textit{31.35} & \textit{46.79} & \textit{81.63} & \textit{63.70} & \textit{43.60} & \textit{-5.32} & \textit{17.78} & \textit{94.11} & \textit{86.48} & \textit{51.10} & \textit{99.21} \\
\midrule

\multirow{8}{*}{beam-sim}& Baseline & \underline{41.78} & \underline{20.21} & \underline{34.64} & \underline{75.88} & \underline{59.87} & 16.62 & -27.06 & -18.88 & 86.26 & 54.74 & 25.05 & 97.59 \\
& FENICE-0.0 & 37.96 & 16.21 & 30.04 & 74.25 & 58.78 & \underline{19.84} & -22.05 & -16.06 & 87.14 & --- & \underline{33.23} & 97.60 \\
& FIZZ-0.0 & 37.27 & 15.68 & 29.53 & 73.74 & 58.57 & 16.79 & -24.40 & -17.94 & 85.92 & 65.08 & --- & 97.54 \\
\cmidrule(lr){2-14}
& FENICE-0.75 & 38.48\textdagger\textdaggerdbl & 16.41\textdagger\textdaggerdbl & 30.23\textdagger\textdaggerdbl & 74.39\textdagger\textdaggerdbl & 59.02\textdagger\textdaggerdbl & 14.67 & \textbf{-19.55}*\textdagger\textdaggerdbl & \textbf{\underline{-8.17}}*\textdagger\textdaggerdbl & \textbf{\underline{87.22}}*\textdaggerdbl & --- & 27.51* & \textbf{97.69}*\textdagger\textdaggerdbl \\
& FIZZ-0.75 & 37.64\textdaggerdbl & 15.92\textdaggerdbl & 29.77\textdaggerdbl & 73.90\textdaggerdbl & 58.69\textdaggerdbl & 16.76 & -22.16*\textdaggerdbl & \textbf{-13.80}*\textdagger\textdaggerdbl & 86.35\textdaggerdbl & \textbf{\underline{65.54}}*\textdaggerdbl & --- & 97.57\textdaggerdbl \\
& MBR-1.0 & 38.13\textdagger\textdaggerdbl & 16.02\textdaggerdbl & 29.67 & 74.03\textdaggerdbl & 58.89\textdagger\textdaggerdbl & 11.86 & \textbf{\underline{-18.49}}*\textdagger\textdaggerdbl & --- & 86.17\textdaggerdbl & 55.52* & 22.90 & \textbf{\underline{97.72}}*\textdagger\textdaggerdbl \\
\cmidrule(lr){2-14}
& Oracle & \textit{54.10} & \textit{32.27} & \textit{47.60} & \textit{81.15} & \textit{63.31} & \textit{49.79} & \textit{-5.26} & \textit{13.82} & \textit{93.72} & \textit{87.59} & \textit{58.15} & \textit{98.65} \\
\midrule

\multirow{8}{*}{llm}& Baseline & 19.03 & 5.11 & \underline{13.19} & 61.83 & 53.73 & 0.84 & -6.02 & 8.72 & 93.08 & 88.04 & 22.11 & 99.89 \\
& FENICE-0.0 & 19.08 & 5.05 & \underline{13.19} & 61.83 & 53.72 & 0.82 & -5.31 & 9.37 & \underline{93.24} & --- & \underline{27.38} & 99.89 \\
& FIZZ-0.0 & 19.07 & 5.03 & \underline{13.19} & 61.84 & 53.72 & 0.78 & -5.63 & 8.46 & 92.99 & \underline{90.41} & --- & 99.89 \\
\cmidrule(lr){2-14}
& FENICE-0.75 & \textbf{\underline{19.09}} & \textbf{5.16}\textdagger\textdaggerdbl & 13.13 & \textbf{\underline{61.91}}*\textdagger\textdaggerdbl & \textbf{\underline{53.74}}\textdagger\textdaggerdbl & \textbf{1.42}*\textdagger\textdaggerdbl & \textbf{-3.38}*\textdagger\textdaggerdbl & \textbf{\underline{19.09}}*\textdagger\textdaggerdbl & 93.22*\textdaggerdbl & --- & 24.35* & 99.89*\textdagger\textdaggerdbl \\
& FIZZ-0.75 & \textbf{\underline{19.09}} & 5.06\textdaggerdbl & 13.18 & \textbf{61.87}\textdaggerdbl & 53.73 & \textbf{0.86}\textdaggerdbl & \textbf{-4.44}*\textdagger\textdaggerdbl & \textbf{13.40}*\textdagger\textdaggerdbl & 93.04 & 90.36* & --- & 99.89\textdaggerdbl \\
& MBR-1.0 & 19.03 & \textbf{\underline{5.17}}*\textdagger\textdaggerdbl & 13.05 & \textbf{61.90}*\textdagger\textdaggerdbl & \textbf{\underline{53.74}} & \textbf{\underline{1.72}}*\textdagger\textdaggerdbl & \textbf{\underline{-2.91}}*\textdagger\textdaggerdbl & --- & 93.14\textdaggerdbl & 88.42* & 21.42 & \textbf{\underline{99.90}}*\textdagger\textdaggerdbl \\
\cmidrule(lr){2-14}
& Oracle & \textit{23.69} & \textit{8.33} & \textit{17.04} & \textit{64.43} & \textit{54.63} & \textit{4.06} & \textit{-0.82} & \textit{32.85} & \textit{95.79} & \textit{96.49} & \textit{50.80} & \textit{99.93} \\

\bottomrule
\end{tabular}
}
\caption{Results for each metric on the XSum dataset. \underline{Underline} indicates the highest scores for each metric and \textbf{bold} indicates better scores than all baselines. *, \textdagger, and \textdaggerdbl ~represent the statistical significance against Baseline, FENICE-0.0, and FIZZ-0.0, respectively (See \S\ref{subsec:exp_setting}). ``---'' indicates the skipped settings because the metrics used in the reranking and evaluation are identical. The abbreviations are the same as those in Table \ref{tab:cnndm_results}.}
\label{tab:xsum_results}
\end{table*}

\paragraph{Evaluation}
We evaluated the top-ranked summary selected by each weight and reranker combination using the two groups of metrics detailed in Table \ref{tab:eval_metrics}, excluding MoverScore, UniEval, and SimCLS. We averaged the scores across all PLMs, grouped by the weight and reranker. To ensure comparability between metrics with different scales, we applied MinMax normalization to the aggregated scores. Finally, we restructured the data into a long format and categorized the metrics according to our defined groups. 

\paragraph{Results}
The results are presented in Table~\ref{tab:weight_results}. The optimal combination weight ($w$) varies significantly by dataset and reference-free metric. On CNN/DM, the optimal weight for both FENICE and FIZZ is \textbf{0.75}, whereas on XSum, FENICE peaks at 0.25. Conversely, SimCLS performed best with a weight of 1.0 on both datasets, indicating that the reference-free score negatively impacted performance compared to using MBR alone. To maintain a unified setting, we selected the configuration that performed best across both datasets. We selected $w = 0.75$ as the default, as it is optimal for CNN/DM and competitive for XSum. Given the negative contribution of SimCLS, we exclude it from the final system comparison.

\subsection{Experiment Results}

The main results are shown in Tables~\ref{tab:cnndm_results} and \ref{tab:xsum_results} for the CNN/DM and XSum datasets, respectively. For settings with multiple models, i.e., \texttt{epsilon} and \texttt{beam-sim}, we averaged the results from all models. The \texttt{Baseline} refers to the non-reranked summary generated by each setting's decoding method. 

The tables show that our methods \textbf{significantly dominate most of the factuality metrics} when compared with the baselines as shown by our methods achieving the highest scores and are often statistically significant against all three baselines. These results show that our method of using pseudo-references as an alternative signal works in improving factuality of a summary. 

On the CNN/DM dataset, the \texttt{FENICE-0.75} configuration significantly improved the FIZZ (Fi) score from 39.36 (Baseline) to 52.44 in the \texttt{epsilon} setting. This trend is even more pronounced on the XSum dataset, where the same configuration raised the FIZZ (Fi) score from 16.91 to 27.79. Similarly, the configuration significantly improved most MENLI scores, except for MENLI-Entailment (EM) score in \texttt{beam-sim} setting. MENLI-Entailment (EM) has the highest improvement in both datasets, where the score increased from 4.46 to 10.44 in CNN/DM dataset and -31.15 to -20.36 in XSum dataset. Given that XSum is characterized by a high frequency of hallucinations due to its abstractiveness \cite{maynez-etal-2020-faithfulness}, our findings confirm that consensus-based optimization can successfully filter out these errors.

Furthermore, this enhancement in factuality does not compromise the summary quality metrics, the scores remained competitive with, or significantly better than, the baselines across most settings. For CNN/DM, the improvements are mainly by the combination methods (FENICE-0.75 and FIZZ-0.75). For XSum, our methods' improvements are small but still significant. Unfortunately, for \texttt{beam-sim}, \texttt{Baseline} is the best for all quality metrics. However, considering that MENLI scores (EM, CM, and SM) have improvements when evaluating against the ``gold'' references, the summaries chosen by our method do not diverge from them.

To establish a theoretical upper bound within the generated candidate pool, we calculated \texttt{Oracle} scores for each metric by selecting the candidate from the pool that maximizes the respective metric against the ``gold'' references or the source document. While the gap is small for some metrics, most show a big difference, with the \texttt{Oracle} scores often being more than double of our method's best scores. This holds true for both the CNN/DM and XSum datasets. This gap indicates that, despite the gains over the baselines, there is still substantial room for improvement in identifying the best possible candidate summary from a given pool. Our additional results (Appendix \ref{appendix:additional_test_res}) further reinforce the robustness of our method, demonstrating its effectiveness regardless of the sampling strategy or generation model.

Our preliminary results on weight sensitivity (see \S\ref{subsec:optimal_weight}) indicate that balancing the weight $w$ between MBR and reference-free metric is necessary to get the best overall performance on both dataset. For the CNN/DM dataset, FENICE reaches its peak performance at both $w=0$ and $w=0.75$, whereas FIZZ and SimCLS show a more linear improvement as $w$ increases, peaking at 0.75 and 1, respectively. Conversely, on the XSum dataset, all three models exhibit higher sensitivity to the weight parameter, with peak performance generally occurring at higher values of $w$, suggesting that MBR are particularly crucial for this dataset.

In this study, we treat candidates $\mathcal{Y}$ and pseudo-references $\mathcal{R}$ as different sets. This raises a question whether the performance gains stems from quality disparities between the candidate pool and the pseudo-reference pool. Our preliminary study using multiple decoding strategies to generate the candidates against epsilon-generated pseudo-references confirms that the improvements achieved by ConSUM are not derived from utilizing superior references, but from the consensus mechanism itself. More detail about our investigation is in Appendix \ref{appendix:optimal_pseudo}.

Lastly, we reported the average length of the generated summaries (see Appendix \ref{appendix:length_summary}) and some sample comparison summaries  (see Appendix \ref{appendix:example_summaries}) to see the result qualitatively. Our findings show that PLMs and LLM have different length of summaries. LLM generated more than 100 tokens for both datasets, while PLMs generated around 80 and 30 tokens for CNN/DM and XSum, respectively. With limited length, the main focus of the PLM-generated summaries changed w.r.t source. On the other hand, LLM-generated summaries are not limited by the prompt, hence the changes include additional point, and correction of factual error.

\begin{figure}[t]
    \centering
    \includegraphics[width=1.0\linewidth]{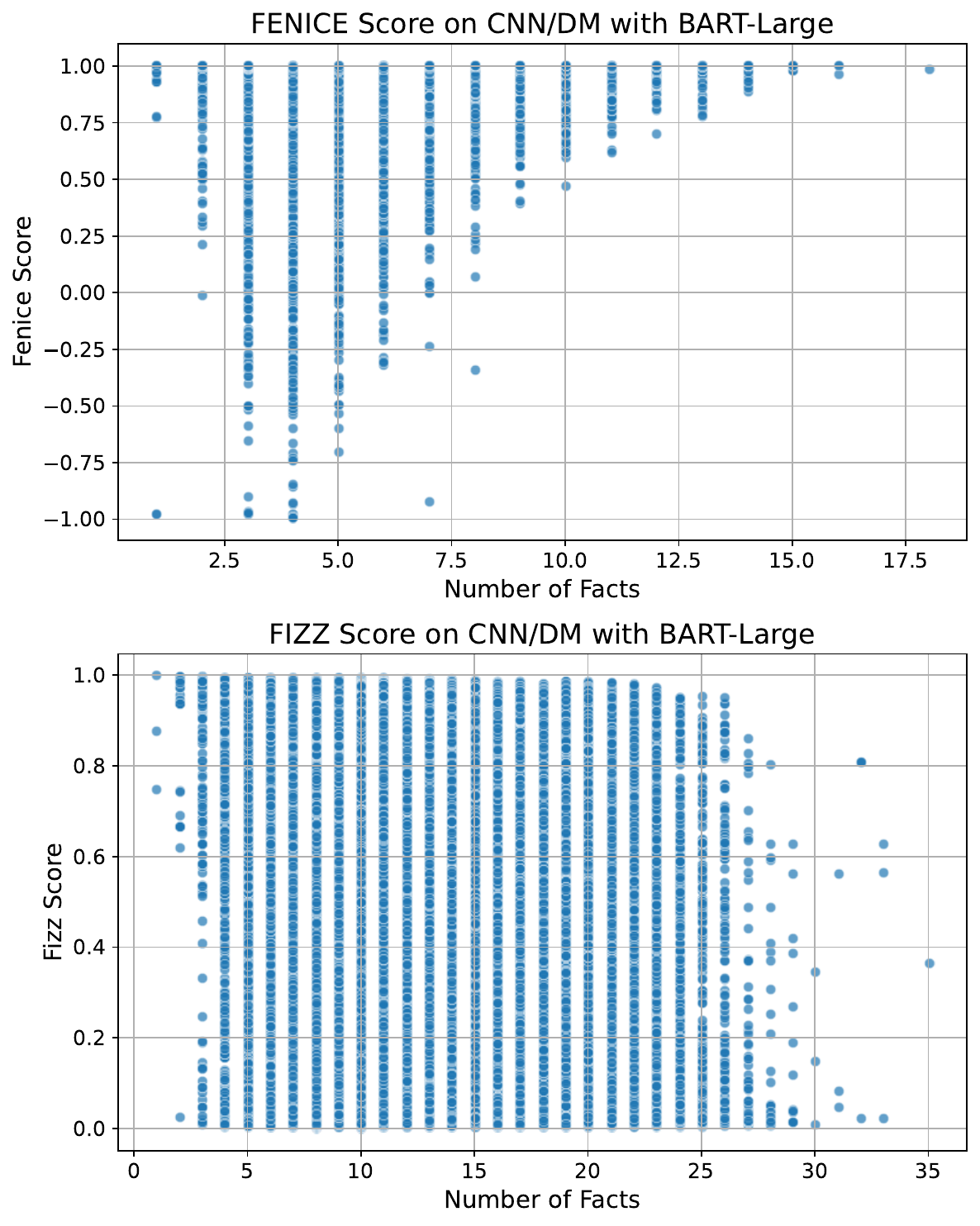}
    \caption{Correlation between the number of facts extracted using each reference-free metric and the respective factuality score for BART-generated summaries on CNN/DM. \textbf{Top:} The correlation for FENICE. \textbf{Bottom:} The correlation for FIZZ.}
    \label{fig:corr_num_facts}
\end{figure}

\subsection{Effect of the Amount of Extracted Facts}

We investigated the different behaviors between FENICE and FIZZ by analyzing the relationship between the number of Atomic Content Units (ACUs) extracted and the final factuality score. Figure~\ref{fig:corr_num_facts} shows a representative plot for this analysis, using summaries from the BART model on the CNN/DM dataset, since we observed similar patterns across all settings. 

Our analysis revealed that there is little to no correlation between the number of ACUs in a summary and the final factuality score. A summary with more ``facts'' is not necessarily rated as more factual. In addition, FENICE and FIZZ operate at vastly different granularities. FENICE typically extracts and evaluates a small number of ACUs (usually 3-6) to determine the score. In contrast, FIZZ consistently decomposes a summary into a much larger number of ACUs. This difference in granularity highlights that, although they share a conceptual foundation, their mechanisms for assessing factuality are fundamentally different.

\subsection{Correlation between Rerankers and Metrics}

To better understand what drives the performance improvements, we analyzed the correlation between $S_{fin}$ scores and each metric's scores. In addition, previous studies~\cite{muller_understanding_2021, kovacs_mitigating_2024} have shown that reranking resulted in bias toward the optimized metric.
One example bias is summary length, thus we also included the following factors to our correlation analysis: The number of facts extracted by FENICE and FIZZ; The length of candidate summaries and source documents.

\begin{figure}[t]
    \centering
    \includegraphics[width=1.0\linewidth]{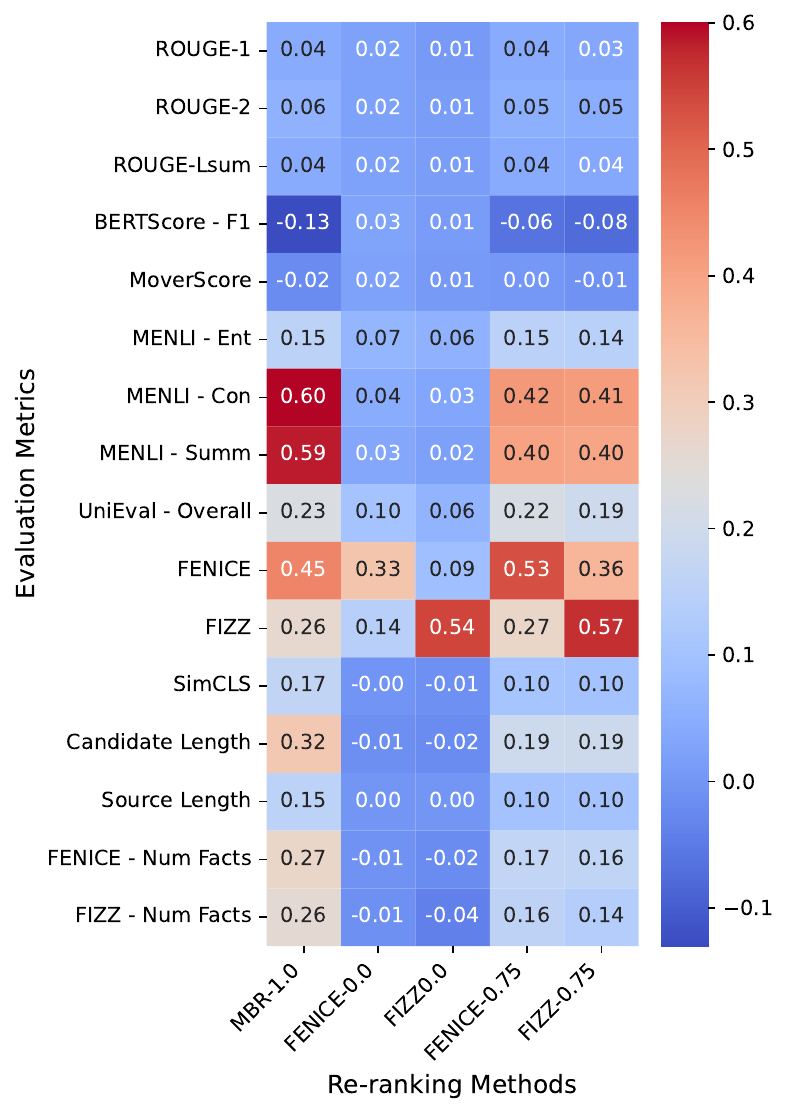}
    \caption{Correlation between rerankers and evaluation metrics. Higher numbers represent positive correlations, lower numbers represent negative correlations, and 0.0 represents no correlation.}
    \label{fig:corr_metrics}
\end{figure}

Instead of word counts, we measured the length using the models' own tokenizers, as this reflects the input the model actually receives. The full correlation matrix is shown in Figure~\ref{fig:corr_metrics}.
Surprisingly, the MENLI score, which we used as the MBR utility function, shows a slight positive correlation with the summary length. This suggests that MENLI, as an NLI-based metric, may have a tendency to favor slightly longer summaries. As expected, the MBR score is highly correlated with the MENLI utility function, and the FENICE and FIZZ scores are highly correlated with their own combined scores. Interestingly, even though our final score is dominated by the MBR component ($w=0.75$), combining it with a reference-free metric increases the correlation with that metric. For example, the correlation between ConSUM (FENICE-0.75) to FENICE metric is higher compared to MBR (MBR-1.0) and FENICE reranker (FENICE-0.0) alone. This suggests that the combination helps to reduce MENLI's inherent self-correlation bias while boosting the influence of the desired reference-free metric.

\section{Human Evaluation}
To validate our findings beyond automatic metrics, we conducted a human evaluation to directly assess which system's outputs are preferred by people. We randomly sampled 50 source documents from the CNN/DM test set. For each source document, we presented annotators with 5 summaries to evaluate: the human-written/gold reference; the top-ranked summary from our best systems: FENICE-0.75, FIZZ-0.75, and MBR-1.0; and the top-ranked summary from the Baseline (w/o using MBR).  
All system-generated summaries were sourced from the BART model using the \texttt{beam-sim} setting, as this configuration yielded the best average performance in our automatic metric tests. Further discussion is presented in Appendix \ref{appendix:additional_test_res}.

We recruited annotators on Amazon Mechanical Turk, with 3 unique annotators evaluating each set of summaries. They were assigned two tasks: (1) Annotators rated each of the 5 summaries on a 1--5 Likert scale across three criteria: Informativeness, Factuality, and Fluency (1 = ``Strongly Disagree'', 5 = ``Strongly Agree''). The definition of each criteria is in Appendix \ref{appendix:aspect_def}. (2) Annotators ranked the same 5 summaries from best (1) to worst (5). We explicitly permitted them to assign duplicate ranks (ties) if they judged two or more summaries to be of equal quality. To ensure high-quality annotations, we recruited only MTurk workers who met the following criteria: HIT Approval Rate $\ge$ 90\% and Number of HITs Approved $\ge$ 50. Details is in Appendix \ref{appendix:mturk_page}.

To validate our human evaluation results, we first measured the Inter-Annotator Agreement (IAA) on the ranking task. We calculated the \textit{Kendall's Tau} correlation \cite{kendall1948rank} between every pair of annotators for each sample, took the maximum of these pairwise scores, and then averaged the results across all 50 samples. This yielded an average correlation score of 0.703, indicating a strong level of agreement among the annotators and confirming the reliability of our findings.

The results for the aspect-based ratings (Informativeness, Factuality, and Fluency) are shown in Table~\ref{tab:human_eval_aspect}. The FENICE-0.75 system achieved the highest overall average scores, making it the most preferred system when considering all aspects together. The aspect-level scores reveal each system's strengths and weaknesses. The Baseline system, while very competitive, also showed a slight weakness in informativeness compared to our method. Interestingly, the gold reference was rated poorly on both informativeness and fluency. We believe that this aligns with existing result \cite{Zhang_EtAl-2024-BenchmarkingLargeLanguage} that denote the low quality of gold references for CNN/DM. 

\begin{figure}[t]
    \centering
    \includegraphics[width=0.95\linewidth]{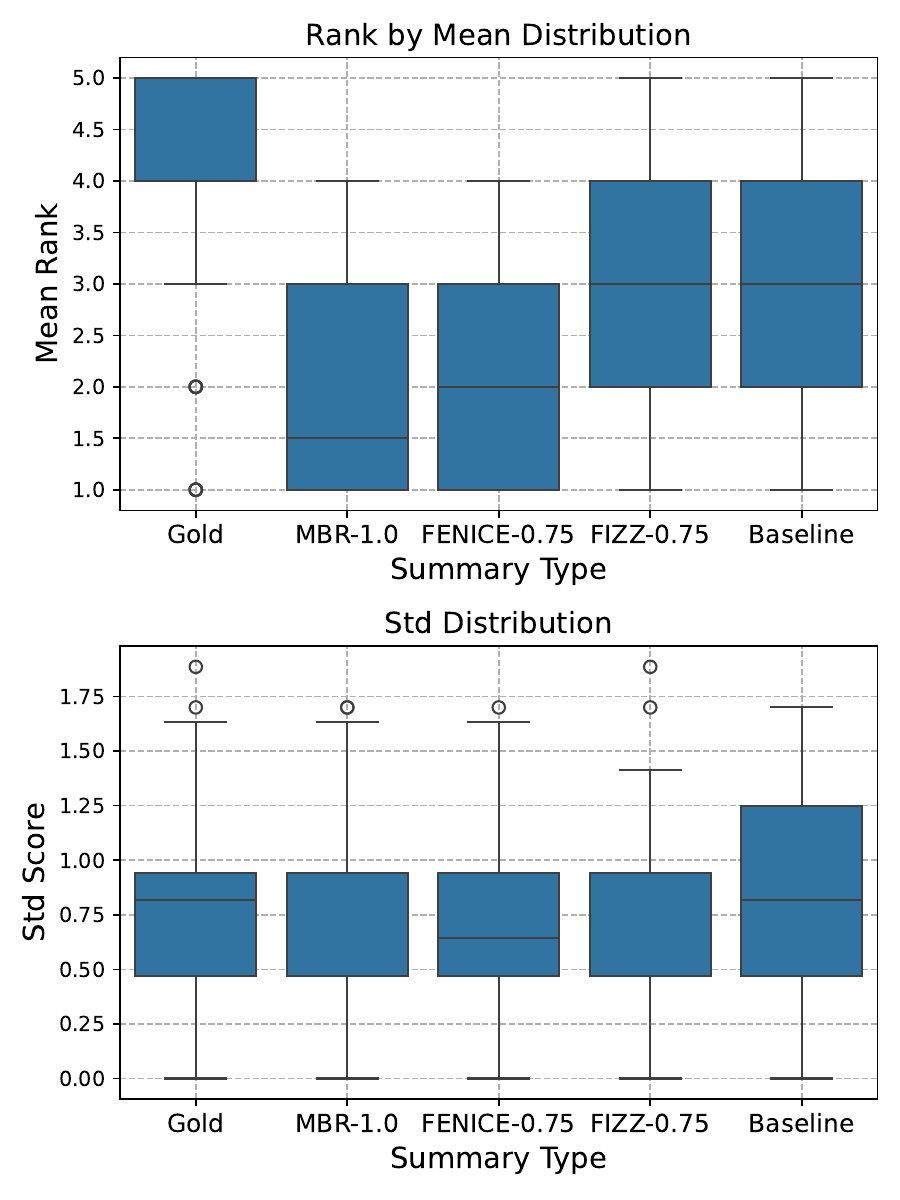}
    \caption{Rank distribution of each system. \textbf{Top:} The rank distribution based on the average ranking between annotators for each system. \textbf{Bottom:} The standard deviation of each annotator for each system.}
    \label{fig:human_eval_rank}
\end{figure}

The ranking results, visualized in Figure~\ref{fig:human_eval_rank}, provide a direct comparison of which summaries annotators preferred overall. 
FENICE-0.75 and MBR-1.0 emerge as the clear winners by having the lowest average rank (lower is better) and a low standard deviation, showing strong consensus between the annotators.
This shows that annotators not only ranked these systems' summaries higher but were also more consistent in their judgment with each other.
The Baseline has similar average rank with FIZZ-0.75, however with bigger standard deviation, showing that the annotators have different ranking for Baseline-generated summaries.
Consistent with the aspect-based ratings, the gold reference was ranked as the worst overall.

\begin{table}[!t]
\centering
\setlength{\tabcolsep}{2pt}
\begin{tabular}{@{}lccc|c@{}}
\toprule
\textbf{System} & \textbf{Fact.} & \textbf{Info.} & \textbf{Fluency} & \textbf{Overall} \\
\midrule
Gold        & 4.57 & 3.32 & 3.87 & 3.92 \\
MBR-1.0     & 4.74 & \textbf{4.31} & 4.66 & 4.57 \\
FENICE-0.75 & \textbf{4.87} & 4.30 & \textbf{4.73} & \textbf{4.63} \\
FIZZ-0.75   & 4.77 & 3.99 & 4.67 & 4.48 \\
Baseline    & 4.79 & 4.19 & 4.71 & 4.56 \\
\bottomrule
\end{tabular}
\caption{Human evaluation results for different systems based on the aspects. The overall score is calculated by averaging all aspects of each system. \textbf{Fact.} - Factuality and \textbf{Info.} - Informativeness.}
\label{tab:human_eval_aspect}
\end{table}

\section{Conclusion}
In this work, we introduced \textbf{ConSUM}, a novel reranking framework designed to address a key limitation in summary evaluation: the reliance on either a source document or reference summaries alone. We hypothesized that the best summary is one that is not only faithful to the source but also represents the consensus of the model's own distribution.
To validate this, we conducted an extensive set of experiments across multiple datasets, sampling methods, and system configurations. Our findings showed that ConSUM improves summary factuality across the board. Notably, in several settings, it achieved this without the common trade-off of sacrificing summary quality.

Finally, our human evaluation results confirmed these quantitative findings. Annotators consistently preferred the summaries produced by ConSUM over strong baselines, both in direct ranking and in aspect-based ratings. This work demonstrated that by combining source-based factuality with model-based consensus, we can generate summaries that are not only more factually reliable but also preferred by human readers.

\section*{Limitations}
Although our method, ConSUM, showed promising results, this work has several limitations. 
Our exploration of MBR settings focused only on the optimal number of candidates and pseudo-references, leaving other factors like the choice of the utility function or the pseudo-reference generation strategies unexplored. This was primarily due to MBR's $O(n^2)$ computational complexity, where increasing the number of candidates or references exponentially increases the required time and resources. Some previous studies \cite{Trabelsi_EtAl-2024-EfficientMinimumBayes, Cheng_Vlachos-2023-FasterMinimumBayes, natsumi-etal-2025-agreement} have proposed techniques to improve computational efficiency in MBR decoding, and these optimizations can be combined with our approach.
We also encountered computational challenges with the FENICE and FIZZ metrics, which struggled with large-scale processing, hindering a more thorough parameter exploration for them. Finally, our experiments were confined to two English news datasets (CNN/DM and XSum). The inconsistent results observed between these two datasets suggest that the effectiveness of our method may vary by domain, highlighting the need for future research on a more diverse range of text types and languages.

\section*{Ethical Considerations}

This study fully complies with the ACL Ethics Policy and addresses all relevant items in the Responsible Research Checklist. All resources used in this work are publicly available and properly licensed, with no concerns regarding licensing. The study does not involve or produce any harmful content.
For annotation, we ensured that all rights to the artifacts were formally transferred to the authors through explicit agreements. Annotators were recruited via a crowdsourcing platform, where compensation terms were clearly stated and agreed upon in advance. All annotators were fairly compensated for their contributions.
Although AI assistants were utilized for minor writing support, such as rephrasing and spell-checking, all original content was manually created by the authors.
Given these points, we affirm that this work raises no ethical concerns.

\section*{Acknowledgment}

This work was supported by JSPS KAKENHI Grant Number JP23H03458.

\bibliography{custom}

\appendix

\section{Preliminary}
\label{appendix:preliminary}
In contrast to NMT, the application and impact of MBR decoding in the text summarization task remain relatively unexplored. This research gap necessitates a preliminary study to identify the optimal hyperparameter configuration for our ConSUM method. Specifically, our study aims to determine the following: (1) the optimal number of candidates; (2) the optimal number of pseudo-references; (3) the optimal weight combination scores. 

\begin{figure}[t]
    \centering
    \includegraphics[width=1\linewidth]{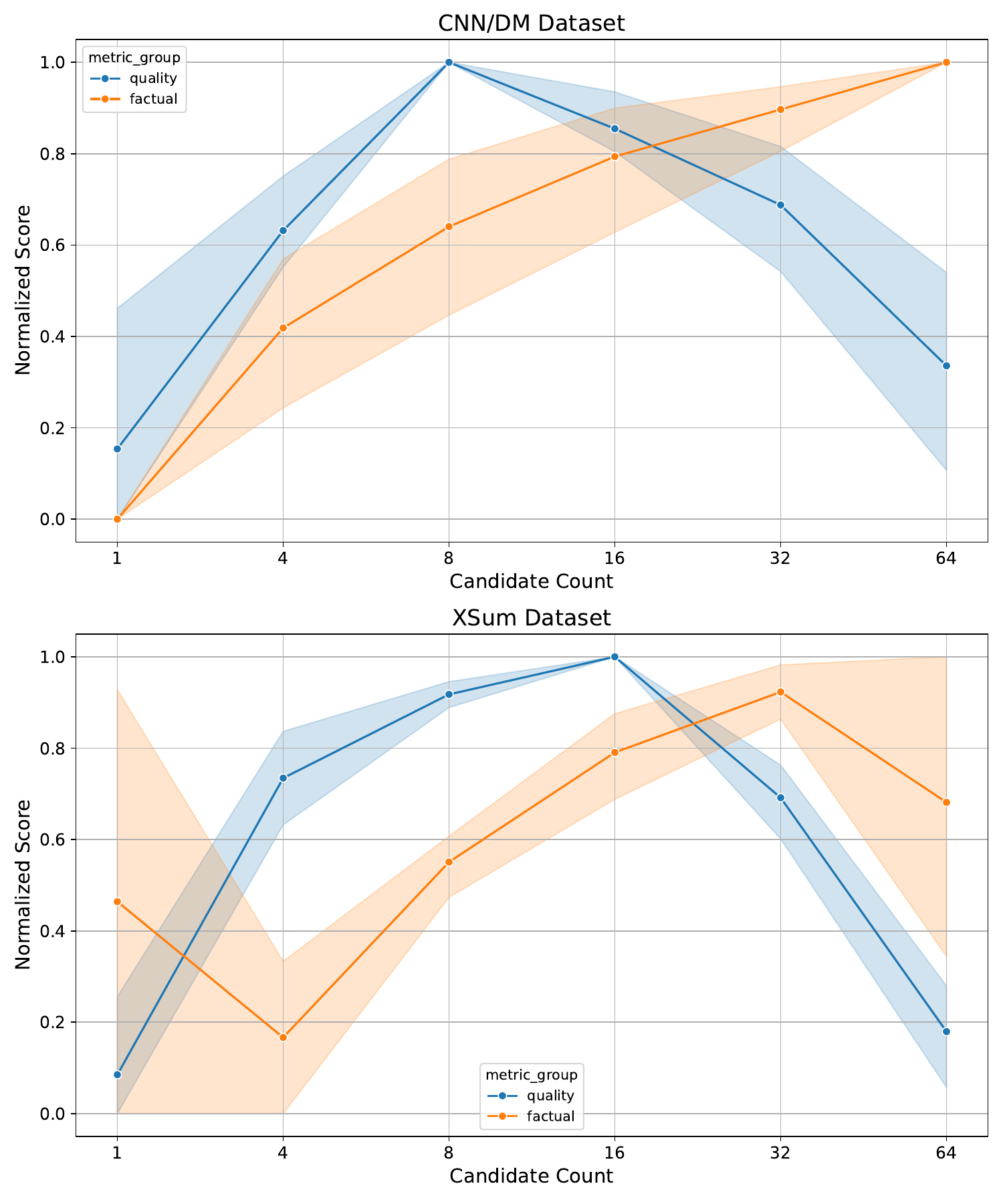}
    \caption{Average of normalized evaluation metrics based on the candidate summaries count on each dataset. \textbf{Top:} The average result on CNN/DM dataset. \textbf{Bottom:} The average result on XSum dataset. The results show the tendency of quality metrics are the same but different for factual metrics.}
    \label{fig:optimal_cand}
\end{figure}

\subsection{Optimal Number of Candidates}
\label{appendix:optimal_cand}

\paragraph{Experiment Settings}
Motivated by NMT findings linking performance to hypothesis count \cite{kovacs_mitigating_2024}, we investigate the effect of candidate set size ($|\mathcal{Y}|$) in MBR decoding. We generated an initial count of 64 samples per document using epsilon sampling ($\epsilon=0.02$) \cite{Freitag_EtAl-2023-EpsilonSamplingRocks} across three PLMs: BART, PEGASUS, and T5-Large. From this pool, we evaluated subsets of sizes {1, 4, 8, 16, 32, 64}, setting the candidate set as the pseudo-reference set ($\mathcal{Y}=\mathcal{R}$) in each configuration.

\paragraph{Evaluation}
We evaluated the top-ranked summary from MBR decoding using the two groups of metrics detailed in Table \ref{tab:eval_metrics}, excluding MoverScore, UniEval, and SimCLS. We averaged the scores across all PLMs, grouped by candidate count. To ensure comparability between metrics with different scales, we applied Min-Max normalization to the aggregated scores. Finally, we restructured the data into a long format and categorized the metrics according to our defined groups.

\paragraph{Results}
The results are presented in Figure \ref{fig:optimal_cand}. In terms of quality metrics, both datasets exhibit a similar trend: performance peaks at 8 and 16 candidates for CNN/DM and XSum, respectively, before gradually declining. However, trends in factuality metrics diverge. On CNN/DM, factuality scores generally improve as the candidate count grows. Conversely, on XSum, scores rise between 4 and 32 candidates but drop thereafter. This disparity likely stems from dataset characteristics: CNN/DM’s extractive nature benefits from larger consensus pools, whereas XSum’s highly abstractive summaries may accumulate noise with excessive candidates. Balancing these optima, we select \textbf{16 candidates} as the default for all subsequent experiments.

\subsection{Optimal Number of Pseudo-References}
\label{appendix:optimal_pseudo}

\paragraph{Experiment Settings}
Fixing the candidate count at the optimal $|\mathcal{Y}|=16$ (Appendix \ref{appendix:optimal_cand}), we investigated the impact of the pseudo-reference set size ($|\mathcal{R}|$). We generated 16 candidate summaries using three strategies: Epsilon Sampling ($\epsilon=0.02$) and Diverse Beam Search (DBS) for PLMs, and Nucleus Sampling for Llama-3. For DBS, we evaluated configurations from prior work~\cite{Dixit_EtAl-2023-ImprovingFactualityAbstractive, liu_simcls_2021} alongside our custom settings (detailed in Table~\ref{tab:sampling-setting}). We compared two reference scenarios: using the candidate set itself as the reference ($\mathcal{Y}=\mathcal{R}$), and utilizing the larger pre-generated pools from Appendix \ref{appendix:optimal_cand} as an external reference set.

\begin{table*}[]
    \centering
    \small
    \setlength{\tabcolsep}{4pt}
    \begin{tabular}{cccccc}
        \toprule
        \textbf{Name} & \textbf{Sampling Used} & \textbf{Models} & \textbf{Hyperparameter} & \textbf{Value Changed} & \textbf{Note} \\
        \midrule
        
        epsilon & Epsilon & BART 
            & epsilon cutoff & 0.02 & -- \\
        & & PEGASUS & do sample & True & \\
        & & T5-Large & num beams & 1 & \\
        & & & num return sequences & 16 & \\
        \midrule
        
        beam-div & DBS & BART 
            & do sample & False & \cite{Dixit_EtAl-2023-ImprovingFactualityAbstractive} \\
        & & PEGASUS & num beams & 4, PEGASUS=8 & \\
        & & T5-Large & num beam groups & 1 & \\
        & & & num return sequences & num beams & \\
        & & & diversity penalty & 0.0 & \\
        \midrule
        
        beam-dbl & DBS & BART
            & do sample & False & -- \\
        & & PEGASUS & num beams & 8, PEGASUS=16 & \\
        & & T5-Large & num beam groups & 2 & \\
        & & & num return sequences & num beams & \\
        & & & diversity penalty & 0.5 & \\
        \midrule

        beam-sim & DBS & BART
            & do sample & False & \cite{liu_simcls_2021} \\
        & & PEGASUS & num beams & 16 & \\
        & & T5-Large & num beam groups & 16 & \\
        & & & num return sequences & num beams & \\
        & & & diversity penalty & 1.0 & \\
        \midrule

        llm & Nucleus & Llama-3 & -- & -- & -- \\
        \bottomrule
    \end{tabular}
    \caption{Parameters and models used for each sampling setting.}
    \label{tab:sampling-setting}
\end{table*}

\begin{figure*}[t]
    \centering
    \includegraphics[width=1\textwidth]{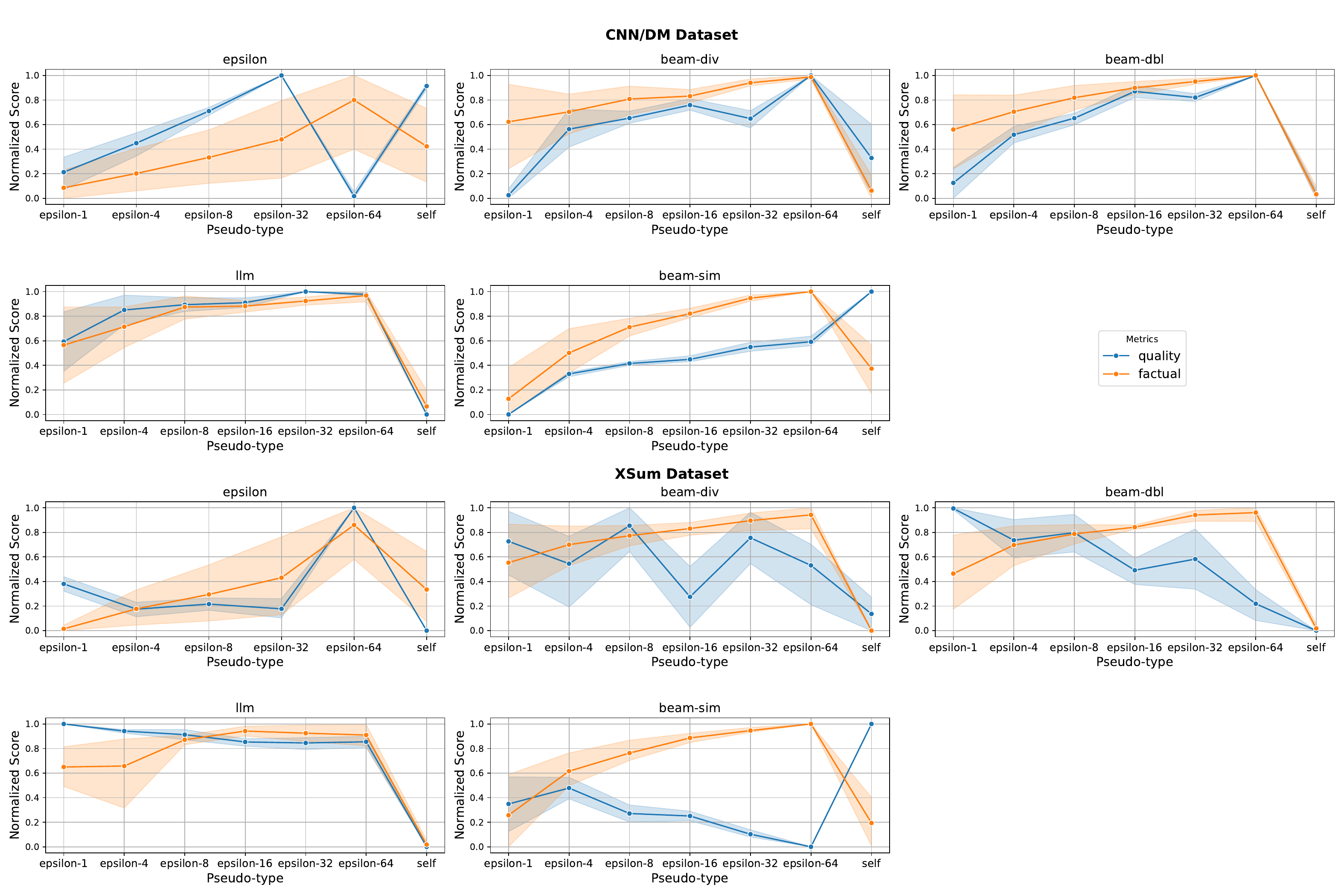}
    \caption{Average of normalized metrics based on the pseudo-reference type. Each chart represent the sampling setting for each dataset. \textbf{Top:} The average metrics for CNN/DM dataset where the sampling settings from upper left to bottom right are epsilon, beam-div, beam-dbl, llm, and beam-sim, respectively. \textbf{Bottom:} The average metrics for XSum dataset where the sampling settings from upper left to bottom right are epsilon, beam-div, beam-dbl, llm, and beam-sim, respectively.}
    \label{fig:optimal_pseudo}
\end{figure*}

\paragraph{Evaluation}
We employed an evaluation similar to that in Appendix \ref{appendix:optimal_cand}. For PLMs, we averaged scores across all models, grouped by the type of pseudo-references used. We then applied the same Min-Max normalization and categorized the results according to our defined metric groups.

\paragraph{Results}
The results, presented in Figure~\ref{fig:optimal_pseudo}, show varying outcomes depending on the sampling strategy. However, a clear pattern emerges for factuality metrics across all settings and datasets: \textbf{performance improves as the number of pseudo-references increases}. This aligns with the principle that a larger evidence set provides a more stable Monte Carlo approximation of the model's true posterior \cite{kamigaito-etal-2025-diversity}.
Furthermore, using epsilon-generated sets as pseudo-references is almost always superior to using the candidate set itself. This confirms that the two sets serve distinct roles: the pseudo-reference set serves as a proxy to ground truth and must be \textbf{diverse and unbiased} to accurately map the model's distribution \cite{kamigaito-etal-2025-diversity}. In contrast, the candidate set serves as the "contestant" pool and benefits from containing multiple high-quality options, even if they are biased toward high-likelihood regions. Decoupling these sets prevents the model from reinforcing its own biases.
While the trend for quality metrics is less consistent, varying by strategy and dataset (e.g., contrasting trends between CNN/DM and XSum), the signal for factuality is robust. Given this clarity, we prioritize the configuration that maximizes factuality. Therefore, a set of \textbf{64 pseudo-references generated via Epsilon Sampling} will be used as the default setting in all subsequent experiments.

\begin{table*}[t]
\centering
\footnotesize %
\setlength{\tabcolsep}{2pt} %
\resizebox{1.0\textwidth}{!}{

\begin{tabular}{@{}ll | l >{\columncolor[gray]{0.9}}l l >{\columncolor[gray]{0.9}}l l >{\columncolor[gray]{0.9}}l l >{\columncolor[gray]{0.9}}l l | l >{\columncolor[gray]{0.9}}l l >{\columncolor[gray]{0.9}}l l >{\columncolor[gray]{0.9}}l l >{\columncolor[gray]{0.9}}l l @{}}
\toprule
& & \multicolumn{9}{c|}{\textbf{CNNDM}} & \multicolumn{9}{c}{\textbf{XSUM}} \\
\cmidrule(lr){3-11} \cmidrule(lr){12-20}
\textbf{Setting} & \textbf{Reranker} & \multicolumn{1}{c}{\textbf{R1}} & \multicolumn{1}{c}{\textbf{R2}} & \multicolumn{1}{c}{\textbf{RL}} & \multicolumn{1}{c}{\textbf{BS}} & \multicolumn{1}{c}{\textbf{EM}} & \multicolumn{1}{c}{\textbf{CM}} & \multicolumn{1}{c}{\textbf{SM}} & \multicolumn{1}{c}{\textbf{Fe}} & \multicolumn{1}{c}{\textbf{Fi}} & \multicolumn{1}{c}{\textbf{R1}} & \multicolumn{1}{c}{\textbf{R2}} & \multicolumn{1}{c}{\textbf{RL}} & \multicolumn{1}{c}{\textbf{BS}} & \multicolumn{1}{c}{\textbf{EM}} & \multicolumn{1}{c}{\textbf{CM}} & \multicolumn{1}{c}{\textbf{SM}} & \multicolumn{1}{c}{\textbf{Fe}} & \multicolumn{1}{c}{\textbf{Fi}} \\
\midrule

\multirow{7}{*}{epsilon}& Baseline & 40.48 & 17.29 & 33.99 & 65.60 & 4.95 & -6.45 & 1.49 & 95.42 & 39.87 & 38.46 & 16.13 & 30.83 & 75.24 & 9.35 & -30.37 & -22.85 & 45.64 & 16.79 \\
& FENICE-0.0 & 41.01 & 17.85 & 34.64 & \underline{66.07} & 6.94 & -4.89 & 2.45 & --- & 56.25 & 38.80 & 16.42 & 31.14 & 75.52 & 16.80 & -23.97 & -17.63 & --- & 28.70 \\
& FIZZ-0.0 & 41.09 & 18.33 & 34.92 & 66.01 & 5.96 & -5.29 & \underline{2.60} & 98.33 & --- & 38.33 & 15.97 & 30.77 & 75.22 & 14.54 & -26.57 & -19.44 & 64.82 & --- \\
\cmidrule(lr){2-20}
& FENICE-0.75 & 41.06* & 17.87* & 34.71* & 65.90* & \textbf{11.62}*\textdagger\textdaggerdbl & \textbf{\underline{-3.59}}*\textdagger\textdaggerdbl & 2.25* & --- & 53.09* & \textbf{\underline{38.93}}* & \textbf{16.59}* & \textbf{\underline{31.39}}* & \textbf{\underline{75.73}}* & \textbf{26.61}*\textdagger\textdaggerdbl & \textbf{-19.54}*\textdagger\textdaggerdbl & \textbf{-16.09}* & --- & 27.63* \\
& FIZZ-0.75 & \textbf{\underline{41.18}}*\textdagger\textdaggerdbl & \textbf{\underline{18.36}}*\textdagger & \textbf{\underline{35.00}}*\textdagger\textdaggerdbl & 66.04* & \textbf{8.62}*\textdagger\textdaggerdbl & \textbf{-4.16}*\textdagger\textdaggerdbl & 2.59* & \textbf{98.44}*\textdaggerdbl & --- & 38.60*\textdagger\textdaggerdbl & 16.27*\textdagger & 31.09*\textdagger\textdaggerdbl & 75.45* & \textbf{20.87}*\textdagger\textdaggerdbl & \textbf{-23.04}*\textdagger\textdaggerdbl & -17.72* & \textbf{66.63}*\textdaggerdbl & --- \\
& MBR-1.0 & 40.77* & 17.60* & 34.42* & 65.63 & \textbf{\underline{12.41}}*\textdagger\textdaggerdbl & \textbf{\underline{-3.59}}*\textdagger\textdaggerdbl & --- & 97.33* & 47.95* & \textbf{38.86}* & \textbf{\underline{16.60}}* & \textbf{\underline{31.39}}* & \textbf{75.68} & \textbf{\underline{28.00}}*\textdagger\textdaggerdbl & \textbf{\underline{-17.49}}*\textdagger\textdaggerdbl & --- & 63.51* & 26.21* \\
& Oracle & 52.34 & 28.95 & 45.79 & 71.41 & 25.84 & -0.41 & 34.08 & \underline{99.77} & \underline{84.33} & 53.89 & 31.44 & 46.93 & 81.65 & 43.77 & -4.93 & 17.79 & \underline{86.80} & \underline{51.34} \\
\midrule

\multirow{7}{*}{beam-div}& Baseline & 36.87 & 18.12 & 32.38 & 63.31 & 13.16 & -5.33 & 2.40 & 97.23 & 72.59 & \underline{42.21} & \underline{20.97} & \underline{35.25} & \underline{73.05} & 17.95 & -26.11 & -17.85 & 56.17 & 25.55 \\
& FENICE-0.0 & 36.95 & 18.05 & 32.44 & 63.43 & \underline{13.96} & -5.22 & 1.72 & --- & 73.62 & 41.94 & 20.50 & 34.87 & 72.97 & \underline{18.74} & -25.26 & -17.57 & --- & 27.70 \\
& FIZZ-0.0 & 36.87 & 18.05 & 32.39 & 63.32 & 13.49 & -5.29 & 2.01 & 97.61 & --- & 41.86 & 20.49 & 34.84 & 72.85 & 18.20 & -25.77 & -17.80 & 59.04 & --- \\
\cmidrule(lr){2-20}
& FENICE-0.75 & \textbf{37.03}*\textdagger\textdaggerdbl & 18.11\textdagger\textdaggerdbl & \textbf{32.50}*\textdagger\textdaggerdbl & \textbf{63.51}*\textdagger\textdaggerdbl & 13.91*\textdaggerdbl & \textbf{-5.04}*\textdagger\textdaggerdbl & \textbf{2.63}*\textdagger\textdaggerdbl & --- & 73.36* & 42.04*\textdagger\textdaggerdbl & 20.53\textdagger\textdaggerdbl & 34.90*\textdagger\textdaggerdbl & 73.02*\textdagger\textdaggerdbl & 17.78*\textdaggerdbl & \textbf{-24.71}*\textdagger\textdaggerdbl & \textbf{-15.99}*\textdagger\textdaggerdbl & --- & 26.94* \\
& FIZZ-0.75 & \textbf{36.98}*\textdaggerdbl & 18.10\textdaggerdbl & \textbf{32.45}*\textdaggerdbl & 63.41*\textdaggerdbl & 13.43* & \textbf{-5.03}*\textdagger\textdaggerdbl & \textbf{3.26}*\textdagger\textdaggerdbl & \textbf{97.64}* & --- & 41.82*\textdaggerdbl & 20.42\textdaggerdbl & 34.76*\textdaggerdbl & 72.84*\textdaggerdbl & 17.46* & -25.28*\textdagger\textdaggerdbl & \textbf{-16.57}*\textdagger\textdaggerdbl & \textbf{59.12}* & --- \\
& MBR-1.0 & \textbf{\underline{37.14}}*\textdagger\textdaggerdbl & \textbf{\underline{18.16}}\textdagger\textdaggerdbl & \textbf{\underline{32.51}}*\textdagger\textdaggerdbl & \textbf{\underline{63.55}}*\textdagger\textdaggerdbl & 13.28 & \textbf{\underline{-4.75}}*\textdagger\textdaggerdbl & --- & 97.51* & 71.21 & 42.04*\textdagger\textdaggerdbl & 20.46\textdagger\textdaggerdbl & 34.83*\textdagger\textdaggerdbl & 72.93*\textdagger\textdaggerdbl & 15.08 & \textbf{\underline{-24.13}}*\textdagger\textdaggerdbl & --- & 56.61* & 24.75 \\
& Oracle & 40.80 & 21.54 & 36.22 & 65.52 & 21.40 & -2.74 & 8.68 & \underline{99.37} & \underline{82.81} & 47.25 & 25.47 & 39.95 & 75.29 & 28.08 & -17.74 & -6.98 & \underline{71.75} & \underline{38.01} \\
\midrule

\multirow{7}{*}{beam-dbl}& Baseline & 36.73 & \underline{18.00} & \underline{32.22} & 63.77 & 13.15 & -5.29 & 2.80 & 97.28 & 73.06 & \underline{42.14} & \underline{20.95} & \underline{35.21} & \underline{75.97} & 17.91 & -26.04 & -17.79 & 56.38 & 25.73 \\
& FENICE-0.0 & 36.56 & 17.48 & 31.93 & 63.77 & \underline{14.31} & -5.15 & 1.50 & --- & 73.41 & 40.89 & 19.26 & 33.59 & 75.47 & \underline{19.72} & -23.42 & -16.27 & --- & 30.20 \\
& FIZZ-0.0 & 36.46 & 17.57 & 31.91 & 63.59 & 13.30 & -5.34 & 1.86 & 97.68 & --- & 40.66 & 19.18 & 33.45 & 75.25 & 18.25 & -24.73 & -17.02 & 62.40 & --- \\
\cmidrule(lr){2-20}
& FENICE-0.75 & \textbf{36.83}*\textdagger\textdaggerdbl & 17.70\textdagger\textdaggerdbl & 32.11\textdagger\textdaggerdbl & \textbf{63.93}*\textdagger\textdaggerdbl & 14.01*\textdaggerdbl & \textbf{-4.66}*\textdagger\textdaggerdbl & \textbf{4.21}*\textdagger\textdaggerdbl & --- & 72.46 & 41.00*\textdagger\textdaggerdbl & 19.28\textdagger\textdaggerdbl & 33.58\textdagger\textdaggerdbl & 75.47*\textdagger\textdaggerdbl & 17.14*\textdaggerdbl & \textbf{-22.39}*\textdagger\textdaggerdbl & \textbf{-12.61}*\textdagger\textdaggerdbl & --- & 27.87 \\
& FIZZ-0.75 & 36.69\textdagger\textdaggerdbl & 17.70\textdagger\textdaggerdbl & 32.03\textdagger\textdaggerdbl & 63.75\textdaggerdbl & 13.45*\textdaggerdbl & \textbf{-4.83}*\textdagger\textdaggerdbl & \textbf{4.30}*\textdagger\textdaggerdbl & \textbf{97.82}*\textdaggerdbl & --- & 40.62\textdagger\textdaggerdbl & 19.05\textdagger\textdaggerdbl & 33.32\textdagger\textdaggerdbl & 75.23\textdaggerdbl & 17.49*\textdaggerdbl & -23.73*\textdagger\textdaggerdbl & \textbf{-14.93}*\textdagger\textdaggerdbl & \textbf{62.67}*\textdaggerdbl & --- \\
& MBR-1.0 & \textbf{\underline{36.99}}*\textdagger\textdaggerdbl & 17.80\textdagger\textdaggerdbl & 32.12\textdagger\textdaggerdbl & \textbf{\underline{63.96}}*\textdagger\textdaggerdbl & 13.05 & \textbf{\underline{-4.43}}*\textdagger\textdaggerdbl & --- & 97.57* & 68.90 & 40.87*\textdagger\textdaggerdbl & 19.03\textdagger\textdaggerdbl & 33.28\textdagger\textdaggerdbl & 75.28*\textdagger\textdaggerdbl & 14.03 & \textbf{\underline{-21.67}}*\textdagger\textdaggerdbl & --- & 57.02* & 24.34 \\
& Oracle & 43.28 & 23.89 & 38.63 & 67.41 & 27.73 & -1.36 & 14.99 & \underline{99.73} & \underline{88.64} & 50.26 & 28.35 & 43.20 & 79.36 & 36.38 & -11.74 & 1.46 & \underline{80.13} & \underline{47.13} \\
\midrule

\multirow{7}{*}{beam-sim}& Baseline & 36.75 & \underline{17.71} & \underline{32.09} & 64.77 & 13.01 & -5.44 & 2.29 & 96.99 & 68.63 & \underline{41.87} & \underline{20.42} & \underline{34.81} & \underline{75.93} & 16.69 & -26.81 & -18.56 & 54.51 & 24.79 \\
& FENICE-0.0 & 35.86 & 16.18 & 30.94 & 64.48 & \underline{15.13} & -5.47 & -0.24 & --- & 69.21 & 38.14 & 16.43 & 30.26 & 74.34 & \underline{19.76} & -21.65 & -15.67 & --- & 32.99 \\
& FIZZ-0.0 & 35.85 & 16.55 & 31.13 & 64.31 & 13.51 & -5.67 & 0.48 & 97.61 & --- & 37.45 & 15.92 & 29.74 & 73.82 & 16.46 & -24.14 & -17.62 & 65.25 & --- \\
\cmidrule(lr){2-20}
& FENICE-0.75 & \textbf{36.82}\textdagger\textdaggerdbl & 16.98\textdagger\textdaggerdbl & 31.63\textdagger\textdaggerdbl & \textbf{\underline{64.92}}*\textdagger\textdaggerdbl & 14.00*\textdaggerdbl & \textbf{-4.37}*\textdagger\textdaggerdbl & \textbf{6.35}*\textdagger\textdaggerdbl & --- & 66.27 & 38.59\textdagger\textdaggerdbl & 16.54\textdagger\textdaggerdbl & 30.31\textdagger\textdaggerdbl & 74.43*\textdagger\textdaggerdbl & 14.79*\textdaggerdbl & \textbf{-19.08}*\textdagger\textdaggerdbl & \textbf{-7.85}*\textdagger\textdaggerdbl & --- & 27.32 \\
& FIZZ-0.75 & 36.51\textdagger\textdaggerdbl & 17.05\textdagger\textdaggerdbl & 31.60\textdagger\textdaggerdbl & 64.67\textdagger\textdaggerdbl & 13.76*\textdaggerdbl & \textbf{-4.77}*\textdagger\textdaggerdbl & \textbf{4.54}*\textdagger\textdaggerdbl & \textbf{97.78}*\textdaggerdbl & --- & 37.81\textdagger\textdaggerdbl & 16.16\textdagger\textdaggerdbl & 30.00\textdagger\textdaggerdbl & 73.98\textdagger\textdaggerdbl & 16.54*\textdaggerdbl & -21.88*\textdagger\textdaggerdbl & \textbf{-13.58}*\textdagger\textdaggerdbl & \textbf{65.76}*\textdaggerdbl & --- \\
& MBR-1.0 & \textbf{\underline{36.98}}*\textdagger\textdaggerdbl & 17.07\textdagger\textdaggerdbl & 31.64\textdagger\textdaggerdbl & \textbf{64.88}*\textdagger\textdaggerdbl & 12.59 & \textbf{\underline{-4.34}}*\textdagger\textdaggerdbl & --- & 97.22* & 61.13 & 38.33*\textdagger\textdaggerdbl & 16.26\textdagger\textdaggerdbl & 29.86\textdagger\textdaggerdbl & 74.09*\textdagger\textdaggerdbl & 11.97 & \textbf{\underline{-18.16}}*\textdagger\textdaggerdbl & --- & 54.97* & 22.75 \\
& Oracle & 46.86 & 27.29 & 42.09 & 69.82 & 41.07 & -0.42 & 22.69 & \underline{99.83} & \underline{92.77} & 54.17 & 32.40 & 47.78 & 81.17 & 49.88 & -4.98 & 14.10 & \underline{87.98} & \underline{58.28} \\
\midrule

\multirow{7}{*}{llm}& Baseline & 35.78 & 13.89 & 29.21 & 64.18 & 5.30 & -2.03 & 20.95 & 98.39 & 24.70 & 19.21 & 5.18 & \underline{13.26} & 61.89 & 0.83 & -6.11 & 8.68 & 88.27 & 22.32 \\
& FENICE-0.0 & 36.05 & 14.07 & 29.48 & 64.31 & 6.05 & -1.90 & 20.96 & --- & 30.80 & \underline{19.24} & 5.09 & 13.24 & 61.91 & 0.85 & -5.34 & 8.85 & --- & 27.76 \\
& FIZZ-0.0 & 36.20 & 14.17 & 29.60 & 64.36 & 5.41 & -2.05 & 20.73 & 98.88 & --- & \underline{19.24} & 5.06 & 13.25 & 61.90 & 0.79 & -5.67 & 8.16 & 90.59 & --- \\
\cmidrule(lr){2-20}
& FENICE-0.75 & \textbf{36.22}*\textdagger & \textbf{14.45}*\textdagger\textdaggerdbl & 29.54* & \textbf{\underline{64.53}}*\textdagger\textdaggerdbl & \textbf{6.24}*\textdagger\textdaggerdbl & \textbf{-1.50}*\textdagger\textdaggerdbl & \textbf{30.36}*\textdagger\textdaggerdbl & --- & 28.02* & 19.20*\textdagger & \textbf{5.20}*\textdagger\textdaggerdbl & 13.17* & \textbf{\underline{61.97}}*\textdagger\textdaggerdbl & \textbf{1.28}*\textdagger\textdaggerdbl & \textbf{-3.50}*\textdagger\textdaggerdbl & \textbf{18.98}*\textdagger\textdaggerdbl & --- & 24.80* \\
& FIZZ-0.75 & \textbf{\underline{36.27}}*\textdagger\textdaggerdbl & \textbf{14.33}*\textdagger\textdaggerdbl & \textbf{\underline{29.63}}*\textdagger & \textbf{64.46}*\textdagger\textdaggerdbl & 5.67*\textdaggerdbl & \textbf{-1.75}*\textdagger\textdaggerdbl & \textbf{24.79}*\textdagger\textdaggerdbl & 98.86* & --- & \underline{19.24}*\textdagger\textdaggerdbl & 5.10*\textdagger\textdaggerdbl & 13.24*\textdagger & \textbf{61.94}*\textdagger\textdaggerdbl & \textbf{0.89}*\textdaggerdbl & \textbf{-4.50}*\textdagger\textdaggerdbl & \textbf{13.17}*\textdagger\textdaggerdbl & 90.56* & --- \\
& MBR-1.0 & 36.12* & \textbf{\underline{14.50}}*\textdagger\textdaggerdbl & 29.40* & \textbf{64.51}*\textdagger\textdaggerdbl & \textbf{\underline{6.41}}*\textdagger\textdaggerdbl & \textbf{\underline{-1.42}}*\textdagger\textdaggerdbl & --- & 98.50* & 24.24 & 19.14* & \textbf{\underline{5.21}}*\textdagger\textdaggerdbl & 13.11* & \textbf{\underline{61.97}}*\textdagger\textdaggerdbl & \textbf{\underline{1.59}}*\textdagger\textdaggerdbl & \textbf{\underline{-3.19}}*\textdagger\textdaggerdbl & --- & 88.19* & 21.68 \\
& Oracle & 41.77 & 19.26 & 34.95 & 67.12 & 15.74 & -0.39 & 48.60 & \underline{99.80} & \underline{55.72} & 23.84 & 8.35 & 17.11 & 64.48 & 3.91 & -0.92 & 32.57 & \underline{96.45} & \underline{50.74} \\

\bottomrule
\end{tabular}
}
\caption{Results on the validation set for each metric, separated by dataset. The bold scores are the highest value scores for each metric. ``---'' indicates the skipped settings because the metrics used in the reranking and evaluation are identical. Each abbreviation represent each metric, \textbf{R1} - ROUGE-1, \textbf{R2} - ROUGE-2, \textbf{RL} - ROUGE-Lsum, \textbf{BS} - BERTScore, \textbf{EM} - MENLI-Entailment, \textbf{CM} - MENLI-Contradiction, \textbf{SM} - MENLI-Summarization, \textbf{Fe} - FENICE, and \textbf{Fi} - FIZZ.}
\label{tab:val_results}
\end{table*}

\subsection{Validation Results}
We conclude our preliminary study by testing the optimal hyperparameters using the validation subset. Specifically, we use 16 candidates for each decoding strategy, 64 pseudo-references generated via Epsilon Sampling, and $w=0.75$ for the combination weight.  

The detailed results for all metrics are available in Table~\ref{tab:val_results}. For CNN/DM dataset, ConSUM improves summary quality for nearly all sampling methods, with the notable exception of Epsilon Sampling. This is likely because the epsilon candidates were being compared against a pseudo-reference set that was too similar to themselves. Our system consistently outperforms the baselines on most factuality metrics, demonstrating its effectiveness at improving factual consistency for this dataset. 

For XSum, the baseline methods often achieve higher quality scores. The main exception is the LLM sampling setting, where ConSUM provides a clear benefit. This suggests that for LLM candidates, which are optimized only for source alignment, our method successfully reranks for better alignment with the model's overall distribution. Similar to the CNN/DM results, \textbf{ConSUM excels in factuality}, winning on nearly all metrics across the different sampling methods. The few exceptions likely occur when baseline summaries happen to have a higher factual overlap with the specific gold references in the validation set.

\section{Evaluation Metrics}
\label{appendix:metric_def}
We divide the evaluation metrics into two groups, Quality and Factuality as shown in Table \ref{tab:eval_metrics}. Below are the explanation for each metric.

\paragraph{ROUGE}
ROUGE \cite{Lin-2004-ROUGEPackageAutomatic} is an n-gram based metric that is widely used in summarization evaluation. We implement the ROUGE using HuggingFace \texttt{evaluate} library.

\paragraph{BERTScore}
BERTScore \cite{Zhang_EtAl-2020-BERTScoreEvaluatingTexta} is a semantic-based similarity metric. It is also often used as a summary quality metric along ROUGE. We implement the BERTScore using HuggingFace \texttt{evaluate} library. In addition, we used the \texttt{DeBERTa-XLarge-MNLI} model~\cite{he2021deberta}, as it is the top-performing model at the time of this research.\footnote{Based on the GitHub \url{https://github.com/Tiiiger/bert_score}, accessed on July 19, 2025.} 

\paragraph{MoverScore}
MoverScore \cite{zhao_moverscore_2019} is a semantic-based similarity metric. Differs from BERTScore, it uses Earth Mover's Distance (EMD) on top of contextualized word embeddings. 

\paragraph{MENLI}
MENLI \cite{Chen_Eger-2023-MENLIRobustEvaluation} stands for \textbf{ME}trics from \textbf{NLI}. It is a NLI-based metric that measures a hypothesis, given the premise in three classes: Entailment -- a hypothesis is true given the premise; Contradiction -- a hypothesis is false given the premise; Neutral -- the relationship is neither entailment nor contradiction. It can be used for various text generation tasks. We utilize the three scores provided by MENLI: entailment, contradiction, and summarization. All associated parameters are adopted from the original work. Details of each parameter are in Appendix \ref{appendix:menli_param}. 

\paragraph{UniEval}
UniEval \cite{zhong_towards_2022} is a unified multi-dimensional evaluator. It measures all dimensions through Boolean Question Answering (QA) problem. It includes 4 different aspects and the total score: Coherence, Consistency, Fluency, Relevance, and Overall.

\paragraph{FENICE}
FENICE \cite{Scire_EtAl-2024-FENICEFactualityEvaluation} stands for \textbf{F}actuality \textbf{E}valuation of summarization based on \textbf{N}atural language \textbf{I}nference and \textbf{C}laim \textbf{E}xtraction. It is a two-step factuality metric comprises claim extraction and NLI alignment. It is based on the concept of Atomic Content Unit (ACU) proposed by \citet{nenkova_evaluating_2004}. For FENICE \cite{Scire_EtAl-2024-FENICEFactualityEvaluation}, on the claim extraction step, we diverge from the original paper's use of ChatGPT and instead use the publicly available T5 distillation model provided by the authors\footnote{\url{https://huggingface.co/Babelscape/t5-base-summarization-claim-extractor}}.

\paragraph{FIZZ}
FIZZ \cite{Yang_EtAl-2024-FIZZFactualInconsistency} stands for \textbf{F}actual \textbf{I}nconsistency Detection by \textbf{Z}oom-in Summary and \textbf{Z}oom-out Document. Similar to FENICE, it is a two-step factuality metric comprises of atomic fact decomposition and NLI alignment. However, it provides more interpretability through comparison at fine-grained atomic fact level. For our FIZZ \cite{Yang_EtAl-2024-FIZZFactualInconsistency} setup, we use the Orca-2 model\footnote{\url{https://huggingface.co/microsoft/Orca-2-7b}} as the decomposer and set the granularity level to 3G.

\begin{figure}[t]
    \centering
    \includegraphics[width=1\linewidth]{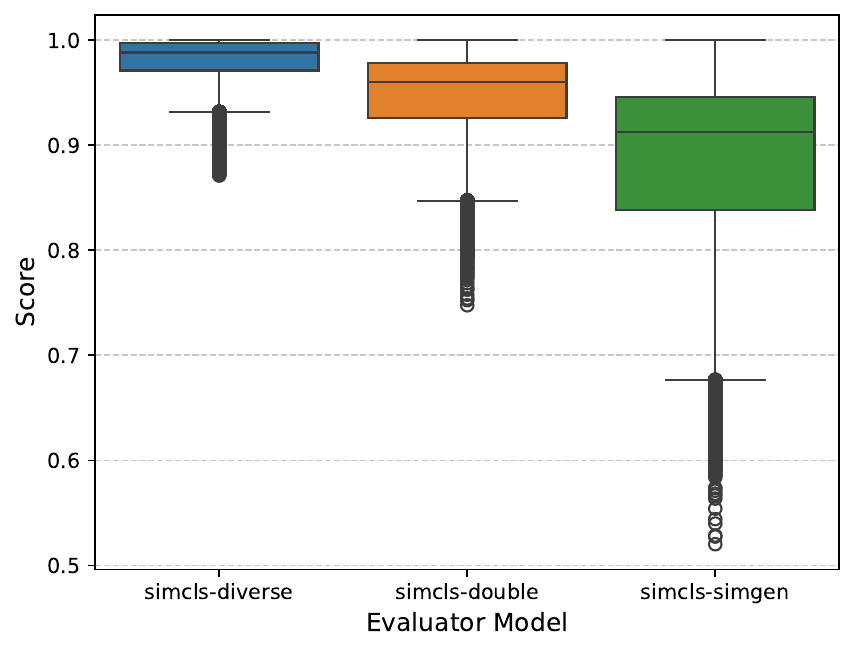}
    \caption{Comparison between SimCLS models by DBS setting}
    \label{fig:simcls_compare}
\end{figure}

\paragraph{SimCLS}
SimCLS \cite{liu_simcls_2021} is a contrastive learning-based scoring model based on RoBERTa, originally proposed for candidate reranking. We trained custom SimCLS models by generating candidates from the training subset of each dataset for every PLM. Specifically, we trained a distinct model for each DBS configuration, resulting in a total of 18 models (2 datasets x 3 models x 3 DBS configuration). As illustrated in Figure~\ref{fig:simcls_compare}, the version trained using the \textbf{\texttt{beam-div} setting} demonstrated the most consistent performance across datasets. Therefore, we utilize this specific SimCLS model for all relevant experiments.

\section{Summarization Models Details}
\label{appendix:summ_models}

\paragraph{BART}
BART stands for Bidirectional and Auto-Regressive Transformer \cite{Lewis_EtAl-2020-BARTDenoisingSequencetoSequence}. It is a denoising autoencoder that is trained by corrupting text with an arbitrary noising function and learning to reconstruct the original text. This objective makes the model excels in text generation tasks, especially abstractive summarization. Our study used the publicly available \texttt{BART-Large} model, fine-tuned to the respective dataset, CNN/DM\footnote{\url{https://huggingface.co/facebook/bart-large-cnn}} and XSum \footnote{\url{https://huggingface.co/facebook/bart-large-xsum}}

\paragraph{PEGASUS}
PEGASUS stands for Pre-training with Extracted Gap-sentences for Abstractive Summarization \cite{Zhang_EtAl-2020-PEGASUSPretrainingExtracted}. It is a transformer-based model specifically designed for abstractive summarization through Gap Sentence Generation (GSG) objective. This objective forces the model to learn a high-level understanding of the source text. Our study used the publicly available PEGASUS model, fine-tuned to the respective dataset, CNN/DM\footnote{\url{https://huggingface.co/google/pegasus-cnn\_dailymail}} and XSum\footnote{\url{https://huggingface.co/google/pegasus-xsum}}

\paragraph{T5}
T5 stands for Text-to-Text Transfer Transformer \cite{Raffel_EtAl-2023-ExploringLimitsTransfer}. It is a model that unifies all NLP tasks into a single text-to-text format. Every task is reframed as problem of generating a target text string from an input text string with a specific prefix to allow the model to be fine-tuned to a wide variety of tasks. 
As there were no reliably-performing, publicly available T5 checkpoints for these summarization datasets at the time of our research, we finetuned the \texttt{t5-large} model ourselves. The process was conducted separately for each dataset, running for 5~epochs with a learning rate of $1 \times 10^{-4}$ on two RTX A6000 GPUs. The performance of our fine-tuned models on the validation sets is presented in Table~\ref{tab:t5_validation_results}.

\paragraph{Llama-3}
Llama-3~\cite{grattafiori2024llama3herdmodels} is a family of Large Language Models (LLMs) developed by Meta. It is a decoder-only transformer that is pre-trained on a massive and diverse dataset. The large amount of parameters enables the model to excel at complex reasoning and instruction following. Specifically, we used the publicly available \texttt{Llama-3-8B-Instruct} model\footnote{\url{https://huggingface.co/meta-llama/Meta-Llama-3-8B-Instruct}}. The prompt used is in Appendix \ref{appendix:llm_prompt}

\section{Sampling Settings}
\label{appendix:sampling_setting}

we explore two types of decoding on the pretrained models: Diverse Beam Search (DBS) \cite{vijayakumar_diverse_2018} and Epsilon Sampling \cite{hewitt-etal-2022-truncation}. In addition, we explore using Large Language Model (LLM) as the summarization model for our method. For DBS, we test multiple parameter configurations inspired by previous work~\cite{Dixit_EtAl-2023-ImprovingFactualityAbstractive, liu_simcls_2021}, in addition to our own settings, which doubled the parameters' value from \texttt{beam-div} setting. For the LLM-based method, we use the Llama-3 model to generate 16 candidate summaries per source document. The detailed configurations for each sampling method are shown in Table~\ref{tab:sampling-setting}.

\begin{table}[t!]
\centering
\begin{tabular}{@{}lcc@{}}
\toprule
\textbf{Metric} & \textbf{XSUM} & \textbf{CNN/DM} \\
\midrule
ROUGE-1   & 36.47 & 40.73 \\
ROUGE-2   & 14.29 & 17.58 \\
ROUGE-L   & 28.69 & 27.27 \\
ROUGE-Lsum& 28.70 & 33.94 \\
\bottomrule
\end{tabular}
\caption{Validation results for the T5-Large model after finetuning on the XSUM and CNN/DM datasets.}
\label{tab:t5_validation_results}
\end{table}

\section{Generated Summary Length}
\label{appendix:length_summary}

The length of model-generated summaries are reported in Table \ref{tab:summ_length}. In this study, we purposely do not apply length restriction to emulate the average usage of summary generation using the respective models. 

\begin{table}[t!]
    \centering
    \begin{tabular}{@{}llcc@{}}
    \toprule
    & & \multicolumn{2}{c}{\textbf{Dataset}} \\
    \cmidrule(lr){3-4}
    \textbf{Setting} & \textbf{Model} & \textbf{CNN/DM} & \textbf{XSum} \\
    \midrule
    \multirow{3}{*}{epsilon} & bart & 79.51 & 26.81 \\
    & pegasus & 65.41 & 24.47 \\
    & t5-large & 89.33 & 30.61 \\
    \midrule
    \multirow{3}{*}{beam-sim} & bart & 76.43 & 26.62 \\
    & pegasus & 73.23 & 23.81 \\
    & t5-large & 20.89 & 20.71 \\
    \midrule
    \multirow{3}{*}{beam-dbl} & bart & 78.69 & 25.84 \\
    & pegasus & 83.17 & 25.14 \\
    & t5-large & 20.89 & 20.74 \\
    \midrule
    \multirow{3}{*}{beam-div} & bart & 78.33 & 25.61 \\
    & pegasus & 79.64 & 24.17 \\
    & t5-large & 20.89 & 20.71 \\
    \midrule
    llm & llama-3 & 138.64 & 115.73 \\
    \bottomrule
    \end{tabular}
    \caption{Average length of model-generated summaries across different settings and datasets.}
    \label{tab:summ_length}
\end{table}

\section{MENLI Parameters}
\label{appendix:menli_param}

We utilize the three scores provided by MENLI: entailment, contradiction, and summarization as evaluation metrics. In addition, we use the same MENLI-summarization as the utility function for our MBR decoding method implemented using the \texttt{mbrs} \cite{Deguchi_EtAl-2024-MbrsLibraryMinimum} library. All associated parameters are adopted from the original work \cite{Chen_Eger-2023-MENLIRobustEvaluation} and the parameters are in Table \ref{tab:menli_params}.

\section{Additional Test Subset Results}
\label{appendix:additional_test_res}
As shown in Appendix \ref{appendix:sampling_setting}, we experimented with 5 different sampling settings. We also explained the definition of each sampling setting in Appendix \ref{appendix:sampling_setting}. We show the performance for all 5 settings in Table~\ref{tab:cnndm_detailed_results} for the CNN/DM dataset and Table~\ref{tab:xsum_detailed_results} for the XSum dataset. In addition, we included the performance of divided by setting, model, and reranker in Table \ref{tab:cnn_model_sett_rerank} and Table \ref{tab:xsum_model_sett_rerank} and divided by setting and model in Table \ref{tab:cnn_model_sett} and Table \ref{tab:xsum_model_sett} for CNN/DM and XSum dataset, respectively.

The additional results using \texttt{beam-div} and \texttt{beam-dbl} exhibit trends similar to, or better than, the \texttt{beam-sim} setting. As detailed in Tables \ref{tab:cnndm_detailed_results} and \ref{tab:xsum_detailed_results}, both configurations outperform the baselines, particularly on the CNN/DM dataset, where our method achieves superior ROUGE scores across all metrics. Furthermore, a model-wise analysis demonstrates consistent improvements, corroborating \textbf{the robustness of our method regardless of the underlying generation model}.

\section{Example Summaries}
\label{appendix:example_summaries}

We sampled the summaries from both BART model and LlaMa-3 model to represent pre-trained model and LLM. Table \ref{tab:example_cnn_beam_bart} and Table \ref{tab:example_xsum_beam_bart} show the generated summaries using BART and \texttt{beam-sim} as the setting. Due to the fine-tuning of the model to the respective datasets, the generated summaries are limited in length. Hence, the changes in the generated summaries are mainly the main focus of the summary. On the other hand, LlaMa-3 summaries are not limited in length, thus the main difference in the generated summaries are the improvement in factual accuracy and addition of key details.

\begin{table}[t!]
\centering
\setlength{\tabcolsep}{4pt} %
\begin{tabular}{@{}lccc@{}}
\toprule
\textbf{Parameter} & \textbf{ent} & \textbf{con} & \textbf{sum} \\
\midrule
\texttt{direction} & \texttt{rh} & \texttt{rh} & \texttt{hr} \\
Formula & $e$ & $-c$ & $e-c$ \\
\texttt{nli\_weight} & 1.0 & 1.0 & 1.0 \\
\texttt{combine\_with} & \texttt{None} & \texttt{None} & \texttt{None} \\
\texttt{model} & \texttt{D} & \texttt{D} & \texttt{D} \\
\bottomrule
\end{tabular}
\caption{Parameter configurations for different MENLI variations.}
\label{tab:menli_params}
\end{table}

\section{Aspect Definition}
\label{appendix:aspect_def}
To define a clear definition of the aspects. We define them as follows:

\textbf{Factuality} -- A summary is factual if all the information presented in it is consistent with the information in the source document. It means no details are hallucinated (made up), contradicted, or distorted from the original text. It means that the meaning from the original source is preserved.

\textbf{Informativeness} -- Measures the inclusivity of crucial information from the document. Highly informative summary includes all key points, main ideas, and essential facts. It ensures that the reader can understand the core content without needing to refer back to the source

\textbf{Fluency} -- A fluent summary is well-written, grammatically correct, and easy to read and understand. The text should flow naturally, use appropriate vocabulary, and have correct sentence structure, capitalization, and punctuation.

\begin{figure*}[!t]
    \centering
    \includegraphics[width=1\textwidth]{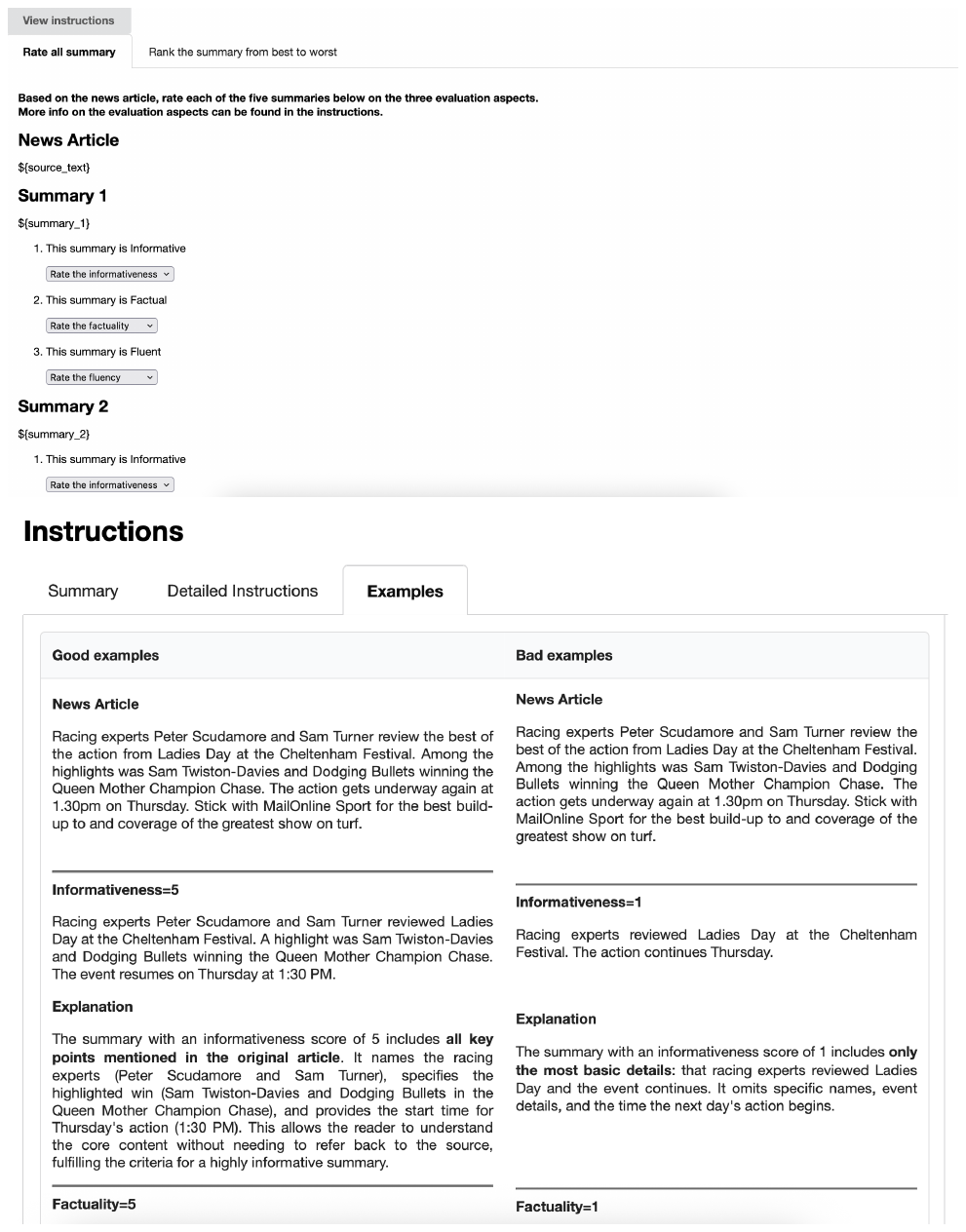}
    \caption{MTurk Page for human evaluation.}
    \label{fig:mturk_page}
\end{figure*}

\section{Human Instruction}
\label{appendix:mturk_page}

Figure \ref{fig:mturk_page} shows the screenshot of the human instruction from the MTurk page.

\section{LLM Prompt}
\label{appendix:llm_prompt}
The prompt used in LLama-3 to generate the summary is as follows. The \texttt{src} refers to the source document/the news article from the dataset:

\lstset{
    backgroundcolor=\color[RGB]{245,245,245},
    breaklines=true,
    breakindent=0pt,
    xrightmargin=5pt,
    basicstyle=\ttfamily\small,
    frame=trbl,
    frameround = tttt,
}
\noindent
\begin{lstlisting}
{
    "role": "system",
    "content": "You are an assistant who replies with a summary to every message.",
},
{"role": "user", "content": f"Summarize the following text: \n\n {src}"}
\end{lstlisting}

\begin{table*}[ht]
\centering
\footnotesize %
\setlength{\tabcolsep}{2pt} %
\resizebox{1.0\textwidth}{!}{

\begin{tabular}{@{}ll | l >{\columncolor[gray]{0.9}}l l >{\columncolor[gray]{0.9}}l l |>{\columncolor[gray]{0.9}}l l >{\columncolor[gray]{0.9}}l l >{\columncolor[gray]{0.9}}l l >{\columncolor[gray]{0.9}}l  @{}}
\toprule
& & \multicolumn{5}{c|}{\textbf{Quality}} & \multicolumn{7}{c}{\textbf{Factuality}} \\
\cmidrule(lr){3-7} \cmidrule(lr){8-14}
\textbf{Setting} & \textbf{Reranker} & \multicolumn{1}{c}{\textbf{R1}} & \multicolumn{1}{c}{\textbf{R2}} & \multicolumn{1}{c}{\textbf{RL}} & \multicolumn{1}{c}{\textbf{BS}} & \multicolumn{1}{c|}{\textbf{MS}} & \multicolumn{1}{c}{\textbf{EM}} & \multicolumn{1}{c}{\textbf{CM}} & \multicolumn{1}{c}{\textbf{SM}} & \multicolumn{1}{c}{\textbf{UE}} & \multicolumn{1}{c}{\textbf{Fe}} & \multicolumn{1}{c}{\textbf{Fi}} & \multicolumn{1}{c}{\textbf{SC}} \\
\midrule

\multirow{7}{*}{epsilon}& Baseline & 39.96 & 16.88 & 33.38 & 65.52 & 58.03 & 4.46 & -6.62 & 1.63 & 81.37 & 95.52 & 39.36 & \underline{99.74} \\
& FENICE-0.0 & 40.52 & 17.46 & 34.06 & \underline{66.02} & 58.19 & 5.96 & -5.21 & 2.14 & \underline{85.18} & --- & \underline{55.83} & 99.71 \\
& FIZZ-0.0 & 40.54 & 17.81 & 34.23 & 65.91 & 58.17 & 5.18 & -5.59 & 2.39 & 84.65 & 98.39 & --- & 99.72 \\
\cmidrule(lr){2-14}
& FENICE-0.75 & \textbf{40.60}* & 17.47* & 34.11* & 65.86* & \textbf{\underline{58.25}}*\textdagger\textdaggerdbl & \textbf{10.44}*\textdagger\textdaggerdbl & \textbf{-3.82}*\textdagger\textdaggerdbl & 2.36* & 83.34* & --- & 52.44* & 99.70 \\
& FIZZ-0.75 & \textbf{\underline{40.67}}*\textdagger\textdaggerdbl & \textbf{\underline{17.87}}*\textdagger & \textbf{\underline{34.36}}*\textdagger\textdaggerdbl & 65.94* & \textbf{58.24}*\textdagger\textdaggerdbl & \textbf{7.68}*\textdagger\textdaggerdbl & \textbf{-4.46}*\textdagger\textdaggerdbl & \textbf{\underline{2.44}}* & 84.16* & \textbf{\underline{98.45}}*\textdaggerdbl & --- & 99.70 \\
& MBR-1.0 & 40.36* & 17.23* & 33.87* & 65.60* & 58.19* & \textbf{\underline{11.29}}*\textdagger\textdaggerdbl & \textbf{\underline{-3.78}}*\textdagger\textdaggerdbl & --- & 81.41 & 97.35* & 47.26* & 99.69 \\
\cmidrule(lr){2-14}
& Oracle & \textit{51.85} & \textit{28.45} & \textit{45.15} & \textit{71.35} & \textit{61.29} & \textit{23.82} & \textit{-0.47} & \textit{35.02} & \textit{95.20} & \textit{99.80} & \textit{84.22} & \textit{99.89} \\
\midrule

\multirow{7}{*}{beam-div}& Baseline & 36.44 & 17.70 & 31.86 & 64.87 & 56.81 & 12.42 & -5.49 & 1.96 & \underline{90.53} & 97.29 & 72.28 & 97.59 \\
& FENICE-0.0 & 36.54 & 17.63 & 31.95 & 65.02 & 56.88 & \underline{13.33} & -5.45 & 1.44 & 90.30 & --- & \underline{73.52} & 97.61 \\
& FIZZ-0.0 & 36.42 & 17.63 & 31.86 & 64.86 & 56.82 & 12.77 & -5.42 & 1.61 & 90.10 & 97.62 & --- & 97.59 \\
\cmidrule(lr){2-14}
& FENICE-0.75 & \textbf{36.60}*\textdagger\textdaggerdbl & 17.67\textdagger & \textbf{31.98}*\textdagger\textdaggerdbl & \textbf{65.05}*\textdagger\textdaggerdbl & \textbf{\underline{56.89}}*\textdaggerdbl & 13.21*\textdaggerdbl & \textbf{-5.23}*\textdagger\textdaggerdbl & \textbf{2.43}*\textdagger\textdaggerdbl & 90.41\textdagger\textdaggerdbl & --- & 73.15* & 97.61*\textdagger\textdaggerdbl \\
& FIZZ-0.75 & 36.53*\textdaggerdbl & 17.68\textdaggerdbl & 31.92*\textdaggerdbl & 64.94*\textdaggerdbl & 56.84*\textdaggerdbl & 12.74* & \textbf{-5.22}*\textdagger\textdaggerdbl & \textbf{\underline{2.95}}*\textdagger\textdaggerdbl & 90.13 & \textbf{\underline{97.69}}*\textdaggerdbl & --- & 97.61*\textdaggerdbl \\
& MBR-1.0 & \textbf{\underline{36.74}}*\textdagger\textdaggerdbl & \textbf{\underline{17.78}}*\textdagger\textdaggerdbl & \textbf{\underline{32.02}}*\textdagger\textdaggerdbl & \textbf{\underline{65.06}}*\textdagger\textdaggerdbl & 56.87*\textdaggerdbl & 12.83* & \textbf{\underline{-4.94}}*\textdagger\textdaggerdbl & --- & 90.39\textdagger\textdaggerdbl & \textbf{97.63}* & 70.88 & \textbf{\underline{97.64}}*\textdagger\textdaggerdbl \\
\cmidrule(lr){2-14}
& Oracle & \textit{40.42} & \textit{21.11} & \textit{35.73} & \textit{66.93} & \textit{57.83} & \textit{20.66} & \textit{-2.84} & \textit{8.46} & \textit{93.78} & \textit{99.45} & \textit{82.71} & \textit{98.07} \\
\midrule

\multirow{7}{*}{beam-dbl}& Baseline & 36.25 & \underline{17.55} & \underline{31.66} & 64.76 & 56.74 & 12.32 & -5.47 & 2.40 & \underline{90.47} & 97.33 & 72.76 & 97.60 \\
& FENICE-0.0 & 36.21 & 17.11 & 31.48 & 64.82 & 56.80 & \underline{13.59} & -5.35 & 1.26 & 90.01 & --- & \underline{73.55} & 97.62 \\
& FIZZ-0.0 & 35.96 & 17.10 & 31.34 & 64.59 & 56.68 & 12.70 & -5.45 & 1.74 & 89.75 & 97.70 & --- & 97.61 \\
\cmidrule(lr){2-14}
& FENICE-0.75 & \textbf{36.39}*\textdagger\textdaggerdbl & 17.28\textdagger\textdaggerdbl & 31.59\textdagger\textdaggerdbl & \textbf{64.93}*\textdagger\textdaggerdbl & \textbf{\underline{56.82}}*\textdagger\textdaggerdbl & 13.29*\textdaggerdbl & \textbf{-4.88}*\textdagger\textdaggerdbl & \textbf{4.06}*\textdagger\textdaggerdbl & 90.31\textdagger\textdaggerdbl & --- & 72.48 & \textbf{97.65}*\textdagger\textdaggerdbl \\
& FIZZ-0.75 & 36.19\textdaggerdbl & 17.23\textdagger\textdaggerdbl & 31.47\textdaggerdbl & 64.73\textdaggerdbl & 56.72\textdaggerdbl & 12.77* & \textbf{-5.05}*\textdagger\textdaggerdbl & \textbf{\underline{4.14}}*\textdagger\textdaggerdbl & 89.93\textdaggerdbl & \textbf{\underline{97.81}}*\textdaggerdbl & --- & \textbf{97.63}*\textdaggerdbl \\
& MBR-1.0 & \textbf{\underline{36.56}}*\textdagger\textdaggerdbl & 17.38\textdagger\textdaggerdbl & 31.59\textdagger\textdaggerdbl & \textbf{\underline{64.95}}*\textdagger\textdaggerdbl & 56.79*\textdaggerdbl & 12.53 & \textbf{\underline{-4.69}}*\textdagger\textdaggerdbl & --- & 90.26\textdagger\textdaggerdbl & \textbf{97.72}* & 68.73 & \textbf{\underline{97.68}}*\textdagger\textdaggerdbl \\
\cmidrule(lr){2-14}
& Oracle & \textit{42.88} & \textit{23.41} & \textit{38.11} & \textit{68.13} & \textit{58.47} & \textit{26.56} & \textit{-1.43} & \textit{14.97} & \textit{95.11} & \textit{99.74} & \textit{88.56} & \textit{98.32} \\
\midrule

\multirow{7}{*}{beam-sim}& Baseline & 36.37 & \underline{17.34} & \underline{31.64} & 64.78 & 56.79 & 12.26 & -5.67 & 2.24 & 89.86 & 97.02 & 68.44 & 97.63 \\
& FENICE-0.0 & 35.69 & 15.94 & 30.67 & 64.61 & 56.74 & \underline{14.56} & -5.49 & -0.13 & 89.54 & --- & \underline{68.89} & 97.65 \\
& FIZZ-0.0 & 35.60 & 16.27 & 30.78 & 64.37 & 56.65 & 12.80 & -5.91 & 0.45 & 88.94 & 97.66 & --- & 97.62 \\
\cmidrule(lr){2-14}
& FENICE-0.75 & \textbf{36.50}*\textdagger\textdaggerdbl & 16.63\textdagger\textdaggerdbl & 31.23\textdagger\textdaggerdbl & \textbf{\underline{64.97}}*\textdagger\textdaggerdbl & \textbf{\underline{56.90}}*\textdagger\textdaggerdbl & 13.37*\textdaggerdbl & \textbf{-4.54}*\textdagger\textdaggerdbl & \textbf{\underline{6.60}}*\textdagger\textdaggerdbl & \textbf{\underline{90.14}}*\textdagger\textdaggerdbl & --- & 65.85 & \textbf{97.71}*\textdagger\textdaggerdbl \\
& FIZZ-0.75 & 36.14\textdagger\textdaggerdbl & 16.65\textdagger\textdaggerdbl & 31.14\textdagger\textdaggerdbl & 64.68\textdagger\textdaggerdbl & 56.79\textdagger\textdaggerdbl & 13.19*\textdaggerdbl & \textbf{-4.99}*\textdagger\textdaggerdbl & \textbf{4.59}*\textdagger\textdaggerdbl & 89.48\textdaggerdbl & \textbf{\underline{97.80}}*\textdaggerdbl & --- & 97.65*\textdaggerdbl \\
& MBR-1.0 & \textbf{\underline{36.56}}*\textdagger\textdaggerdbl & 16.65\textdagger\textdaggerdbl & 31.14\textdagger\textdaggerdbl & \textbf{64.88}*\textdagger\textdaggerdbl & \textbf{56.87}*\textdagger\textdaggerdbl & 12.03 & \textbf{\underline{-4.44}}*\textdagger\textdaggerdbl & --- & 89.67\textdaggerdbl & 97.32* & 60.60 & \textbf{\underline{97.73}}*\textdagger\textdaggerdbl \\
\cmidrule(lr){2-14}
& Oracle & \textit{46.68} & \textit{27.01} & \textit{41.80} & \textit{69.90} & \textit{59.54} & \textit{39.80} & \textit{-0.47} & \textit{23.46} & \textit{96.12} & \textit{99.86} & \textit{92.69} & \textit{98.62} \\
\midrule

\multirow{7}{*}{llm}& Baseline & 34.83 & 13.51 & 28.34 & 64.04 & 56.56 & 4.80 & -2.12 & 21.70 & 92.58 & 98.43 & 24.98 & 99.91 \\
& FENICE-0.0 & 35.15 & 13.77 & 28.66 & 64.20 & 56.64 & 5.53 & -2.05 & 21.19 & 92.63 & --- & \underline{31.12} & 99.90 \\
& FIZZ-0.0 & 35.26 & 13.80 & 28.74 & 64.23 & 56.66 & 5.13 & -2.27 & 20.92 & 92.55 & 98.89 & --- & 99.90 \\
\cmidrule(lr){2-14}
& FENICE-0.75 & \textbf{35.31}*\textdagger & \textbf{\underline{14.11}}*\textdagger\textdaggerdbl & 28.71* & \textbf{\underline{64.38}}*\textdagger\textdaggerdbl & 56.66*\textdagger & \textbf{5.95}*\textdagger\textdaggerdbl & \textbf{-1.60}*\textdagger\textdaggerdbl & \textbf{\underline{31.14}}*\textdagger\textdaggerdbl & \textbf{92.91}*\textdagger\textdaggerdbl & --- & 28.50* & 99.91*\textdagger\textdaggerdbl \\
& FIZZ-0.75 & \textbf{\underline{35.36}}*\textdagger\textdaggerdbl & \textbf{13.98}*\textdagger\textdaggerdbl & \textbf{\underline{28.80}}*\textdagger\textdaggerdbl & \textbf{64.34}*\textdagger\textdaggerdbl & \textbf{\underline{56.68}}*\textdagger\textdaggerdbl & 5.34*\textdaggerdbl & \textbf{-1.95}*\textdaggerdbl & \textbf{25.08}*\textdagger\textdaggerdbl & \textbf{92.70}*\textdaggerdbl & \textbf{\underline{98.90}}* & --- & 99.90\textdaggerdbl \\
& MBR-1.0 & 35.15* & \textbf{14.07}*\textdagger\textdaggerdbl & 28.52* & \textbf{64.36}*\textdagger\textdaggerdbl & 56.63* & \textbf{\underline{6.02}}*\textdagger\textdaggerdbl & \textbf{\underline{-1.51}}*\textdagger\textdaggerdbl & --- & \textbf{\underline{92.99}}*\textdagger\textdaggerdbl & 98.50 & 24.73 & \textbf{\underline{99.92}}*\textdagger\textdaggerdbl \\
\cmidrule(lr){2-14}
& Oracle & \textit{40.88} & \textit{18.83} & \textit{34.05} & \textit{67.03} & \textit{57.93} & \textit{14.81} & \textit{-0.45} & \textit{49.41} & \textit{95.63} & \textit{99.81} & \textit{55.71} & \textit{99.95} \\

\bottomrule
\end{tabular}
}
\caption{Results on the test set for each metric on the CNN/DM dataset. \underline{Underline} indicates the highest scores for each metric and \textbf{bold} indicates better scores than all baselines. *, \textdagger, and \textdaggerdbl ~represent the statistical significance against Baseline, FENICE-0.0, and FIZZ-0.0, respectively (See \S\ref{subsec:exp_setting}). ``---'' indicates the skipped settings because the metrics used in the reranking and evaluation are identical. Each abbreviation represents the following metric, \textbf{R1} - ROUGE-1, \textbf{R2} - ROUGE-2, \textbf{RL} - ROUGE-L, \textbf{BS} - BERTScore, \textbf{MS} - MoverScore, \textbf{EM} - MENLI-Entailment, \textbf{CM} - MENLI-Contradiction, \textbf{SM} - MENLI-Summarization, \textbf{UE} - UniEval-Overall, \textbf{Fe} - FENICE, \textbf{Fi} - FIZZ, and \textbf{SC} - SimCLS.}
\label{tab:cnndm_detailed_results}
\end{table*}

\begin{table*}[ht]
\centering
\footnotesize %
\setlength{\tabcolsep}{2pt} %
\resizebox{1.0\textwidth}{!}{

\begin{tabular}{@{}ll | l >{\columncolor[gray]{0.9}}l l >{\columncolor[gray]{0.9}}l l |>{\columncolor[gray]{0.9}}l l >{\columncolor[gray]{0.9}}l l >{\columncolor[gray]{0.9}}l l >{\columncolor[gray]{0.9}}l  @{}}
\toprule
& & \multicolumn{5}{c|}{\textbf{Quality}} & \multicolumn{7}{c}{\textbf{Factuality}} \\
\cmidrule(lr){3-7} \cmidrule(lr){8-14}
\textbf{Setting} & \textbf{Reranker} & \multicolumn{1}{c}{\textbf{R1}} & \multicolumn{1}{c}{\textbf{R2}} & \multicolumn{1}{c}{\textbf{RL}} & \multicolumn{1}{c}{\textbf{BS}} & \multicolumn{1}{c|}{\textbf{MS}} & \multicolumn{1}{c}{\textbf{EM}} & \multicolumn{1}{c}{\textbf{CM}} & \multicolumn{1}{c}{\textbf{SM}} & \multicolumn{1}{c}{\textbf{UE}} & \multicolumn{1}{c}{\textbf{Fe}} & \multicolumn{1}{c}{\textbf{Fi}} & \multicolumn{1}{c}{\textbf{SC}} \\
\midrule

\multirow{7}{*}{epsilon}& Baseline & 38.27 & 15.91 & 30.60 & 75.20 & 59.38 & 9.50 & -31.15 & -23.49 & 85.08 & 45.67 & 16.91 & \underline{98.50} \\
& FENICE-0.0 & 38.67 & 16.29 & 30.96 & 75.47 & 59.44 & 17.41 & -24.20 & -17.95 & \underline{88.13} & --- & \underline{28.84} & 98.41 \\
& FIZZ-0.0 & 38.12 & 15.85 & 30.56 & 75.14 & 59.28 & 14.50 & -27.22 & -20.10 & 87.12 & 64.70 & --- & 98.39 \\
\cmidrule(lr){2-14}
& FENICE-0.75 & \textbf{\underline{38.68}}*\textdaggerdbl & \textbf{\underline{16.44}}*\textdagger\textdaggerdbl & \textbf{\underline{31.18}}*\textdagger\textdaggerdbl & \textbf{\underline{75.66}}*\textdagger\textdaggerdbl & \textbf{\underline{59.45}}*\textdaggerdbl & \textbf{27.04}*\textdagger\textdaggerdbl & \textbf{-20.36}*\textdagger\textdaggerdbl & \textbf{\underline{-16.40}}*\textdagger\textdaggerdbl & 88.01*\textdaggerdbl & --- & 27.79* & 98.32 \\
& FIZZ-0.75 & 38.41\textdaggerdbl & 16.17*\textdaggerdbl & 30.91*\textdaggerdbl & 75.41*\textdaggerdbl & 59.37\textdaggerdbl & \textbf{21.03}*\textdagger\textdaggerdbl & \textbf{-23.57}*\textdagger\textdaggerdbl & \underline{-18.07}*\textdaggerdbl & 87.63*\textdaggerdbl & \textbf{\underline{66.68}}*\textdaggerdbl & --- & 98.34 \\
& MBR-1.0 & 38.57*\textdaggerdbl & \textbf{16.38}*\textdaggerdbl & \textbf{31.11}*\textdagger\textdaggerdbl & \textbf{75.59}*\textdagger\textdaggerdbl & 59.39\textdaggerdbl & \textbf{\underline{28.20}}*\textdagger\textdaggerdbl & \textbf{\underline{-18.07}}*\textdagger\textdaggerdbl & --- & 87.53*\textdaggerdbl & 63.02* & 26.21* & 98.31 \\
\cmidrule(lr){2-14}
& Oracle & \textit{53.82} & \textit{31.35} & \textit{46.79} & \textit{81.63} & \textit{63.70} & \textit{43.60} & \textit{-5.32} & \textit{17.78} & \textit{94.11} & \textit{86.48} & \textit{51.10} & \textit{99.21} \\
\midrule

\multirow{7}{*}{beam-div}& Baseline & \underline{42.17} & \underline{20.83} & \underline{35.17} & \underline{75.99} & \underline{59.88} & 17.95 & -26.22 & -18.19 & 86.49 & 55.75 & 25.78 & 97.56 \\
& FENICE-0.0 & 41.84 & 20.36 & 34.75 & 75.90 & 59.79 & \underline{19.16} & -25.27 & -17.49 & 86.88 & --- & \underline{27.99} & 97.56 \\
& FIZZ-0.0 & 41.74 & 20.28 & 34.72 & 75.81 & 59.75 & 18.44 & -25.74 & -17.88 & 86.54 & 58.59 & --- & 97.55 \\
\cmidrule(lr){2-14}
& FENICE-0.75 & 41.95\textdagger\textdaggerdbl & 20.37\textdaggerdbl & 34.80\textdaggerdbl & 75.91\textdaggerdbl & 59.84\textdagger\textdaggerdbl & 18.06 & \textbf{-24.71}*\textdagger\textdaggerdbl & \textbf{\underline{-15.97}}*\textdagger\textdaggerdbl & \textbf{\underline{86.91}}*\textdaggerdbl & --- & 27.21* & \textbf{97.57}*\textdagger\textdaggerdbl \\
& FIZZ-0.75 & 41.75 & 20.20 & 34.67 & 75.80 & 59.77\textdaggerdbl & 17.92 & \textbf{-25.12}*\textdaggerdbl & \textbf{-16.69}*\textdagger\textdaggerdbl & 86.61*\textdaggerdbl & \textbf{\underline{58.76}}* & --- & 97.56\textdaggerdbl \\
& MBR-1.0 & 41.98\textdagger\textdaggerdbl & 20.32 & 34.74 & 75.86\textdaggerdbl & \underline{59.88}\textdagger\textdaggerdbl & 15.33 & \textbf{\underline{-24.14}}*\textdagger\textdaggerdbl & --- & 86.73*\textdaggerdbl & 56.11 & 24.75 & \textbf{\underline{97.59}}*\textdagger\textdaggerdbl \\
\cmidrule(lr){2-14}
& Oracle & \textit{47.14} & \textit{25.25} & \textit{39.83} & \textit{78.01} & \textit{61.24} & \textit{28.41} & \textit{-17.80} & \textit{-7.09} & \textit{90.15} & \textit{70.70} & \textit{38.00} & \textit{98.05} \\
\midrule

\multirow{7}{*}{beam-dbl}& Baseline & \underline{42.09} & \underline{20.80} & \underline{35.12} & \underline{75.91} & \underline{59.86} & 17.94 & -26.11 & -18.02 & 86.40 & 55.98 & 25.95 & 97.56 \\
& FENICE-0.0 & 40.70 & 19.15 & 33.39 & 75.39 & 59.45 & \underline{19.93} & -23.63 & -16.66 & 87.18 & --- & \underline{30.45} & 97.56 \\
& FIZZ-0.0 & 40.53 & 19.00 & 33.29 & 75.19 & 59.36 & 18.45 & -24.89 & -17.38 & 86.50 & 61.79 & --- & 97.55 \\
\cmidrule(lr){2-14}
& FENICE-0.75 & 40.92\textdagger\textdaggerdbl & 19.16\textdaggerdbl & 33.40 & 75.42\textdaggerdbl & 59.56\textdagger\textdaggerdbl & 17.28 & \textbf{-22.38}*\textdagger\textdaggerdbl & \textbf{\underline{-12.85}}*\textdagger\textdaggerdbl & \textbf{\underline{87.19}}*\textdaggerdbl & --- & 28.00* & \textbf{97.60}*\textdagger\textdaggerdbl \\
& FIZZ-0.75 & 40.53 & 18.91 & 33.20 & 75.18 & 59.39\textdaggerdbl & 17.84 & -23.78*\textdaggerdbl & \textbf{-15.12}*\textdagger\textdaggerdbl & 86.64*\textdaggerdbl & \textbf{\underline{62.11}}*\textdaggerdbl & --- & 97.56\textdaggerdbl \\
& MBR-1.0 & 40.84\textdagger\textdaggerdbl & 18.97 & 33.21 & 75.29\textdaggerdbl & 59.54\textdagger\textdaggerdbl & 14.20 & \textbf{\underline{-21.79}}*\textdagger\textdaggerdbl & --- & 86.73*\textdaggerdbl & 57.15* & 24.52 & \textbf{\underline{97.64}}*\textdagger\textdaggerdbl \\
\cmidrule(lr){2-14}
& Oracle & \textit{50.11} & \textit{28.10} & \textit{43.04} & \textit{79.32} & \textit{62.06} & \textit{36.43} & \textit{-11.95} & \textit{1.30} & \textit{91.97} & \textit{79.38} & \textit{46.95} & \textit{98.33} \\
\midrule

\multirow{7}{*}{beam-sim}& Baseline & \underline{41.78} & \underline{20.21} & \underline{34.64} & \underline{75.88} & \underline{59.87} & 16.62 & -27.06 & -18.88 & 86.26 & 54.74 & 25.05 & 97.59 \\
& FENICE-0.0 & 37.96 & 16.21 & 30.04 & 74.25 & 58.78 & \underline{19.84} & -22.05 & -16.06 & 87.14 & --- & \underline{33.23} & 97.60 \\
& FIZZ-0.0 & 37.27 & 15.68 & 29.53 & 73.74 & 58.57 & 16.79 & -24.40 & -17.94 & 85.92 & 65.08 & --- & 97.54 \\
\cmidrule(lr){2-14}
& FENICE-0.75 & 38.48\textdagger\textdaggerdbl & 16.41\textdagger\textdaggerdbl & 30.23\textdagger\textdaggerdbl & 74.39\textdagger\textdaggerdbl & 59.02\textdagger\textdaggerdbl & 14.67 & \textbf{-19.55}*\textdagger\textdaggerdbl & \textbf{\underline{-8.17}}*\textdagger\textdaggerdbl & \textbf{\underline{87.22}}*\textdaggerdbl & --- & 27.51* & \textbf{97.69}*\textdagger\textdaggerdbl \\
& FIZZ-0.75 & 37.64\textdaggerdbl & 15.92\textdaggerdbl & 29.77\textdaggerdbl & 73.90\textdaggerdbl & 58.69\textdaggerdbl & 16.76 & -22.16*\textdaggerdbl & \textbf{-13.80}*\textdagger\textdaggerdbl & 86.35\textdaggerdbl & \textbf{\underline{65.54}}*\textdaggerdbl & --- & 97.57\textdaggerdbl \\
& MBR-1.0 & 38.13\textdagger\textdaggerdbl & 16.02\textdaggerdbl & 29.67 & 74.03\textdaggerdbl & 58.89\textdagger\textdaggerdbl & 11.86 & \textbf{\underline{-18.49}}*\textdagger\textdaggerdbl & --- & 86.17\textdaggerdbl & 55.52* & 22.90 & \textbf{\underline{97.72}}*\textdagger\textdaggerdbl \\
\cmidrule(lr){2-14}
& Oracle & \textit{54.10} & \textit{32.27} & \textit{47.60} & \textit{81.15} & \textit{63.31} & \textit{49.79} & \textit{-5.26} & \textit{13.82} & \textit{93.72} & \textit{87.59} & \textit{58.15} & \textit{98.65} \\
\midrule

\multirow{7}{*}{llm}& Baseline & 19.03 & 5.11 & \underline{13.19} & 61.83 & 53.73 & 0.84 & -6.02 & 8.72 & 93.08 & 88.04 & 22.11 & 99.89 \\
& FENICE-0.0 & 19.08 & 5.05 & \underline{13.19} & 61.83 & 53.72 & 0.82 & -5.31 & 9.37 & \underline{93.24} & --- & \underline{27.38} & 99.89 \\
& FIZZ-0.0 & 19.07 & 5.03 & \underline{13.19} & 61.84 & 53.72 & 0.78 & -5.63 & 8.46 & 92.99 & \underline{90.41} & --- & 99.89 \\
\cmidrule(lr){2-14}
& FENICE-0.75 & \textbf{\underline{19.09}} & \textbf{5.16}\textdagger\textdaggerdbl & 13.13 & \textbf{\underline{61.91}}*\textdagger\textdaggerdbl & \textbf{\underline{53.74}}\textdagger\textdaggerdbl & \textbf{1.42}*\textdagger\textdaggerdbl & \textbf{-3.38}*\textdagger\textdaggerdbl & \textbf{\underline{19.09}}*\textdagger\textdaggerdbl & 93.22*\textdaggerdbl & --- & 24.35* & 99.89*\textdagger\textdaggerdbl \\
& FIZZ-0.75 & \textbf{\underline{19.09}} & 5.06\textdaggerdbl & 13.18 & \textbf{61.87}\textdaggerdbl & 53.73 & \textbf{0.86}\textdaggerdbl & \textbf{-4.44}*\textdagger\textdaggerdbl & \textbf{13.40}*\textdagger\textdaggerdbl & 93.04 & 90.36* & --- & 99.89\textdaggerdbl \\
& MBR-1.0 & 19.03 & \textbf{\underline{5.17}}*\textdagger\textdaggerdbl & 13.05 & \textbf{61.90}*\textdagger\textdaggerdbl & \textbf{\underline{53.74}} & \textbf{\underline{1.72}}*\textdagger\textdaggerdbl & \textbf{\underline{-2.91}}*\textdagger\textdaggerdbl & --- & 93.14\textdaggerdbl & 88.42* & 21.42 & \textbf{\underline{99.90}}*\textdagger\textdaggerdbl \\
\cmidrule(lr){2-14}
& Oracle & \textit{23.69} & \textit{8.33} & \textit{17.04} & \textit{64.43} & \textit{54.63} & \textit{4.06} & \textit{-0.82} & \textit{32.85} & \textit{95.79} & \textit{96.49} & \textit{50.80} & \textit{99.93} \\

\bottomrule
\end{tabular}
}
\caption{Results on the test set for each metric on XSum dataset. \underline{Underline} indicates the highest scores for each metric and \textbf{bold} indicates better scores than all baselines. *, \textdagger, and \textdaggerdbl ~represent the statistical significance against Baseline, FENICE-0.0, and FIZZ-0.0, respectively (See \S\ref{subsec:exp_setting}). ``---'' indicates the skipped settings because the metrics used in the reranking and evaluation are identical. The abbreviations are the same as those in Table \ref{tab:cnndm_detailed_results}.}
\label{tab:xsum_detailed_results}
\end{table*}

\begin{table*}[t]
\centering
\footnotesize %
\setlength{\tabcolsep}{2pt} %
\resizebox{1.0\textwidth}{!}{

\begin{tabular}{@{}lll | c >{\columncolor[gray]{0.9}}c c >{\columncolor[gray]{0.9}}c c |>{\columncolor[gray]{0.9}}c c >{\columncolor[gray]{0.9}}c c >{\columncolor[gray]{0.9}}c c >{\columncolor[gray]{0.9}}c  @{}}
\toprule
& & & \multicolumn{5}{c|}{\textbf{Quality}} & \multicolumn{7}{c}{\textbf{Factuality}} \\
\cmidrule(lr){4-8} \cmidrule(lr){9-15}
\textbf{Setting} & \textbf{Model} & \textbf{Reranker} & \textbf{R1} & \textbf{R2} & \textbf{RL} & \textbf{BS} & \textbf{MS} & \textbf{EM} & \textbf{CM} & \textbf{SM} & \textbf{UE} & \textbf{Fe} & \textbf{Fi} & \textbf{SC} \\
\midrule

\multirow{21}{*}{epsilon} & \multirow{7}{*}{BART}& Baseline & 40.48 & 17.10 & 33.68 & 65.85 & 58.19 & 3.50 & -6.20 & 3.21 & 85.49 & 96.18 & 39.74 & \underline{99.84} \\
&   & FENICE-0.0 & 41.05 & 17.71 & 34.39 & 66.29 & 58.37 & 4.78 & -4.94 & 3.88 & \underline{87.99} & --- & \underline{55.42} & 99.81 \\
&   & FIZZ-0.0 & 41.07 & 18.06 & 34.59 & 66.15 & 58.32 & 4.02 & -5.07 & 4.07 & 87.56 & 98.68 & --- & 99.82 \\
\cmidrule(lr){3-15}
&   & FENICE-0.75 & \underline{41.38} & 17.88 & 34.67 & \underline{66.31} & \underline{58.50} & 8.39 & \underline{-3.31} & \underline{4.13} & 87.49 & --- & 51.78 & 99.80 \\
&   & FIZZ-0.75 & 41.27 & \underline{18.15} & \underline{34.77} & 66.25 & 58.41 & 5.89 & -4.02 & 4.02 & 87.65 & \underline{98.72} & --- & 99.81 \\
&   & MBR-1.0 & 41.20 & 17.66 & 34.47 & 66.12 & 58.46 & \underline{9.24} & -3.35 & --- & 86.53 & 97.64 & 46.55 & 99.80 \\
&   & Oracle & 52.49 & 28.94 & 45.59 & 71.74 & 61.58 & 19.70 & -0.39 & 42.04 & 94.88 & 99.81 & 83.51 & 99.94 \\
\cmidrule(lr){2-15}
 & \multirow{7}{*}{PEGASUS}& Baseline & 38.78 & 16.06 & 32.56 & 65.04 & 57.65 & 5.43 & -6.78 & -1.15 & 76.12 & 94.82 & 38.64 & \underline{99.67} \\
&   & FENICE-0.0 & 39.31 & 16.61 & 33.23 & \underline{65.46} & 57.76 & 7.07 & -5.42 & -0.31 & \underline{81.18} & --- & \underline{55.60} & 99.65 \\
&   & FIZZ-0.0 & 39.30 & 17.00 & 33.33 & 65.33 & 57.74 & 6.17 & -6.08 & -0.57 & 80.12 & 98.00 & --- & 99.65 \\
\cmidrule(lr){3-15}
&   & FENICE-0.75 & 39.37 & 16.58 & 33.20 & 65.28 & \underline{57.79} & 12.15 & -3.90 & \underline{0.30} & 78.41 & --- & 52.61 & 99.63 \\
&   & FIZZ-0.75 & \underline{39.43} & \underline{17.05} & \underline{33.45} & 65.33 & \underline{57.79} & 8.97 & -4.76 & -0.02 & 79.28 & \underline{98.04} & --- & 99.63 \\
&   & MBR-1.0 & 39.07 & 16.28 & 32.93 & 65.02 & 57.72 & \underline{12.99} & \underline{-3.86} & --- & 75.60 & 96.87 & 47.38 & 99.62 \\
&   & Oracle & 50.87 & 27.67 & 44.48 & 70.83 & 60.80 & 28.03 & -0.50 & 26.06 & 95.96 & 99.79 & 85.93 & 99.86 \\
\cmidrule(lr){2-15}
 & \multirow{7}{*}{T5-Large}& Baseline & 40.62 & 17.47 & 33.90 & 65.69 & 58.26 & 4.46 & -6.87 & 2.82 & 82.50 & 95.56 & 39.68 & \underline{99.71} \\
&   & FENICE-0.0 & 41.21 & 18.05 & 34.57 & \underline{66.32} & 58.45 & 6.01 & -5.28 & 2.83 & \underline{86.37} & --- & \underline{56.46} & 99.68 \\
&   & FIZZ-0.0 & 41.25 & 18.38 & 34.76 & 66.24 & 58.44 & 5.36 & -5.62 & \underline{3.66} & 86.26 & 98.48 & --- & 99.69 \\
\cmidrule(lr){3-15}
&   & FENICE-0.75 & 41.06 & 17.96 & 34.46 & 65.98 & 58.46 & 10.79 & -4.23 & 2.66 & 84.11 & --- & 52.94 & 99.66 \\
&   & FIZZ-0.75 & \underline{41.32} & \underline{18.40} & \underline{34.86} & 66.23 & \underline{58.50} & 8.20 & -4.60 & 3.32 & 85.54 & \underline{98.58} & --- & 99.67 \\
&   & MBR-1.0 & 40.80 & 17.74 & 34.21 & 65.66 & 58.39 & \underline{11.64} & \underline{-4.15} & --- & 82.08 & 97.54 & 47.85 & 99.66 \\
&   & Oracle & 52.21 & 28.74 & 45.38 & 71.49 & 61.49 & 23.73 & -0.51 & 36.94 & 94.76 & 99.79 & 83.21 & 99.88 \\
\midrule

\multirow{21}{*}{beam-sim} & \multirow{7}{*}{BART}& Baseline & 42.28 & 20.13 & 35.95 & 66.53 & 58.43 & 3.45 & -4.10 & 9.80 & 90.68 & 98.95 & \underline{66.36} & \underline{99.89} \\
&   & FENICE-0.0 & 42.29 & 19.41 & 35.81 & 66.75 & 58.56 & \underline{4.99} & -4.29 & 4.61 & 90.54 & --- & 64.17 & 99.81 \\
&   & FIZZ-0.0 & 42.03 & 19.54 & 35.72 & 66.49 & 58.42 & 4.24 & -4.40 & 5.51 & 90.08 & \underline{99.19} & --- & 99.83 \\
\cmidrule(lr){3-15}
&   & FENICE-0.75 & 43.09 & 20.19 & 36.33 & 67.06 & 58.69 & 4.01 & -3.43 & \underline{13.64} & \underline{90.89} & --- & 61.91 & 99.86 \\
&   & FIZZ-0.75 & 42.62 & 20.05 & 36.10 & 66.79 & 58.55 & 3.95 & -3.76 & 11.17 & 90.38 & 99.18 & --- & 99.85 \\
&   & MBR-1.0 & \underline{43.36} & \underline{20.44} & \underline{36.44} & \underline{67.15} & \underline{58.70} & 3.38 & \underline{-3.41} & --- & 90.42 & 98.64 & 56.62 & 99.88 \\
&   & Oracle & 52.91 & 30.04 & 46.50 & 71.94 & 61.65 & 18.40 & -0.42 & 37.09 & 95.65 & 99.85 & 89.75 & 99.93 \\
\cmidrule(lr){2-15}
 & \multirow{7}{*}{PEGASUS}& Baseline & 42.05 & \underline{19.80} & \underline{35.96} & 66.72 & \underline{58.22} & 5.23 & -5.69 & 3.48 & 93.10 & 98.28 & 57.32 & 99.64 \\
&   & FENICE-0.0 & 40.96 & 18.21 & 34.57 & 66.35 & 57.98 & \underline{7.43} & -5.44 & 1.12 & 93.09 & --- & \underline{60.75} & 99.56 \\
&   & FIZZ-0.0 & 41.10 & 18.66 & 34.90 & 66.29 & 57.96 & 6.22 & -5.65 & 2.33 & 92.47 & \underline{98.91} & --- & 99.59 \\
\cmidrule(lr){3-15}
&   & FENICE-0.75 & 42.21 & 19.28 & 35.38 & \underline{66.83} & 58.17 & 5.45 & -4.27 & \underline{9.92} & \underline{93.53} & --- & 55.83 & 99.67 \\
&   & FIZZ-0.75 & 41.79 & 19.20 & 35.31 & 66.60 & 58.10 & 5.71 & -4.78 & 7.09 & 93.00 & 98.87 & --- & 99.64 \\
&   & MBR-1.0 & \underline{42.33} & 19.30 & 35.27 & 66.82 & 58.13 & 4.55 & \underline{-3.96} & --- & \underline{93.53} & 98.03 & 48.71 & \underline{99.71} \\
&   & Oracle & 51.98 & 28.92 & 45.65 & 71.58 & 60.96 & 25.59 & -0.54 & 30.35 & 97.70 & 99.85 & 90.38 & 99.84 \\
\cmidrule(lr){2-15}
 & \multirow{7}{*}{T5-Large}& Baseline & \underline{24.78} & \underline{12.10} & \underline{23.01} & \underline{61.09} & 53.72 & 28.08 & -7.23 & -6.56 & 85.81 & 93.83 & 81.63 & 93.36 \\
&   & FENICE-0.0 & 23.83 & 10.20 & 21.64 & 60.74 & 53.69 & \underline{31.25} & -6.75 & -6.11 & 85.00 & --- & \underline{81.75} & 93.58 \\
&   & FIZZ-0.0 & 23.67 & 10.62 & 21.72 & 60.32 & 53.57 & 27.94 & -7.68 & -6.49 & 84.27 & 94.87 & --- & 93.43 \\
\cmidrule(lr){3-15}
&   & FENICE-0.75 & 24.20 & 10.43 & 21.97 & 61.02 & \underline{53.83} & 30.64 & \underline{-5.92} & \underline{-3.76} & \underline{86.01} & --- & 79.81 & \underline{93.59} \\
&   & FIZZ-0.75 & 24.01 & 10.70 & 22.00 & 60.65 & 53.71 & 29.90 & -6.45 & -4.48 & 85.05 & \underline{95.34} & --- & 93.46 \\
&   & MBR-1.0 & 23.98 & 10.21 & 21.71 & 60.67 & 53.79 & 28.15 & -5.96 & --- & 85.04 & 95.29 & 76.46 & \underline{93.59} \\
&   & Oracle & 35.13 & 22.08 & 33.24 & 66.19 & 56.01 & 75.43 & -0.46 & 2.94 & 95.00 & 99.89 & 97.94 & 96.09 \\
\midrule

\multirow{7}{*}{llm} & \multirow{7}{*}{LLaMA-3}& Baseline & 34.83 & 13.51 & 28.34 & 64.04 & 56.56 & 4.80 & -2.12 & 21.70 & 92.58 & 98.43 & 24.98 & 99.91 \\
&   & FENICE-0.0 & 35.15 & 13.77 & 28.66 & 64.20 & 56.64 & 5.53 & -2.05 & 21.19 & 92.63 & --- & \underline{31.12} & 99.90 \\
&   & FIZZ-0.0 & 35.26 & 13.80 & 28.74 & 64.23 & 56.66 & 5.13 & -2.27 & 20.92 & 92.55 & 98.89 & --- & 99.90 \\
\cmidrule(lr){3-15}
&   & FENICE-0.75 & 35.31 & \underline{14.11} & 28.71 & \underline{64.38} & 56.66 & 5.95 & -1.60 & \underline{31.14} & 92.91 & --- & 28.50 & 99.91 \\
&   & FIZZ-0.75 & \underline{35.36} & 13.98 & \underline{28.80} & 64.34 & \underline{56.68} & 5.34 & -1.95 & 25.08 & 92.70 & \underline{98.90} & --- & 99.90 \\
&   & MBR-1.0 & 35.15 & 14.07 & 28.52 & 64.36 & 56.63 & \underline{6.02} & \underline{-1.51} & --- & \underline{92.99} & 98.50 & 24.73 & \underline{99.92} \\
&   & Oracle & 40.88 & 18.83 & 34.05 & 67.03 & 57.93 & 14.81 & -0.45 & 49.41 & 95.63 & 99.81 & 55.71 & 99.95 \\

\bottomrule
\end{tabular}
}
\caption{Results on the test set divided by setting, model, and reranker for each metric on CNN/DM dataset. \underline{Underline} scores are the highest scores for each metric. ``---'' indicates the skipped settings because the metrics used in the reranking and evaluation are identical. The abbreviations are the same as those in Table \ref{tab:cnndm_detailed_results}.}
\label{tab:cnn_model_sett_rerank}
\end{table*}

\begin{table*}[t]
\centering
\footnotesize %
\setlength{\tabcolsep}{2pt} %
\resizebox{1.0\textwidth}{!}{

\begin{tabular}{@{}lll | c >{\columncolor[gray]{0.9}}c c >{\columncolor[gray]{0.9}}c c |>{\columncolor[gray]{0.9}}c c >{\columncolor[gray]{0.9}}c c >{\columncolor[gray]{0.9}}c c >{\columncolor[gray]{0.9}}c  @{}}
\toprule
& & & \multicolumn{5}{c|}{\textbf{Quality}} & \multicolumn{7}{c}{\textbf{Factuality}} \\
\cmidrule(lr){4-8} \cmidrule(lr){9-15}
\textbf{Setting} & \textbf{Model} & \textbf{Reranker} & \textbf{R1} & \textbf{R2} & \textbf{RL} & \textbf{BS} & \textbf{MS} & \textbf{EM} & \textbf{CM} & \textbf{SM} & \textbf{UE} & \textbf{Fe} & \textbf{Fi} & \textbf{SC} \\
\midrule

\multirow{21}{*}{epsilon} & \multirow{7}{*}{BART}& Baseline & 37.98 & 15.54 & 30.31 & 75.30 & 59.34 & 9.64 & -31.78 & -24.35 & 84.92 & 44.58 & 16.70 & \underline{98.83} \\
&   & FENICE-0.0 & \underline{38.60} & 16.10 & 30.89 & 75.66 & \underline{59.42} & 18.43 & -24.64 & -18.73 & \underline{88.31} & --- & \underline{28.45} & 98.75 \\
&   & FIZZ-0.0 & 38.00 & 15.56 & 30.42 & 75.30 & 59.23 & 15.08 & -27.75 & -20.89 & 87.31 & 64.41 & --- & 98.73 \\
\cmidrule(lr){3-15}
&   & FENICE-0.75 & 38.51 & \underline{16.11} & \underline{30.96} & \underline{75.79} & 59.41 & 27.98 & -20.43 & \underline{-17.02} & 88.22 & --- & 27.83 & 98.68 \\
&   & FIZZ-0.75 & 38.27 & 15.87 & 30.72 & 75.55 & 59.33 & 21.57 & -24.02 & -18.48 & 87.89 & \underline{66.55} & --- & 98.70 \\
&   & MBR-1.0 & 38.39 & 16.03 & 30.87 & 75.67 & 59.33 & \underline{28.79} & \underline{-18.04} & --- & 87.71 & 62.40 & 26.30 & 98.67 \\
&   & Oracle & 53.84 & 31.24 & 46.79 & 81.76 & 63.67 & 45.39 & -5.23 & 17.29 & 94.10 & 86.49 & 51.02 & 99.41 \\
\cmidrule(lr){2-15}
 & \multirow{7}{*}{PEGASUS}& Baseline & 40.38 & 18.00 & 32.79 & 76.38 & 60.03 & 12.80 & -28.28 & -19.67 & 86.17 & 45.23 & 16.40 & \underline{98.22} \\
&   & FENICE-0.0 & 40.63 & 18.23 & 32.96 & 76.56 & \underline{60.07} & 22.30 & -22.49 & -15.27 & 88.86 & --- & \underline{27.79} & 98.15 \\
&   & FIZZ-0.0 & 40.14 & 17.91 & 32.69 & 76.34 & 59.94 & 19.05 & -25.30 & -17.22 & 87.96 & 63.92 & --- & 98.11 \\
\cmidrule(lr){3-15}
&   & FENICE-0.75 & \underline{40.69} & \underline{18.50} & \underline{33.30} & \underline{76.82} & \underline{60.07} & 33.89 & -18.87 & \underline{-13.72} & \underline{88.87} & --- & 26.66 & 98.06 \\
&   & FIZZ-0.75 & 40.55 & 18.30 & 33.13 & 76.63 & 60.03 & 27.26 & -21.51 & -15.30 & 88.48 & \underline{65.99} & --- & 98.07 \\
&   & MBR-1.0 & 40.60 & 18.44 & 33.22 & 76.79 & 60.01 & \underline{35.76} & \underline{-16.47} & --- & 88.52 & 62.84 & 25.22 & 98.04 \\
&   & Oracle & 56.42 & 34.49 & 49.72 & 82.90 & 64.74 & 52.59 & -4.94 & 21.69 & 94.16 & 85.49 & 48.99 & 99.16 \\
\cmidrule(lr){2-15}
 & \multirow{7}{*}{T5-Large}& Baseline & 36.45 & 14.19 & 28.71 & 73.91 & 58.78 & 6.07 & -33.41 & -26.45 & 84.14 & 47.21 & 17.64 & \underline{98.43} \\
&   & FENICE-0.0 & 36.78 & 14.54 & 29.05 & 74.19 & 58.83 & 11.51 & -25.48 & -19.85 & \underline{87.22} & --- & \underline{30.28} & 98.34 \\
&   & FIZZ-0.0 & 36.20 & 14.08 & 28.56 & 73.78 & 58.66 & 9.37 & -28.60 & -22.19 & 86.09 & 65.77 & --- & 98.32 \\
\cmidrule(lr){3-15}
&   & FENICE-0.75 & \underline{36.84} & \underline{14.72} & \underline{29.28} & \underline{74.38} & \underline{58.87} & 19.27 & -21.78 & \underline{-18.47} & 86.94 & --- & 28.90 & 98.22 \\
&   & FIZZ-0.75 & 36.42 & 14.34 & 28.89 & 74.04 & 58.74 & 14.25 & -25.20 & -20.43 & 86.51 & \underline{67.51} & --- & 98.26 \\
&   & MBR-1.0 & 36.72 & 14.66 & 29.23 & 74.30 & 58.83 & \underline{20.04} & \underline{-19.71} & --- & 86.38 & 63.83 & 27.10 & 98.21 \\
&   & Oracle & 51.21 & 28.33 & 43.86 & 80.23 & 62.69 & 32.80 & -5.78 & 14.35 & 94.06 & 87.46 & 53.30 & 99.07 \\
\midrule

\multirow{21}{*}{beam-sim} & \multirow{7}{*}{BART}& Baseline & \underline{43.15} & \underline{20.78} & \underline{35.41} & \underline{77.40} & \underline{60.54} & 14.36 & -29.07 & -20.56 & 88.69 & 55.14 & 22.07 & 98.73 \\
&   & FENICE-0.0 & 39.40 & 16.83 & 30.87 & 75.71 & 59.40 & \underline{18.16} & -24.03 & -17.56 & 89.19 & --- & \underline{29.87} & 98.74 \\
&   & FIZZ-0.0 & 38.89 & 16.45 & 30.57 & 75.45 & 59.29 & 15.27 & -26.75 & -19.99 & 88.56 & 65.88 & --- & 98.73 \\
\cmidrule(lr){3-15}
&   & FENICE-0.75 & 39.66 & 16.86 & 30.79 & 75.73 & 59.57 & 11.90 & -21.36 & \underline{-9.21} & \underline{89.36} & --- & 23.94 & 98.88 \\
&   & FIZZ-0.75 & 39.30 & 16.69 & 30.82 & 75.61 & 59.41 & 14.88 & -24.72 & -15.93 & 88.88 & \underline{66.34} & --- & 98.76 \\
&   & MBR-1.0 & 39.28 & 16.43 & 30.17 & 75.37 & 59.45 & 8.85 & \underline{-20.32} & --- & 88.37 & 55.04 & 19.34 & \underline{98.94} \\
&   & Oracle & 54.90 & 32.52 & 48.12 & 82.20 & 63.90 & 45.34 & -6.24 & 15.68 & 94.33 & 86.54 & 52.85 & 99.38 \\
\cmidrule(lr){2-15}
 & \multirow{7}{*}{PEGASUS}& Baseline & \underline{45.30} & \underline{23.38} & \underline{37.88} & \underline{78.48} & \underline{61.32} & 20.29 & -25.77 & -16.05 & 88.76 & 54.43 & 21.57 & 98.01 \\
&   & FENICE-0.0 & 40.86 & 18.54 & 32.36 & 76.47 & 59.91 & \underline{22.79} & -21.66 & -14.40 & \underline{89.54} & --- & \underline{29.17} & 98.13 \\
&   & FIZZ-0.0 & 40.45 & 18.19 & 32.12 & 76.30 & 59.83 & 20.49 & -23.57 & -15.36 & 88.89 & 65.48 & --- & 98.07 \\
\cmidrule(lr){3-15}
&   & FENICE-0.75 & 41.73 & 19.04 & 32.84 & 76.77 & 60.28 & 15.07 & -19.67 & \underline{-5.52} & 89.51 & --- & 22.50 & 98.28 \\
&   & FIZZ-0.75 & 40.87 & 18.48 & 32.36 & 76.46 & 59.98 & 19.85 & -21.67 & -11.24 & 89.15 & \underline{65.51} & --- & 98.10 \\
&   & MBR-1.0 & 41.48 & 18.64 & 32.35 & 76.54 & 60.20 & 11.54 & \underline{-18.78} & --- & 88.73 & 53.34 & 17.97 & \underline{98.33} \\
&   & Oracle & 57.37 & 35.66 & 50.90 & 83.28 & 64.95 & 53.17 & -5.81 & 19.47 & 94.35 & 85.62 & 51.05 & 99.07 \\
\cmidrule(lr){2-15}
 & \multirow{7}{*}{T5-Large}& Baseline & \underline{36.87} & \underline{16.47} & \underline{30.64} & \underline{71.77} & \underline{57.75} & 15.20 & -26.34 & -20.05 & 81.35 & 54.66 & 31.52 & \underline{96.02} \\
&   & FENICE-0.0 & 33.64 & 13.26 & 26.88 & 70.57 & 57.03 & \underline{18.59} & -20.46 & -16.22 & 82.68 & --- & \underline{40.64} & 95.93 \\
&   & FIZZ-0.0 & 32.46 & 12.40 & 25.91 & 69.48 & 56.61 & 14.62 & -22.89 & -18.47 & 80.32 & 63.89 & --- & 95.83 \\
\cmidrule(lr){3-15}
&   & FENICE-0.75 & 34.06 & 13.33 & 27.05 & 70.68 & 57.21 & 17.03 & -17.60 & \underline{-9.78} & \underline{82.80} & --- & 36.10 & 95.91 \\
&   & FIZZ-0.75 & 32.76 & 12.60 & 26.13 & 69.62 & 56.68 & 15.55 & -20.08 & -14.22 & 81.01 & \underline{64.77} & --- & 95.85 \\
&   & MBR-1.0 & 33.64 & 13.00 & 26.50 & 70.17 & 57.04 & 15.18 & \underline{-16.36} & --- & 81.42 & 58.18 & 31.38 & 95.89 \\
&   & Oracle & 50.02 & 28.62 & 43.79 & 77.99 & 61.06 & 50.85 & -3.74 & 6.31 & 92.50 & 90.63 & 70.57 & 97.50 \\
\midrule

\multirow{7}{*}{llm} & \multirow{7}{*}{LLaMA-3}& Baseline & 19.03 & 5.11 & \underline{13.19} & 61.83 & 53.73 & 0.84 & -6.02 & 8.72 & 93.08 & 88.04 & 22.11 & 99.89 \\
&   & FENICE-0.0 & 19.08 & 5.05 & \underline{13.19} & 61.83 & 53.72 & 0.82 & -5.31 & 9.37 & \underline{93.24} & --- & \underline{27.38} & 99.89 \\
&   & FIZZ-0.0 & 19.07 & 5.03 & \underline{13.19} & 61.84 & 53.72 & 0.78 & -5.63 & 8.46 & 92.99 & \underline{90.41} & --- & 99.89 \\
\cmidrule(lr){3-15}
&   & FENICE-0.75 & \underline{19.09} & 5.16 & 13.13 & \underline{61.91} & \underline{53.74} & 1.42 & -3.38 & \underline{19.09} & 93.22 & --- & 24.35 & 99.89 \\
&   & FIZZ-0.75 & \underline{19.09} & 5.06 & 13.18 & 61.87 & 53.73 & 0.86 & -4.44 & 13.40 & 93.04 & 90.36 & --- & 99.89 \\
&   & MBR-1.0 & 19.03 & \underline{5.17} & 13.05 & 61.90 & \underline{53.74} & \underline{1.72} & \underline{-2.91} & --- & 93.14 & 88.42 & 21.42 & \underline{99.90} \\
&   & Oracle & 23.69 & 8.33 & 17.04 & 64.43 & 54.63 & 4.06 & -0.82 & 32.85 & 95.79 & 96.49 & 50.80 & 99.93 \\

\bottomrule
\end{tabular}
}
\caption{Results on the test set divided by setting, model, and reranker for each metric on XSum dataset. \underline{Underline} scores are the highest scores for each metric. ``---'' indicates the skipped settings because the metrics used in the reranking and evaluation are identical. The abbreviations are the same as those in Table \ref{tab:cnndm_detailed_results}.}
\label{tab:xsum_model_sett_rerank}
\end{table*}

\begin{table*}[t]
\centering
\footnotesize %
\setlength{\tabcolsep}{2pt} %
\resizebox{1.0\textwidth}{!}{

\begin{tabular}{@{}ll | c >{\columncolor[gray]{0.9}}c c >{\columncolor[gray]{0.9}}c c |>{\columncolor[gray]{0.9}}c c >{\columncolor[gray]{0.9}}c c >{\columncolor[gray]{0.9}}c c >{\columncolor[gray]{0.9}}c  @{}}
\toprule
& & \multicolumn{5}{c|}{\textbf{Quality}} & \multicolumn{7}{c}{\textbf{Factuality}} \\
\cmidrule(lr){3-7} \cmidrule(lr){8-14}
\textbf{Setting} & \textbf{Model} & \textbf{R1} & \textbf{R2} & \textbf{RL} & \textbf{BS} & \textbf{MS} & \textbf{EM} & \textbf{CM} & \textbf{SM} & \textbf{UE} & \textbf{Fe} & \textbf{Fi} & \textbf{SC} \\
\midrule

\multirow{3}{*}{epsilon}& BART & 42.71 & 19.36 & 36.02 & 66.96 & 58.84 & 7.93 & -3.90 & 9.31 & 88.23 & 98.64 & 63.13 & 99.83 \\
& PEGASUS & 40.88 & 18.18 & 34.74 & 66.04 & 58.18 & 11.54 & -4.47 & 3.45 & 80.95 & 98.12 & 64.26 & 99.67 \\
& T5-Large & 42.64 & 19.53 & 36.02 & 66.80 & \underline{58.86} & 10.03 & -4.47 & 7.79 & 85.95 & 98.47 & 63.40 & 99.71 \\
\midrule

\multirow{3}{*}{beam-div}& BART & 43.17 & 20.99 & 36.99 & 67.12 & 58.73 & 3.89 & -3.99 & 8.68 & 91.59 & 99.21 & 72.44 & 99.85 \\
& PEGASUS & 42.71 & 20.90 & 36.79 & 67.00 & 58.37 & 6.03 & -4.62 & 6.68 & \textbf{93.80} & 99.07 & 70.79 & 99.66 \\
& T5-Large & 25.41 & 12.62 & 23.65 & 61.63 & 53.87 & 32.07 & -6.21 & -5.31 & 87.02 & 96.83 & 87.25 & 93.51 \\
\midrule

\multirow{3}{*}{beam-dbl}& BART & \underline{43.44} & \underline{21.13} & \underline{37.12} & 67.22 & 58.77 & 4.33 & -3.67 & 10.82 & 91.76 & \textbf{99.36} & 75.26 & \underline{99.86} \\
& PEGASUS & 42.66 & 20.77 & 36.58 & 66.95 & 58.30 & 6.57 & -4.31 & 9.34 & 93.61 & 99.16 & 73.52 & 99.69 \\
& T5-Large & 25.52 & 12.55 & 23.67 & 61.65 & 53.94 & \underline{33.57} & -5.87 & -4.83 & 87.14 & 97.08 & \textbf{88.11} & 93.64 \\
\midrule

\multirow{3}{*}{beam-sim}& BART & \textbf{44.08} & \textbf{21.40} & \textbf{37.55} & \textbf{67.53} & \textbf{59.00} & 6.06 & -3.40 & \underline{14.27} & 91.23 & \underline{99.35} & 73.84 & \underline{99.86} \\
& PEGASUS & 43.20 & 20.48 & 36.72 & \underline{67.31} & 58.50 & 8.60 & -4.33 & 9.58 & \underline{93.77} & 99.07 & 70.26 & 99.66 \\
& T5-Large & 25.66 & 12.34 & 23.61 & 61.53 & 54.04 & \textbf{35.91} & -5.78 & -3.99 & 86.60 & 96.99 & \underline{87.56} & 93.87 \\
\midrule

\multirow{1}{*}{llm}& LLaMA-3 & 35.99 & 14.58 & 29.40 & 64.65 & 56.83 & 6.80 & \textbf{-1.71} & \textbf{29.13} & 93.14 & 99.15 & 39.35 & \textbf{99.91} \\

\bottomrule
\end{tabular}
}
\caption{Results on the test set divided by setting, model, and reranker for each metric on CNN/DM dataset. \textbf{Bold} and \underline{Underline} represent the highest and the second highest scores, respectively. The abbreviations are the same as those in Table \ref{tab:cnndm_detailed_results}.}
\label{tab:cnn_model_sett}
\end{table*}

\begin{table*}[t]
\centering
\footnotesize %
\setlength{\tabcolsep}{2pt} %
\resizebox{1.0\textwidth}{!}{

\begin{tabular}{@{}ll | c >{\columncolor[gray]{0.9}}c c >{\columncolor[gray]{0.9}}c c |>{\columncolor[gray]{0.9}}c c >{\columncolor[gray]{0.9}}c c >{\columncolor[gray]{0.9}}c c >{\columncolor[gray]{0.9}}c  @{}}
\toprule
& & \multicolumn{5}{c|}{\textbf{Quality}} & \multicolumn{7}{c}{\textbf{Factuality}} \\
\cmidrule(lr){3-7} \cmidrule(lr){8-14}
\textbf{Setting} & \textbf{Model} & \textbf{R1} & \textbf{R2} & \textbf{RL} & \textbf{BS} & \textbf{MS} & \textbf{EM} & \textbf{CM} & \textbf{SM} & \textbf{UE} & \textbf{Fe} & \textbf{Fi} & \textbf{SC} \\
\midrule

\multirow{3}{*}{epsilon}& BART & 40.51 & 18.06 & 32.99 & 76.43 & 59.96 & \underline{23.84} & -21.70 & -13.99 & 88.35 & 70.71 & 35.67 & 98.82 \\
& PEGASUS & 42.77 & 20.55 & 35.40 & 77.49 & 60.70 & \textbf{29.09} & -19.69 & -10.33 & 89.00 & 70.29 & 34.33 & 98.26 \\
& T5-Large & 38.66 & 16.41 & 31.08 & 74.98 & 59.34 & 16.19 & -22.85 & -15.85 & 87.33 & 72.05 & 37.30 & 98.41 \\
\midrule

\multirow{3}{*}{beam-div}& BART & 43.90 & 21.57 & 36.24 & 77.75 & 60.71 & 17.27 & -26.80 & -17.82 & 89.47 & 61.13 & 27.12 & 98.74 \\
& PEGASUS & \textbf{46.26} & \textbf{24.42} & \textbf{38.80} & \textbf{78.64} & \textbf{61.51} & 22.73 & -22.54 & -11.40 & 89.50 & 63.62 & 28.01 & 98.12 \\
& T5-Large & 37.80 & 17.27 & 31.54 & 72.15 & 57.85 & 17.98 & -23.10 & -16.76 & 82.59 & 64.27 & 38.96 & 96.04 \\
\midrule

\multirow{3}{*}{beam-dbl}& BART & 43.80 & 21.32 & 36.00 & 77.66 & 60.65 & 18.41 & -24.85 & -15.07 & 89.81 & 66.19 & 30.98 & 98.77 \\
& PEGASUS & \underline{45.42} & \underline{23.50} & \underline{37.69} & \underline{78.15} & \underline{61.22} & 23.12 & -20.59 & -8.69 & 89.59 & 67.87 & 31.67 & 98.20 \\
& T5-Large & 37.52 & 16.94 & 31.15 & 72.05 & 57.80 & 19.36 & -20.79 & -14.41 & 83.16 & 69.27 & \underline{44.17} & 96.09 \\
\midrule

\multirow{3}{*}{beam-sim}& BART & 42.08 & 19.51 & 33.82 & 76.78 & 60.22 & 18.40 & -21.78 & -10.57 & 89.63 & 71.47 & 36.07 & \underline{98.88} \\
& PEGASUS & 44.01 & 21.70 & 35.83 & 77.76 & 60.92 & 23.31 & -19.56 & -6.50 & \underline{89.85} & 70.56 & 34.73 & 98.28 \\
& T5-Large & 36.21 & 15.67 & 29.56 & 71.47 & 57.63 & 21.00 & -18.21 & -11.41 & 83.15 & \underline{73.10} & \textbf{49.99} & 96.13 \\
\midrule

\multirow{1}{*}{llm}& LLaMA-3 & 19.73 & 5.56 & 13.71 & 62.23 & 53.86 & 1.50 & \textbf{-4.07} & \textbf{16.24} & \textbf{93.50} & \textbf{92.32} & 35.22 & \textbf{99.90} \\

\bottomrule
\end{tabular}
}
\caption{Results on the test set divided by setting, model, and reranker for each metric on XSum dataset. \textbf{Bold} and \underline{Underline} represent the highest and the second highest scores, respectively. The abbreviations are the same as those in Table \ref{tab:cnndm_detailed_results}.}
\label{tab:xsum_model_sett}
\end{table*}

\begin{table*}[ht]
    \centering
    \footnotesize %
    \setlength{\tabcolsep}{2pt} %
    \begin{tabular}{p{1.0\textwidth}}
    \toprule
    \textbf{Source} \\
    (CNN)A photo of a baby boy being pulled from the rubble of the Nepal earthquake has become the defining image of a disaster that has devastated the country. His tiny face is coated in dust from the debris that crashed around him as the earth shook on Saturday, April 25, killing more than 5,000 people and injuring at least twice as many more. His name is Sonit Awal, and he's just four months old. On Saturday, Sonit was at home in Muldhoka, Bhaktapur, east of the capital Kathmandu, when the 7.8-magnitude quake sent tremors through Nepal and neighboring countries, according to Kathmandu Today. The house collapsed, burying the child, leaving his father Shyam Awal frantically searching for him amid the rubble. A desperate Awal called the Nepalese Army, and they too scrambled through the dirt and debris until midnight, before giving up hope and leaving. Sonit's father had lost all hope of finding his son alive, the paper reported, until he heard faint cries from below. On Sunday morning, the soldiers returned and at 10 a.m they lifted Sonit clear of the fallen rocks, beams, bricks and dust that had trapped him for 22 hours. His rescuers held him high in their hands to the sun, though his eyes were still wedged tightly shut. Sonit's face was exposed during the whole ordeal, though a hooded top and shawl protected his head and body during the cold night spent under rubble. The baby was taken to Bhaktapur Hospital and found to be uninjured; . His unlikely rescue provides some hope to a country that has experienced so much loss. \\
    \midrule
    \textbf{Before MBR Summary} 
    \par\vspace{1em}
    \textbf{Baseline Summary:} \par
    Sonit Awal was buried under rubble when his home collapsed in the Nepal earthquake. His father called the army, who spent 22 hours searching for him before giving up hope. The baby was rescued Sunday morning and found to be uninjured. The 7.8-magnitude quake killed more than 5,000 people and injured at least twice as many more. \par \vspace{1em}
    \textbf{FIZZ-0.0 Summary:} \par
    Sonit Awal, 4 months, was buried under rubble when his home collapsed. His father frantically searched for him for 22 hours before calling the army. The soldiers pulled him from the rubble on Sunday morning. He was uninjured. The 7.8-magnitude quake killed more than 5,000 people and injured at least twice as many. \par \vspace{1em}
    \textbf{FENICE-0.0 Summary:} \par
    Baby Sonit Awal was trapped under rubble for 22 hours after his house collapsed in Nepal's earthquake. His father called the army, which spent the night searching for him. The baby was rescued and taken to a hospital, uninjured. More than 5,000 people were killed in the earthquake, which struck on Saturday. \\
    \midrule
    \textbf{After MBR Summary}
    \par\vspace{1em}
    \textbf{MBR-1.0 Summary:} \par
    Sonit Awal, 4 months old, was buried under rubble when Nepal's house collapsed. The baby's father frantically searched for him for 22 hours before calling the army. Sonit was rescued Sunday morning, uninjured, and taken to a hospital. The 7.8-magnitude quake killed more than 5,000 people and injured at least twice as many more. \par \vspace{1em}
    \textbf{FIZZ-0.75 Summary:} \par
    Sonit Awal, 4 months, was buried under rubble when his home collapsed. His father frantically searched for him for 22 hours before calling the army. The soldiers pulled him from the rubble on Sunday morning. He was uninjured. The 7.8-magnitude quake killed more than 5,000 people and injured at least twice as many. \par \vspace{1em}
    \textbf{FENICE-0.75 Summary:} \par
    Baby Sonit Awal was trapped under rubble for 22 hours after his house collapsed in Nepal's earthquake. His father called the army, which spent the night searching for him. The baby was rescued and taken to a hospital, uninjured. More than 5,000 people were killed in the earthquake, which struck on Saturday. \\
    \bottomrule
    \end{tabular}
    \caption{Example CNN/DM dataset summaries generated by BART model using the beam-sim setting. }
    \label{tab:example_cnn_beam_bart}
\end{table*}

\begin{table*}[ht]
    \centering
    \footnotesize %
    \setlength{\tabcolsep}{2pt} %
    \begin{tabular}{p{1.0\textwidth}}
    \toprule
    \textbf{Source} \\
    The mix-up led to the body of Philip Bradburn being cremated instead of that of Conservative MEP Philip Bradbourn. A failure to provide written records created confusion between the two similar sounding surnames, the Heart of England NHS Trust said. The trust and Central England Co-operative apologised for the blunder. Former MEP Mr Bradbourn died at Good Hope Hospital, Sutton Coldfield on 20 December. Mr Bradburn died at University Hospital Birmingham four days later and his body was sent to a funeral directors run by Central England Co-operative. As Good Hope's mortuary was nearing capacity, a request was made by hospital staff to move four bodies - including Mr Bradbourn's - to the funeral company. The names of those to be moved were given over the phone but not followed up with an email listing their names and addresses, the report found. Updates on this story and more from Birmingham and the Black Country. The undertaker collected Mr Bradbourn's body, but four days later returned it to Good Hope, when his family requested to change his burial to a cremation. Meanwhile, Mr Bradburn's body was sent back to the hospital over concerns it had been at the funeral directors for some time. When it arrived at Good Hope from the funeral directors, paperwork carrying his surname was overwritten with the surname of the politician. Doctors looked at these papers and signed off the cremation for Mr Bradbourn but the body of Mr Bradburn was released. Central England Co-operative said: "Our priority has been to work closely with all concerned so that we can learn from this unfortunate incident and build further appropriate safeguards for the future." Dr Andrew Catto, of Heart of England NHS Foundation Trust said there had been a "rare and complex set of circumstances". He said: "We are very sorry that this incredibly distressing situation has happened." \\
    \midrule
    \textbf{Before MBR Summary} 
    \par\vspace{1em}
    \textbf{Baseline Summary:} \par
    An NHS trust has apologised after the body of a former Conservative MEP was mistakenly sent to the funeral of a man with the same surname. \par \vspace{1em}
    \textbf{FIZZ-0.0 Summary:} \par
    Staff at a Birmingham hospital made a "distressing mistake" when they signed off the cremation of a politician's body, a report has found. \par \vspace{1em}
    \textbf{FENICE-0.0 Summary:} \par
    An NHS trust and funeral company have apologised after the body of a former Conservative MEP was mistakenly sent to the funeral of a man with the same surname. \\
    \midrule
    \textbf{After MBR Summary}
    \par\vspace{1em}
    \textbf{MBR-1.0 Summary:} \par
    An MP's body was sent to a funeral company for cremation when hospital staff mistook it for another man's, a report has found. \par \vspace{1em}
    \textbf{FIZZ-0.75 Summary:} \par
    An MP's body was sent to a funeral company for cremation when hospital staff mistook it for another man's, a report has found. \par \vspace{1em}
    \textbf{FENICE-0.75 Summary:} \par
    An MP's body was sent to a funeral company for cremation when hospital staff mistook it for another man's, a report has found. \\
    \bottomrule
    \end{tabular}
    \caption{Example XSUM dataset summaries generated by BART model using the beam-sim setting. }
    \label{tab:example_xsum_beam_bart}
\end{table*}

\begin{table*}[ht]
    \centering
    \footnotesize %
    \setlength{\tabcolsep}{2pt} %
    \begin{tabular}{p{1.0\textwidth}}
    \toprule
    \textbf{Source} \\
    West Ham wonderkid Reece Oxford pushed himself closer to a long-awaited first-team debut after scoring for the Under 21s against Manchester United. The 16-year-old, who has been on the bench for the senior side this season, scored with a header in the Hammers 3-2 defeat against the Red Devils on Tuesday night. The goal will push Oxford's claims for a senior debut before the end of the season. Reece Oxford, pictured here earlier in the season, has moved closer to a first-team debut for West Ham . Oxford scored for West Ham's under 21 team in their 3-2 defeat to Manchester United on Tuesday night . Oxford signed his first professional contract earlier this season, snubbing interest from all of the Premier League's big guns. The central defender, who has been likened to Rio Ferdinand, has tipped to be a future England international. Oxford regularly trains with the first team and is hopeful Sam Allardyce will give him his first team debut before the end of campaign. \\
    \midrule
    \textbf{Before MBR Summary} 
    \par\vspace{1em}
    \textbf{Baseline Summary:} \par
    Reece Oxford, a 16-year-old defender for West Ham, scored a header for the Under 21 team in a 3-2 loss to Manchester United, bringing him closer to a potential first-team debut. Oxford has already been on the bench for the senior side this season and signed his first professional contract, turning down interest from top Premier League clubs. He regularly trains with the first team and hopes to make his debut under manager Sam Allardyce before the end of the season. \par \vspace{1em}
    \textbf{FIZZ-0.0 Summary:} \par
    Reece Oxford, a 16-year-old West Ham player, scored a header for the Under 21 team in a 3-2 loss to Manchester United, increasing his chances of making his senior debut before the end of the season. Oxford has been training with the first team and has signed his first professional contract, snubbing interest from top Premier League clubs. He is expected to be a future England international and hopes to make his debut under manager Sam Allardyce. \par \vspace{1em}
    \textbf{FENICE-0.0 Summary:} \par
    Reece Oxford, a 16-year-old West Ham defender, scored a goal for the Under 21 team in a 3-2 loss to Manchester United, bringing him closer to a potential first-team debut. Oxford has already been on the bench for the senior team this season and has been training with the first team. He signed his first professional contract earlier this season, despite interest from other Premier League teams, and is tipped to be a future England international. \\
    \midrule
    \textbf{After MBR Summary}
    \par\vspace{1em}
    \textbf{MBR-1.0 Summary:} \par
    Reece Oxford, a 16-year-old West Ham defender, scored a header for the Under 21 team in a 3-2 loss to Manchester United, bringing him closer to a potential first-team debut. Oxford has already been on the senior team's bench this season and has signed his first professional contract, snubbing interest from top Premier League clubs. He has been tipped to become a future England international and hopes to make his first-team debut under manager Sam Allardyce before the end of the season. \par \vspace{1em}
    \textbf{FIZZ-0.75 Summary:} \par
    Reece Oxford, a 16-year-old West Ham player, scored a header for the Under 21 team in a 3-2 loss to Manchester United, increasing his chances of making his senior debut before the end of the season. Oxford has been training with the first team and has signed his first professional contract, snubbing interest from top Premier League clubs. He is expected to be a future England international and hopes to make his debut under manager Sam Allardyce. \par \vspace{1em}
    \textbf{FENICE-0.75 Summary:} \par
    Reece Oxford, a 16-year-old West Ham defender, scored a goal for the Under 21 team in a 3-2 loss to Manchester United, bringing him closer to a potential first-team debut. Oxford has already been on the bench for the senior team this season and has been training with the first team. He signed his first professional contract earlier this season, despite interest from other Premier League teams, and is tipped to be a future England international. \\
    \bottomrule
    \end{tabular}
    \caption{Example CNN/DM summaries generated by Llama-3 model using the llm setting. }
    \label{tab:example_cnn_llm}
\end{table*}

\begin{table*}[ht]
    \centering
    \footnotesize %
    \setlength{\tabcolsep}{2pt} %
    \begin{tabular}{p{1.0\textwidth}}
    \toprule
    \textbf{Source} \\
    The mix-up led to the body of Philip Bradburn being cremated instead of that of Conservative MEP Philip Bradbourn. A failure to provide written records created confusion between the two similar sounding surnames, the Heart of England NHS Trust said. The trust and Central England Co-operative apologised for the blunder. Former MEP Mr Bradbourn died at Good Hope Hospital, Sutton Coldfield on 20 December. Mr Bradburn died at University Hospital Birmingham four days later and his body was sent to a funeral directors run by Central England Co-operative. As Good Hope's mortuary was nearing capacity, a request was made by hospital staff to move four bodies - including Mr Bradbourn's - to the funeral company. The names of those to be moved were given over the phone but not followed up with an email listing their names and addresses, the report found. Updates on this story and more from Birmingham and the Black Country. The undertaker collected Mr Bradbourn's body, but four days later returned it to Good Hope, when his family requested to change his burial to a cremation. Meanwhile, Mr Bradburn's body was sent back to the hospital over concerns it had been at the funeral directors for some time. When it arrived at Good Hope from the funeral directors, paperwork carrying his surname was overwritten with the surname of the politician. Doctors looked at these papers and signed off the cremation for Mr Bradbourn but the body of Mr Bradburn was released. Central England Co-operative said: "Our priority has been to work closely with all concerned so that we can learn from this unfortunate incident and build further appropriate safeguards for the future." Dr Andrew Catto, of Heart of England NHS Foundation Trust said there had been a "rare and complex set of circumstances". He said: "We are very sorry that this incredibly distressing situation has happened." \\
    \midrule
    \textbf{Before MBR Summary} 
    \par\vspace{1em}
    \textbf{Baseline Summary:} \par
    The Heart of England NHS Trust and Central England Co-operative apologized for a mix-up that led to the cremation of the wrong body. The mistake occurred when the names of two men with similar surnames, Philip Bradburn and Philip Bradbourn, were not properly recorded. Bradburn's body was sent to a funeral director, but was later returned to the hospital due to concerns it had been there too long. Meanwhile, Bradbourn's body was cremated, despite his family requesting a burial. The trust and co-operative are working to learn from the incident and implement additional safeguards to prevent similar mistakes in the future. \par \vspace{1em}
    \textbf{FIZZ-0.0 Summary:} \par
    A mix-up led to the cremation of Philip Bradburn instead of Conservative MEP Philip Bradbourn due to a failure to provide written records and similar sounding surnames. The mistake occurred when four bodies, including Bradbourn's, were moved from Good Hope Hospital to a funeral directors run by Central England Co-operative. The undertaker collected Bradbourn's body, but it was later returned to the hospital when his family requested a cremation. Meanwhile, Bradburn's body was sent back to the hospital, and paperwork was overwritten with Bradbourn's surname, leading to the cremation of Bradburn's body. Both the Heart of England NHS Trust and Central England Co-operative have apologized for the blunder. \par \vspace{1em}
    \textbf{FENICE-0.0 Summary:} \par
    The Heart of England NHS Trust and Central England Co-operative apologized for a mistake that led to the cremation of Philip Bradburn's body instead of Conservative MEP Philip Bradbourn's. The mix-up occurred due to a failure to provide written records, resulting in confusion between the two similar-sounding surnames. Bradbourn's body was moved to a funeral director's and later returned to the hospital, while Bradburn's body was sent to the funeral director's and eventually cremated. The incident was described as a "rare and complex set of circumstances" and both organizations are working to learn from the mistake and implement further safeguards. \\
    \midrule
    \textbf{After MBR Summary}
    \par\vspace{1em}
    \textbf{MBR-1.0 Summary:} \par
    The Heart of England NHS Trust and Central England Co-operative apologized for a mistake that led to the cremation of Philip Bradburn's body instead of Conservative MEP Philip Bradbourn's. The mix-up occurred due to a failure to provide written records, with similar sounding surnames contributing to the confusion. Bradburn died on December 24, and his body was sent to a funeral director, while Bradbourn died on December 20. The mistake was only discovered when Bradbourn's family requested a change from burial to cremation, and it was found that Bradburn's body had been cremated instead. \par \vspace{1em}
    \textbf{FIZZ-0.75 Summary:} \par
    A mix-up led to the cremation of Philip Bradburn instead of Conservative MEP Philip Bradbourn due to a failure to provide written records and similar sounding surnames. The mistake occurred when four bodies, including Bradbourn's, were moved from Good Hope Hospital to a funeral directors run by Central England Co-operative. The undertaker collected Bradbourn's body, but it was later returned to the hospital when his family requested a cremation. Meanwhile, Bradburn's body was sent back to the hospital, and paperwork was overwritten with Bradbourn's surname, leading to the cremation of Bradburn's body. Both the Heart of England NHS Trust and Central England Co-operative have apologized for the blunder. \par \vspace{1em}
    \textbf{FENICE-0.75 Summary:} \par
    The Heart of England NHS Trust and Central England Co-operative apologized for a mistake that led to the cremation of Philip Bradburn's body instead of Conservative MEP Philip Bradbourn's. The error occurred due to a failure to provide written records and confusion between the two similar-sounding surnames. Bradbourn's body was moved to a funeral director's, but his family later requested a cremation, while Bradburn's body was mistakenly released for cremation. The trust and co-operative are working to learn from the incident and implement additional safeguards to prevent similar mistakes in the future. \\
    \bottomrule
    \end{tabular}
    \caption{Example XSUM summaries generated by Llama-3 model using the llm setting. }
    \label{tab:example_xsum_llm}
\end{table*}

\end{document}